\documentclass[format=acmsmall, review=false, screen=true]{acmart}

\usepackage{booktabs} % For formal tables
\usepackage{enumitem}

\usepackage{url}
\usepackage{mathrsfs}
\usepackage[mathscr]{eucal}
\usepackage{multirow}
\usepackage{xcolor,colortbl}

\usepackage{bm}
\usepackage{tikz}
\usetikzlibrary{arrows,positioning,automata,calc,shapes}
\usepackage{pgfplots}
\pgfplotsset{compat=newest, scaled z ticks=false} 
\pgfplotsset{plot coordinates/math parser=false}
\newlength\figureheight 
 \newlength\figurewidth

\usepackage{cleveref}

\usepackage[ruled]{algorithm2e} % For algorithms
\usepackage{setspace}

\SetAlFnt{\small}
\SetAlCapFnt{\small}
\SetAlCapNameFnt{\small}
\SetAlCapHSkip{0pt}
\IncMargin{-\parindent}

\renewcommand\footnotetextcopyrightpermission[1]{} % removes footnote with conference information in first column

\setcopyright{none}
\acmDOI{}

\begin{document}
% Title portion. Note the short title for running heads 
% * <song_3134@tamu.edu> 2018-04-23T01:25:38.283Z:
%
% ^.

\title{Tensor Completion Algorithms in Big Data Analytics}
%\author{Qingquan Song,{\small$^\dagger$} Hancheng Ge,{\small$^\dagger$} James Caverlee,{\small$^\dagger$} Xia Hu{\small$^{\dagger,\ddagger}$}}
%\affiliation{\\
%  \institution{$^\mathsmaller{\dagger}$Department of Computer Science and Engineering, Texas A\&M University, \\ $^\mathsmaller{\ddagger}$Center for Remote Health Technologies and Systems, Texas A\&M Engineering Experiment Station} 
 % }

%  \author{Qingquan Song, Hancheng Ge, James Caverlee, Xia Hu}
%\affiliation{\\
%  \institution{Department of Computer Science and Engineering, Texas A\&M University} 
%  }
% \email{{song_3134,hge,caverlee,xiahu}@tamu.edu}

  \author{Qingquan Song, Hancheng Ge, James Caverlee, Xia Hu}
\affiliation{\\
  \institution{Department of Computer Science and Engineering, Texas A\&M University \\ \{song\_3134, hge, caverlee, xiahu\}@tamu.edu } 
  }
   %  \email{{song_3134,hge,caverlee,xiahu}@tamu.edu}

\renewcommand{\shortauthors}{Q. Song et al.}

\fancyhead{}

\begin{abstract}
Tensor completion is a problem of filling the missing or unobserved entries of partially observed tensors.  Due to the multidimensional character of tensors in describing complex datasets, tensor completion algorithms and their applications have received wide attention and achievement in areas like data mining, computer vision, signal processing, and neuroscience. In this survey, we provide a modern overview of recent advances in tensor completion algorithms from the perspective of big data analytics characterized by diverse variety, large volume, and high velocity. We characterize these advances from four perspectives: general tensor completion algorithms, tensor completion with auxiliary information (variety), scalable tensor completion algorithms (volume), and dynamic tensor completion algorithms (velocity). Further, we identify several tensor completion applications on real-world data-driven problems and present some common experimental frameworks popularized in the literature. Our goal is to summarize these popular methods and introduce them to researchers and practitioners for promoting future research and applications. We conclude with a discussion of key challenges and promising research directions in this community for future exploration.
%and give an available repository for practitioners
\end{abstract}

%\begin{CCSXML}
%<ccs2012>
%  <concept_id>10010520.10010553.10010562</concept_id>
%  <concept_desc>Computer systems organization~Embedded systems</concept_desc>
%  <concept_significance>500</concept_significance>
%  <concept_id>10010520.10010575.10010755</concept_id>
%  <concept_desc>Computer systems organization~Redundancy</concept_desc>
%  <concept_significance>300</concept_significance>
%  <concept_id>10010520.10010553.10010554</concept_id>
%  <concept_desc>Computer systems organization~Robotics</concept_desc>
%  <concept_significance>100</concept_significance>
%  <concept_id>10003033.10003083.10003095</concept_id>
%  <concept_desc>Networks~Network reliability</concept_desc>
%  <concept_significance>100</concept_significance>
%</ccs2012>  
%\end{CCSXML}
%
%\ccsdesc[500]{Computer systems organization~Embedded systems}



\maketitle

\renewcommand{\shortauthors}{Q. Song et al.}

\section{Introduction}
Tensor analysis has garnered increasing attention in recent years. In one direction, tensor  decomposition -- which aims to distill tensors into (interpretable) soft representations --  has been widely studied from both a theoretical and application-specific perspective~\cite{Kolda:Tensor,lu2011survey,Papalexakis:Tensor,sidiropoulos2016tensor}. In a related direction and the focus of this survey, \textit{tensor completion} aims to impute missing or unobserved entries of a partially observed tensor. Tensor completion is one of the most actively studied problems in tensor related research and yet diffusely presented across many different research domains. On the one hand, multidimensional datasets are often raw and incomplete owing to various unpredictable or unavoidable reasons such as mal-operations, limited permissions, and data missing at random~\cite{liu2009tensor,Ge:Tensor,liu2014factor}. On the other hand, in practice, due to the multiway property of modern datasets, tensor completion natural arises in data-driven applications such as image completion and video compression.

%-- is growing in research attention but diffusely across many research areas 

 %Practical demand raises the issue of handling missing data in multi-way analysis evolving into the problem of tensor completion.   }

%Tensor completion or imputation diffusely presents in various applications~\cite{??,??,??}.

%{\blue Why tensor decomposition can not be directly applied in tensor completion? }

%The objectives of those two directions (completion vs decomposition) seem very different: distilling a multi-aspect dataset into interpretable soft clusterings vs filling the missing entries.

%such as link prediction, recommender system, urban computing, information diffusion, image completion, video completion, compressed sensing, healthcare, chemometrics, climate data analytics, etc. 

In the past few decades, the matrix completion problem -- a special case of tensor completion --  has been well-studied. Mature algorithms~\cite{cai2010singular}, theoretical foundations~\cite{candes2012exact} and various applications~\cite{candes2010matrix} pave the way for solving the completion problem in high-order tensors. Intuitively, the tensor completion problem could be solved with matrix completion algorithms by downgrading the problem into a matrix level, typically by either slicing a tensor into multiple small matrices or unfolding it into one big matrix. However, several problems distinguish tensor completion from being treated as a straightforward extension of the matrix completion problem. First, it has been shown that matrix completion algorithms may break the multi-way structure of a tensor and lose the redundancy among modes to improve the imputation accuracy~\cite{signoretto2011tensor}. While many tensor-based algorithms directly build upon matrix completion algorithms~\cite{mu2014square,xu2013parallel}, their key focus is out of the context of matrix level, i.e., trying to develop delicate ways of matricization to keep the multi-way properties of a tensor object while using matrix-based completion algorithms. Second, the high-order characteristics introduce even higher space and computational complexity, which may prevent the usage of traditional matrix completion algorithms.

% the matricization process may cause the scalability issue
%  cause the effectiness decay 
%the first choice would break the multi-way structure of a tensor and lose the connection of because of the data conn

%completing slicing a high order tensor into lower level matrices or matricizing it into a large transformed into the matrix level by either {\blue Why matrix completion cannot be directly used?}

\begin{figure}
\centering
\vspace{-0.1cm}
\includegraphics[height=3.7cm, width=13cm]{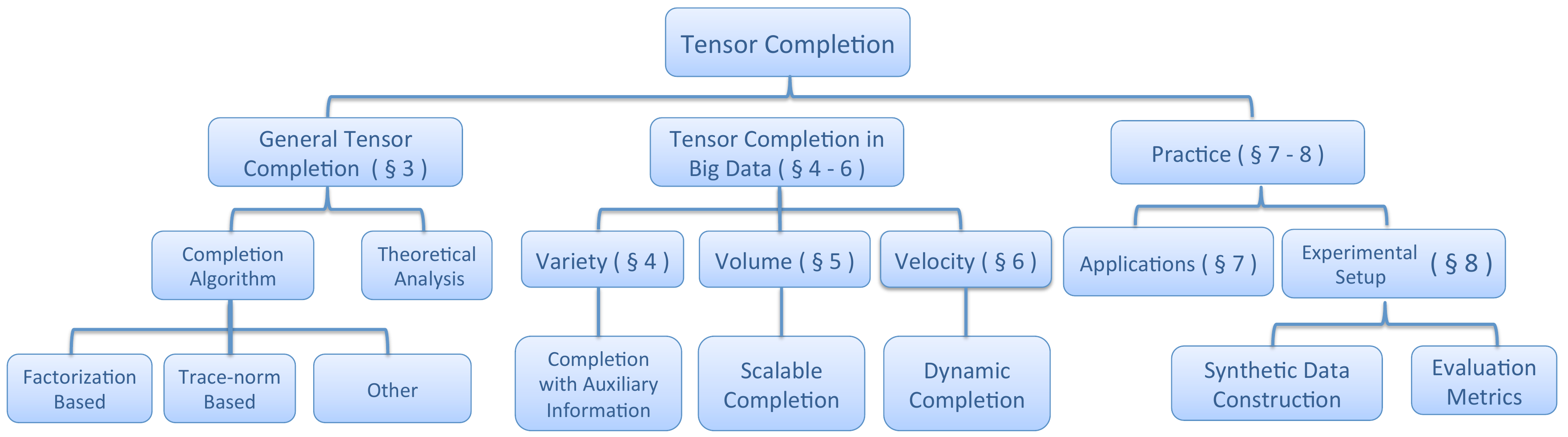} %\vspace{-.4cm} 
\vspace{-0.2cm} 
\caption{Outline of this survey.}
\label{fig:outline}  
\vspace{-0.4cm} 
\end{figure}

In some early works~\cite{bro1998multi,tomasi2005parafac,andersson1998improving,smilde2005multi}, tensor completion is often considered as a byproduct when dealing with missing data in the tensor decomposition problem. However,  due to the distinct objective and challenges, it gradually becomes an independent topic jumping out of the context of tensor decomposition. Concretely, tensor completion is a specific task, whose ultimate goal is to fill the missing or unobserved entries while tensor decomposition is an intermediate step to distill the tensors into (interpretable) soft representations to benefit subsequent tasks. It is feasible to use tensor decomposition methods to solve the completion problem. However, without careful treatment of the missing entries, some decomposition methods may not achieve desirable completion results, e.g., such as the vanilla alternating least square methods for CANDECOMP/PARAFAC decomposition~\cite{carroll1970analysis,harshman1970foundations}.

Though many efforts have focused on the tensor completion problem in recent years, to the best of our knowledge, we still lack a comprehensive survey to capture the many advances across domains. Existing surveys rest on the matrix level~\cite{johnson1990matrix,laurent2009matrix, koren2009matrix, su2009survey} or focus on related topics, e.g., tensor decomposition~\cite{lu2011survey, Kolda:Tensor,grasedyck2013literature,kroonenberg2008applied,kroonenberg1983three,henrion1994n,kiers2001three,comon2004canonical,faber2003recent,Papalexakis:Tensor,sidiropoulos2016tensor,smilde2005multi,acar2009unsupervised}. Moreover, with the rapid growth of real-world big data applications demonstrating variety, velocity, and volume, there has been a commensurate rise in tensor completion algorithms that are capable of leveraging heterogeneous data sources, handling real large-scale datasets, as well as tracking dynamical changes over time. The incremental interest in big data analytics~\cite{kolda2008scalable,Sael:Tensor,kang2012gigatensor} motivates us to provide a view from big data perspective in the present survey which differs from most of the existing reviews. Hence, the goal of this survey is to give an overview of general and advanced high-order tensor completion methods and their applications in different fields. We aim at summarizing the state-of-the-art tensor completion methods for promoting the research process in this area as well as providing a handy reference handbook for both researchers and practitioners. 

%Python repository as a handy tool 
%We aim at summarizing the state-of-the-art tensor completion methods for promoting the research process in this area while providing an available Python repository as a handy tool for both researchers and practitioners. 

Tightly coupling with the ``3V'' challenges~\cite{beyer2011gartner}
 \footnote{As described recently in~\cite{yin2015big}, Big data has ``5V'' challenges including ``variety'', ``volume'', ``velocity'', ``veracity'', and ``value''.  Veracity refers to the quality or uncertainty (trustworthiness) of the collected data and value represents the worth of the data being extracted. We focus on the first three for the convenience of dividing the algorithms from the model perspective. } in big data analytics including \textbf{variety}, \textbf{volume}, and \textbf{velocity}, we organize the structure of this article as shown in Figure~\ref{fig:outline}. Notations, primary multilinear algebra operations are first introduced to formulate the tensor completion problem and its variations. A concise summary of general tensor completion algorithms is provided along with several statistical assumptions and theoretical analysis. Subsequently, coping with the ``3V'' challenges in big data analytics, three categories of advanced completion methods are introduced including tensor completion with auxiliary information (\textbf{variety}), scalable tensor completion algorithms (\textbf{volume}), and dynamic completion methods (\textbf{velocity}). Real-world applications in different areas are outlined in Section~\ref{sec:app}, and several  experimental simulations and metrics are presented in Section~\ref{sec:exp}. Finally, we describe open challenges and promising directions.

\section{Tensor Completion Problem}
In this section, we provide the formal definition of the tensor completion problem. We begin by introducing several tensor operations, basic definitions, and notations following~\cite{Kolda:Tensor}. Table~\ref{Symbols} summarizes the notations used throughout this article.

\begin{table}[t!]
  \begin{center}
    \begin{tabular}[0.1\textwidth]{ c|c } \hline 
        Notations        &  Definitions  \\ \hline \hline
     $\pmb{\mathscr{X}} \in \mathbb{R}^{I_1\times I_2 \times \ldots \times I_N}$  & $N^{\text{th}}$-order tensor \\ \hline 
     $\mathbf{X}_{(n)} \in \mathbb{R}^{I_n\times (\Pi_{i\ne n}^N I_i) }$  & Mode-$n$ unfolding matrix of tensor $\pmb{\mathscr{X}}$   \\ \hline 
    $\pmb{\mathscr{X}}_{i_1,\ldots,i_N}$ or $\pmb{\mathscr{X}}({i_1,\ldots,i_N})$ & An entry of tensor $\pmb{\mathscr{X}}$ indexed by $[i_1,\ldots,i_N]$ \\ \hline
    $<\cdot, \cdot>$  & Inner product \\ \hline
     $\circ$  &Outer product \\ \hline
     $ \llbracket \cdot  \rrbracket $ & Kruskal operator \\\hline  %, e.g., $\pmb{\mathscr{X}} \approx   \llbracket \mathbf{A}_1, \ldots, \mathbf{A}_N \rrbracket$ \\ \hline 
     $\otimes$  &Kronecker product \\ \hline 
     $\odot$  &Khatri-Rao product \\ \hline 
     $\ast$ & Hadamard product \\ \hline 
     $\|\cdot \|_\text{F}$ & Frobenius norm \\ \hline 
     $\|\cdot \|_{\ast}$ & Trace norm \\ \hline

%     $(\mathbf{A}_k)^{\odot_{k\neq n}}$  &$\mathbf{A}_N\odot  \ldots  \odot \mathbf{A}_{n+1} \odot \mathbf{A}_{n-1} \odot \ldots \odot \mathbf{A}_{1}$ \\ \hline 
%     $(\mathbf{A}_k)^{\ast_{k\neq n}}$ &$\mathbf{A}_N \ast  \ldots  \ast \mathbf{A}_{n+1} \ast \mathbf{A}_{n-1} \ast \ldots \ast \mathbf{A}_{1}$ \\ \hline 
      %$\ast$  & Hadamard Product  \\ \hline 
     %$\llbracket A,B,C\rrbracket $  & the final embedding representation  \\ \hline 
      \hline  
    \end{tabular}
  \end{center}
   \vspace{-2pt}
        \caption{ Main symbols and operations.}
\label{Symbols}
\vspace{-19pt}
\end{table}

\subsection{Preliminaries}
%\subsection{Basic Tensor Operation, Definition and Notation}

\noindent \textbf{Definition 2.1.1 (Tensor).} A tensor can be defined in different ways at various levels of abstraction~\cite{kline1990mathematical,lee2012introduction}. We follow the most general way and define it as a \underline{\emph{multidimensional array}}. The dimensionality of it is described as its \underline{\emph{order}}. An $N^{\mbox{th}}$-order tensor is an N-way array, also known as N-dimensional or N-mode tensor, denoted by $\pmb{\mathscr{X}}$. We use the term  \underline{\emph{order}} to refer to the dimensionality of a tensor (e.g., $N^{\mbox{th}}$-order tensor), and the term \underline{\emph{mode}} to describe operations on a specific dimension (e.g., mode-$n$ product).

\begin{figure}
\centering
\vspace{-0cm}
\includegraphics[height=2.5cm, width=12.5cm]{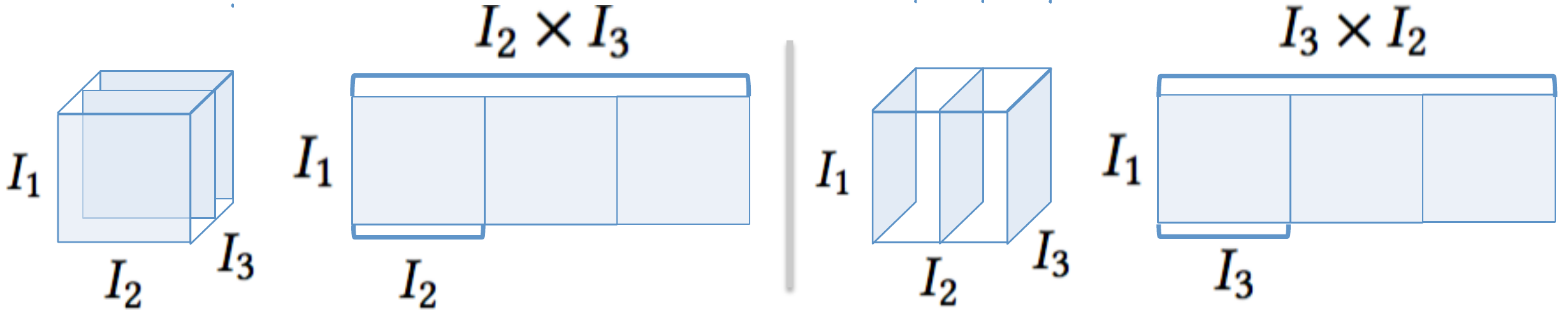} %\vspace{-.4cm} 
\vspace{-0.2cm} 
\caption{Forward (left) and backward (right) mode-one matricization ($ \mathbf{X}_{(1)} $) of the tensor $\pmb{\mathscr{X}}   \in \mathbb{R}^{I_1\times I_2 \times I_3}$.     }
\label{fig:mat}  
\vspace{0.0cm} 
\end{figure}

\noindent \textbf{Definition 2.1.2 (Tensor Matricization).} Tensor matricization is to unfold a tensor into a matrix format with a predefined ordering of its modes. The most commonly used tensor matricization is mode-$n$ matricization (a.k.a., mode-$n$ unfolding), which is to unfold a tensor $\pmb{\mathscr{X}}\in \mathbb{R}^{I_1\times  \ldots \times I_N}$ along its $n^\text{th}$ mode into a matrix denoted as $\mathbf{X}_{(n)}$ of size $I_n \times (\prod_{k \neq n} I_k)$. The ordering of the other modes except mode $n$ can be arranged randomly to construct the column of $\mathbf{X}_{(n)}$. Two commonly used arrangements are forward~\cite{kiers2000towards} (i.e. $n+1,\ldots,N, 1, \ldots, n-1$) and backward~\cite{de2000multilinear} (i.e. $n-1,\ldots,1, N, \ldots, n+1$). We depict such a forward and backward mode-one unfolding of a third-order tensor $\pmb{\mathscr{X}} \in \mathbb{R}^{I_1\times I_2 \times I_3} $ in Figure~\ref{fig:mat}. In general, the specific permutation is not important as long as it is consistent across related calculations~\cite{Kolda:Tensor}. More comprehensive comparison and visualizations can be found in~\cite{bader2006algorithm}.

%$\pmb{\mathscr{X}}(:,:,1)$
%
%$\pmb{\mathscr{X}}(:,:,2)$
%
%$\pmb{\mathscr{X}}(:,:,3)$  
%
%$\pmb{\mathscr{X}}(:,:,I_3)$ 
%
%$\pmb{\mathscr{X}}(:,1,:)$
%
%$\pmb{\mathscr{X}}(:,2,:)$
%
%$\pmb{\mathscr{X}}(:,3,:)$  
%
%$\pmb{\mathscr{X}}(:,I_2,:)$ 
%
%
%$\pmb{\mathscr{X}}(1,:,:)$
%
%$\pmb{\mathscr{X}}(2,:,:)$
%
%$\pmb{\mathscr{X}}(3,:,:)$  
%
%$\pmb{\mathscr{X}}(I_1,:,:)$ 
%
%
%$\pmb{\mathscr{X}}$
%
%
%$I_1$
%
%$I_2$
%
%$I_3$
%
%$I_1$
%
%$I_2\times I_3$
%
%$I_3\times I_2$
%
%$\mathbf{X}_{(1)}$
%
%$\mathbf{X}_{(2)}$
%
%$\mathbf{X}_{(3)}$
%
%$\pmb{\mathscr{X}}\in \mathbb{R}^{I_1\times I_2\times I_3}$
%
%{\blue add a figure for matrix unfolding}

\noindent \textbf{Definition 2.1.3 (Product Operations)}. 
\begin{itemize}[leftmargin=*]
\item Outer Product: product of vectors denoted by $\circ$. For $N$ column vectors ${\bf a}^{(1)}\in\mathbb{R}^{I_1}, \ldots ,{\bf a}^{(N)} \in\mathbb{R}^{I_N}$. The outer product among them is defined as:
\begin{equation}
\pmb{\mathscr{T}} = {\bf a}^{(1)} \circ \ldots \circ {\bf a}^{(N)}, ~\pmb{\mathscr{T}}_{i_1,\ldots,i_N}={\bf a}^{(1)}_{i_1}\cdots {\bf a}^{(N)}_{i_N}, 
\end{equation}
where $\pmb{\mathscr{T}} \in \mathbb{R}^{I_1\times \ldots \times I_N}$. It is the same as matrix multiplication when $N=2$.

%For of $N$ arbitrary column vectors ${\bf a}$_1 and ${\bf b}$ It is the same with matrix multiplication , denoted by ${\bf a} \circ {\bf b}$.
\item Kronecker Product: product of two matrices. Given two matrices ${\bf A}$ and ${\bf B}$ of sizes $m \times n$ and $p \times q$, respectively, their kronecker product is defined as: 
\begin{equation}
\mathbf{A}_{m\times n}\otimes\mathbf{B}_{p \times q} = \begin{bmatrix} a_{11} \mathbf{B} & \cdots & a_{1n}\mathbf{B} \\ \vdots & \ddots & \vdots \\ a_{m1} \mathbf{B} & \cdots & a_{mn} \mathbf{B} \end{bmatrix}_{mn \times pq}, ~\mbox{where}~ \mathbf{A}_{m\times n} = \begin{bmatrix} a_{11}  & \cdots & a_{1n} \\ \vdots & \ddots & \vdots \\ a_{m1} & \cdots & a_{mn} \end{bmatrix}_{m \times n}.   
\end{equation}

\item Khatri-Rao Product: block-wise Kronecker product. Here, we treat it as column-wise Kronecker product of two matrices with same column size, which is defined as:
\begin{equation}
\mathbf{A}_{m\times N} \odot \mathbf{B}_{p\times N}=[\mathbf{a}_1\otimes\mathbf{b}_1,\cdots,\mathbf{a}_N\otimes\mathbf{b}_N]_{mp\times N}, 
\end{equation}
where $\mathbf{A}_{m\times N} =[\mathbf{a}_1,\cdots,\mathbf{a}_N]_{m\times N}$ and $\mathbf{B}_{p\times N} =[\mathbf{b}_1,\cdots,\mathbf{b}_N]_{p\times N}$. 

\item Hadamard Product: element-wise product of two tensors with the same size, denoted as $\pmb{\mathscr{A}} \ast \pmb{\mathscr{B}} $. Its inverse division operation is denoted as $\pmb{\mathscr{A}} \oslash \pmb{\mathscr{B}}$.

\item mode-$n$ Product: multiplication  of a given tensor $\pmb{\mathscr{X}}\in \mathbb{R}^{I_1\times I_2 \times \ldots \times I_N}$ on its $n^{\text{th}}$-mode with a matrix $\mathbf{U}\in \mathbb{R}^{I_n\times J}$, which is denoted as $\pmb{\mathscr{Z}}=\pmb{\mathscr{X}}\times_n \mathbf{U}$, where $\pmb{\mathscr{Z}}\in \mathbb{R}^{I_1 \ldots I _{n-1}  \times J \times  I_{n+1} \ldots  I_N}$. The elementwise result is described as:
\begin{equation}
\pmb{\mathscr{Z}}_{i_1,\ldots,i_{n-1}, j,  i_{n+1}, \ldots , i_N } =  \sum_{k=1}^{I_n}   \pmb{\mathscr{X}}_{i_1,\ldots,i_{n-1}, k,  i_{n+1}, \ldots , i_N } \mathbf{U}_{k,j}.
\end{equation}

\item General Tensor Product (Multiplication): the general tensor (outer) product of tensors. Given two tensors $\pmb{\mathscr{X}}\in \mathbb{R}^{I_1\times  \ldots \times I_N}$ and $\pmb{\mathscr{Y}}\in \mathbb{R}^{J_1\times  \ldots \times J_N}$, there tensor product is defined as $\pmb{\mathscr{Z}}=\pmb{\mathscr{X}}  \circ \pmb{\mathscr{Y}}$, where $\pmb{\mathscr{Z}}\in \mathbb{R}^{I_1  \times \ldots \times I_N \times J_1 \times \ldots \times J_N}$ and the elementwise result is described as:
\begin{equation}
\pmb{\mathscr{Z}}_{i_1,\ldots,i_N, j_1, \ldots , j_N } =   \pmb{\mathscr{X}}_{i_1,\ldots,i_N}   \pmb{\mathscr{Y}}_{j_1, \ldots , j_N }. 
\end{equation}

\end{itemize}

See also Bader and Kolda~\cite{bader2006algorithm} for a detailed treatment of tensor multiplication.

\noindent \textbf{Definition 2.1.4 (Rank-1 Tensor).} Also called simple tensor~\cite{hernandez2010simple} or decomposable tensor~\cite{hackbusch2012tensor}. It is an $N^{\mbox{th}}$-order tensor $\pmb{\mathscr{X}}~(N \in \mathbb{Z}_{+})$ which could be written as the outer product of $N$ vectors, i.e., $\pmb{\mathscr{X}} = \mathbf{a}^{(1)} \circ \mathbf{a}^{(2)} \circ \ldots \circ \mathbf{a}^{(N)}$.

\noindent \textbf{Definition 2.1.5  (Tensor (CP) Rank).} 
The tensor rank (i.e. tensor CP rank) of a tensor $\pmb{\mathscr{X}}$ is defined as the minimum number of summations of rank-one tensors that generate  $\pmb{\mathscr{X}}$~\cite{Hitchcock:1,kruskal1977three}, i.e., 
\begin{equation}
\begin{aligned}
\text{rank}_{\text{CP}}(\pmb{\mathscr{X}}) \triangleq \min \big\{ R \in \mathbb{Z}_{+}:
 \exists~\{\mathbf{a}_r^{(n)}\},  s.t.~\pmb{\mathscr{X}}=\sum_{r=1}^R \mathbf{a}_r^{(1)} \circ \mathbf{a}_r^{(2)} \circ \ldots \circ \mathbf{a}_r^{(N)}  \big\}.
\end{aligned}
\end{equation}

\noindent \textbf{Definition 2.1.6 (Tensor $n$-rank)).} As calculating the tensor rank is an NP-hard problem~\cite{haastad1990tensor}, the tensor $n$-rank was introduced by Kruskal~\cite{Kruskal1989rank} as a special case of multiplex rank introduced by Hitchcock~\cite{Hitchcock:2}. The tensor $n$-rank is defined as the rank (usually column rank) of $\mathbf{X}_{(n)}$, i.e., $\text{rank}_n(\pmb{\mathscr{X}})=\text{rank}(\mathbf{X}_{(n)})$. Another similar definition called Tucker rank or multilinear rank~\cite{kasai2016low}, which is introduced earlier by Tucker~\cite{tucker1966some} is defined as $\text{rank}_{\text tc}(\pmb{\mathscr{X}})= (\text{rank}(\mathbf{X}_{(1)}),\ldots,\text{rank}(\mathbf{X}_{(N)}))$.

\noindent \textbf{Definition 2.1.7 (Tensor Inner Product).} The inner product of two tensors $\pmb{\mathscr{X}}$ and $\pmb{\mathscr{Y}}$ of same size is defined as $<\pmb{\mathscr{X}},\pmb{\mathscr{Y}}>$. Unless otherwise specified, we treat it as dot product defined as follows:
\begin{equation}
\begin{aligned}
<\pmb{\mathscr{X}},\pmb{\mathscr{Y}}>=\sum_{i_1=1}^{I_1} \sum_{i_2=1}^{I_2}\cdots \sum_{i_N=1}^{I_N} \pmb{\mathscr{X}}_{i_1,i_2,\ldots,i_N} \pmb{\mathscr{Y}}_{i_1,i_2,\ldots,i_N}     .
\label{equ:CP0}
\end{aligned}
\end{equation}

\noindent \textbf{Definition 2.1.8 (Tensor Frobenius Norm).} Generalized from matrix Frobenius norm, the F-norm of a tensor $\pmb{\mathscr{X}}$ is defined as: 
\begin{equation}
\begin{aligned}
\| \pmb{\mathscr{X}} \|_{\text{F}} =\sqrt{<\pmb{\mathscr{X}},\pmb{\mathscr{X}}>}=\sqrt{\sum_{i_1=1}^{I_1} \sum_{i_2=1}^{I_2}\cdots \sum_{i_N=1}^{I_N} \pmb{\mathscr{X}}_{i_1,i_2,\ldots,i_N}^2}.
\label{equ:CP}
\end{aligned}
\end{equation}

%For more details regarding the tensor decomposition problem, readers are referred to \cite{Kolda:Tensor,Papalexakis:Tensor,sidiropoulos2016tensor}, since here we are interested in tensor completion problem.

%{\red \noindent \textbf{Definition 2.1.9 (Matrix Trace (Nuclear) Norm).} The matrix trace (nuclear) norm is defined as ... . For the tensor case, as ..., different researchers provide different definitions, which will be further described in Section~\ref{sec:trace}.  }{\blue do we need this? or define the tensor trace norm in a different way?}{\blue  Noisy tensor completion via the sum-of-squares hierarchy~\cite{barak2016noisy}}

\subsection{Tensor Completion Problem}\label{subsec:TCP}
Tensor completion is defined as the problem of filling the missing elements of partially observed tensors. As its particular matrix case~\cite{candes2012exact}, to avoid being an underdetermined and intractable problem, \emph{low rank} is often a necessary hypothesis to restrict the degree of freedoms of the missing entries~\cite{signoretto2011tensor,gandy2011tensor, liu2013tensor,kressner2014low}. Since a tensor has different types of rank definitions, to give a relatively general mathematical formulation of the low-rank tensor completion (LRTC) problem, we first summarize several most of the most popular definitions~\cite{liu2013tensor, gandy2011tensor} into a unified optimization problem and then specify the variants derived from it by answering several questions. %{\blue do we need to mention more about why we only focus on low-rank?}

%The LRTC problem is sometimes considered as a byproduct of the tensor decomposition problem with missing values but could be  formulated as a more general imputation problem with given rank (or its upper bound) as follows:

\noindent \textbf{Definition 2.2.1 (Low-rank Tensor Completion Problem).} Given a low-rank (either CP rank or other ranks) tensor $\pmb{\mathscr{T}} $ with missing entries, the goal of completing it can be formulated as the following optimization problem:  %{\blue do we need to explain why people minimize rank?}
\begin{equation}
\begin{aligned}
\centering
& \underset{ \pmb{\mathscr{X}} }{\text{minimize}}
&&   \text{rank}_{\ast}(\pmb{\mathscr{X}})\\
& \text{subject to}
&& \pmb{\mathscr{X}}_{\pmb{\Omega}}= \pmb{\mathscr{T}}_{\pmb{\Omega}}, 
\end{aligned}
\label{equ:TCP0}
\end{equation}
where $\text{rank}_{\ast}$ denotes a specific type of tensor rank based on the rank assumption of the given tensor $\pmb{\mathscr{T}}$, $\pmb{\mathscr{X}}$ represents the completed low-rank tensor of $\pmb{\mathscr{T}}$, and $\pmb{\Omega}$ is an index set denoting the indices of observations. The intuitive explanation of the above equation is that: we expect to find a tensor $\pmb{\mathscr{X}}$ with the minimum rank, which subjects to the equality constraints given by the observations (a.k.a., measurements). This equation is a straightforward generalization of the well-understood matrix completion problem~\cite{candes2012exact} and could be treated as the starting point of almost every existing variant definitions of LRTC problem. To delve into these variants, we start by asking three questions with respect to the above equation and try to explore the answers from these variants. (1) What type of the rank should we use? (2) Are there any other constraints defined based on the observations that we could presume? (3) In what condition, could we expect to achieve a unique and exact completion? We mainly focus on the first two questions here to extend our discussion and leave the last question into Section~\ref{sec:stat} on account of its tight correlation with various statistical assumptions. 

%For the last question, since they are highly correlated with various statistical assumptions, we only briefly discuss them here and leave the specification in Section~\ref{sec:stat} after introducing some basic tensor completion algorithms.
%to provide the varietal mathematical definitions of the tensor completion problem

\subsubsection{Rank Variants}

We first discuss the first question (i.e., What type of the rank should we use?) and describe some variants of Equation~\eqref{equ:TCP0} based on different rank selections. In the preliminary part, we have defined two different types of ranks: the tensor (CP) rank and the tensor n-rank. As they have covered most of existing popular articles, we now focus on introducing the main variant optimization problem utilizing these two types of ranks.

\emph{(1) Tensor (CP) rank and its corresponding variants of the LRTC problem. }

As existing works have demonstrated that calculating the tensor (CP) rank is an NP-hard problem~\cite{haastad1990tensor, Kolda:Tensor}, directly considering the minimization problem defined in Equation~\eqref{equ:TCP0} with CP rank is unpractical. To avoid this problem, some researchers~\cite{krishnamurthy2013low,jain2014provable,barak2016noisy,ashraphijuo2017fundamental} assume the CP rank of the target tensor is fixed to achieve a relative milder problem. By fixing the rank of $\pmb{\mathscr{T}}$, they are able to substitute the original objective function with some polynomial time computable norms of the tensor $\pmb{\mathscr{X}}$~\cite{barak2016noisy} or treat the CP rank as the constraint and inversely minimize the difference between the true observations and their predictions as follows~\cite{jain2014provable}:
\begin{equation}
\begin{aligned}
\centering
& \underset{ \pmb{\mathscr{X}} }{\text{minimize}}
&& \mathcal{D}( \,\pmb{\mathscr{T}}_{\pmb{\Omega}}\,   ,  \, \pmb{\mathscr{X}}_{\pmb{\Omega}} ) \\
& \text{subject to}
&& \text{rank}_{\text{CP}}(\pmb{\mathscr{X}})=r,
\end{aligned}
\label{equ:TCP1}
\end{equation}
where $\mathcal{D}$ is the error measure between $\pmb{\mathscr{X}}_{\pmb{\Omega}}$ and $\pmb{\mathscr{T}}_{\pmb{\Omega}}$, which is often defined as the Frobenius norm of their difference under the assumption of Gaussian noise, i.e.,  $\mathcal{D}( \,\pmb{\mathscr{T}}_{\pmb{\Omega}}\,   ,  \, \pmb{\mathscr{X}}_{\pmb{\Omega}} ) = \|   \pmb{\mathscr{X}}_{\pmb{\Omega}} - \pmb{\mathscr{T}}_{\pmb{\Omega}} \|_\text{F}$ . Though fixing CP rank helps solve the problem, this assumption is always not an ideal choice since estimating the tensor rank is often very hard ~\cite{Kruskal1989rank}, which confines the applicability of corresponding completion algorithms. Thus, some researchers only assume the predefined CP rank is the rank upper bound~\cite{liu2014factor} or propose the algorithms, which could dynamically conduct the rank search during the completion process~\cite{yokota2016smooth}.

\emph{(2) Tensor n-rank and its corresponding variants of the LRTC problem. }

Comparing with other ranks, Tensor n-rank (or Tucker rank) is perhaps the most widely adopted rank assumption in existing tensor completion literature~\cite{liu2013tensor,signoretto2010nuclear,gandy2011tensor,kressner2014low}. 
As it is defined based on the matricizations of the tensor, many matrix completion techniques such as trace-norm based methods could be generalized into the high-order level~\cite{liu2013tensor,gandy2011tensor,signoretto2010nuclear}. Most of the existing work targets on solving the following tensor n-rank minimization problem described in~\cite{gandy2011tensor}: 
\begin{equation}
\begin{aligned}
\centering
& \underset{ \pmb{\mathscr{X}} }{\text{minimize}}
%&&  f( \text{rank}_1(\pmb{\mathscr{X}}),\text{rank}_2(\pmb{\mathscr{X}}),\ldots,\text{rank}_N(\pmb{\mathscr{X}}))\\
&&  f( \text{rank}_{\text{tc}}(\pmb{\mathscr{X}}))\\
& \text{subject to}
&& \pmb{\mathscr{X}}_{\pmb{\Omega}}= \pmb{\mathscr{T}}_{\pmb{\Omega}},
%&& \mathcal{A}(\pmb{\mathscr{X}})= \pmb{\mathscr{B}},
\end{aligned}
\label{equ:TCP3}
\end{equation}
where $f$ is a predefined function for the tensor n-rank: $\text{rank}_{\text tc}(\pmb{\mathscr{X}})= (\text{rank}(\mathbf{X}_{(1)}),\ldots,\text{rank}(\mathbf{X}_{(N)}))$. For example, the most commonly used function is the rank summation function~\cite{gandy2011tensor, romera2013new}, i.e.,  $f( \text{rank}_{\text{tc}}(\pmb{\mathscr{X}})) = \sum_{i=1}^N \text{rank}(\mathbf{X}_{(i)})$. Standing upon the above optimization problem, there are also several variant ways of using tensor n-rank to formulate the LRTC problem. For example, Kressner et al.~\cite{kressner2014low} assume the tensor n-rank to be fixed, i.e., $\text{rank}_\text{tc}(\pmb{\mathscr{X}})  = (r_1,\ldots,r_N)$ and then use the similar way defined in Equation~\eqref{equ:TCP1} to formulate the completion problem. This definition allows them to explicitly leverage the tensor decomposition power to achieve the completion purpose. A correlated variant is to assume the tensor n-rank to be constrained~\cite{mu2014square}, i.e.,$\text{rank}_\text{tc}(\pmb{\mathscr{X}}) \preceq (r_1,\ldots,r_N)$, where $\preceq$ is an element-wise notation. Though tensor n-rank has been widely employed, recent works~\cite{phien2016efficient} also point out a crucial drawback that it only takes into consideration the ranks of matrices that are constructed based on the unbalanced matricization scheme, i.e., one mode versus the rest. 

%{\red QQ: I DONT UNDERSTAND THIS SENTENCE: It helps to benefit the proves of some statistical properties such as the lower bound of the number of observations for exact tensor recovery.} 

The drawbacks of utilizing tensor CP rank and tensor n-rank encourage researchers to propose other  types of ranks such as tensor-train (TT) rank~\cite{oseledets2011tensor,phien2016efficient,grasedyck2015alternating} and tensor tubal rank~\cite{zhang2017exact}. However, as the rank of a high-order tensor is still not a well-understood area, which is contrary to the matrix case, there is still no absolute conclusion that applying one rank is always better than another. Readers who are interested in the variant mathematical definition of the LRTC problem using other ranks could explore the corresponding papers for a more detailed introduction.

%(2) Tensor-train (TT) rank and Tensor Tubal rank
%Tensor-train rank is defined b
%~\cite{oseledets2011tensor,phien2016efficient} ~\cite{zhang2017exact}

%where $\pmb{\Omega}$ denotes the sampling set

\subsubsection{Constraint Variants}
Another question which could introduce different variants for the general optimization problem~\eqref{equ:TCP0} is what kind of constraints defined based on the observations that we could presume? In Equation~\eqref{equ:TCP0}, the constraints are intuitively defined as $\pmb{\mathscr{X}}_{\pmb{\Omega}}= \pmb{\mathscr{T}}_{\pmb{\Omega}} $ based on the observations. This simple constraint usually implies two statistical assumptions: (1) The situation we considered is the noise-free case~\cite{liu2013tensor}. (2) We assume the observations are uniformly random sampled based on Bernoulli distribution~\cite{huang2014provable}.  Although these two assumptions are widely used in the LRTC literature, there are also several popular variants. Firstly, as data in real-world applications are often corrupted with various of noises, different noise assumptions, especially Gaussian noises, are commonly used by researchers~\cite{gandy2011tensor,goldfarb2014robust,barak2016noisy}, leading to the following noise-contained formulation:
\begin{equation}
\begin{aligned}
\centering
& \underset{ \pmb{\mathscr{X}} }{\text{minimize}}
&&   \text{rank}_{\ast}(\pmb{\mathscr{X}})\\
& \text{subject to}
&& \pmb{\mathscr{X}}_{\pmb{\Omega}}= \pmb{\mathscr{T}}_{\pmb{\Omega}} + \pmb{\mathscr{E}}_{\pmb{\Omega}},
\end{aligned}
\label{equ:TCP2}
\end{equation}
where $\pmb{\mathscr{E}}$ denotes the noise term. Similar definition is also used in Robust Tensor Completion analysis~\cite{goldfarb2014robust}, which will be discussed in Section~\ref{subsubsec:rob}. Secondly, the way of sampling observations could also be changed and reflected in the constraints. For example, some researchers~\cite{mu2014square, eldar2012uniqueness} use the Gaussian measurement operator $\mathcal{G}$ to denote this constraint as a sampling ensemble $\mathcal{G}[\pmb{\mathscr{X}}]=\mathcal{G}[\pmb{\mathscr{T}}]$. In this way, the constraint entries are no longer based on random sampling approach but based on Gaussian random sampling. Since the sampling strategy is also highly correlated with our last question (i.e., In what condition, we could expect to achieve a unique and exact completion?), we leave it to Section~\ref{sec:stat} for further discussion. Besides the above variants, the constraint in Equation~\eqref{equ:TCP0} could also be expressed in different ways such as $\mathcal{A}(\pmb{\mathscr{X}})= \pmb{\mathscr{B}}$ in \cite{gandy2011tensor} and $\mathcal{P}_{\pmb{\Omega}}(\pmb{\mathscr{X}})=\mathcal{P}_{\pmb{\Omega}}(\pmb{\mathscr{T}})$ in~\cite{jain2014provable,kressner2014low}, where $\mathcal{A}$ and $\mathcal{P}_{\pmb{\Omega}}$ are linear projection operators indicating the sampling operation along with the index set $\pmb{\Omega}$.

\section{General Tensor Completion Algorithms}
\label{sec:general}

After formally defining the LRTC problem, we now introduce some general tensor completion algorithms serving as a stepping stone for more detailed introductions of advanced algorithms in subsequent sections. We start from summarizing some fundamental tensor completion approaches by categorizing them into decomposition based approaches, trace-norm based approaches, and some other variants. These methods are by no means necessarily more straightforward, and non-scalable compared with the advance tensor completion methods. The reason we put them here is either due to their general and primary effect in laying the foundation of building up advanced tools or because of their original focus is not from the perspective of big data analytics.

%There is a rich variety of tensor decompositions in the literature. In this section, we provide a comprehensive overview of the most widely used decompositions in data mining, from a practitioner’s point of view.

\subsection{Decomposition Based Approaches}
Tensor decompositions or factorizations are powerful tools for extracting meaningful, latent structures in heterogeneous, multi-aspect data~\cite{Kolda:Tensor, Papalexakis:Tensor}. Since real-world datasets are often raw and incomplete, many early researches are conducted on tensor decomposition with missing values~\cite{bro1998multi, walczak2001dealing, tomasi2005parafac, acar2011scalable, jain2014provable}, which could be regarded as the pioneer problem of LRTC. We focus on two most widely used decomposition methods -- CP and Tucker decomposition -- to begin the overview. Readers interested in tensor decompositions and their applications could go further into~\cite{Kolda:Tensor,Papalexakis:Tensor,sidiropoulos2016tensor,acar2009unsupervised} for a more comprehensive introduction. Not surprisingly, other decomposition approaches could also be applied to solving the completion problem such as hierarchical tensor representations~\cite{rauhut2015tensor,da2013hierarchical,rauhut2017low}, PARAFAC2 models~\cite{rauhut2015tensor},  and so on.

%Rauhut et al.~\cite{rauhut2015tensor}{\red Tensor completion in hierarchical tensor representations}

%{\red Evangelia Pantraki and Constantine Kotropoulos. 2015. Automatic image tagging and recommendation via PARAFAC2}

%hierarchical tucker~\cite{da2013hierarchical}

%{\red Low Rank Tensor Recovery via Iterative Hard Thresholding} Tensor-train+hierarchical
%iterative hard thresholding method for fitting the factors of a Tucker decomposition~\cite{rauhut2017low}

% is , which can be formulated as~\eqref{equ:TCP1}, where $\pmb{\mathscr{X}}$ is the factorized low-rank approximation.

%Before , we first provide an compendious overview of several most commonly used decompositions including the canonical polyadic (CP) decomposition~\cite{Hitchcock:1,carroll1970analysis,harshman1970foundations}, 

\subsubsection{CANDECOMP/PARAFAC (CP) Based Methods}

CP Decomposition was first proposed by Hitchcock~\cite{Hitchcock:1} and further discussed by Carroll~\cite{carroll1970analysis} and Harshman~\cite{harshman1970foundations}. It is formally defined as: Given an $N^{\text{th}}$-order tensor  $\pmb{\mathscr{X}}$, its CP decomposition is an approximation of $N$ loading matrices $\mathbf{A}_n \in \mathbb{R}^{I_n\times R}, n=1,\ldots, N$, such that,
\begin{equation}
\begin{aligned}
\pmb{\mathscr{X}} \approx   \llbracket \mathbf{A}_1, \ldots, \mathbf{A}_N \rrbracket = \sum_{r=1}^R \mathbf{a}_r^{(1)} \circ \mathbf{a}_r^{(2)} \circ \ldots \circ \mathbf{a}_r^{(N)},
\label{equ:CP2}
\end{aligned}
\end{equation}
where $R$ is a positive integer denoting an upper bound of the rank of $\pmb{\mathscr{X}}$, $\mathbf{a}_r^{(n)}$ is the $r$-th column of matrix $\mathbf{A}_n$. We can further unfold $\pmb{\mathscr{X}}$ along its $n^{\text{th}}$ mode as follows,
\begin{equation}
\begin{aligned}
\pmb{\mathscr{X}} \quad \overset{\text{unfold}}{\Longrightarrow} \quad \mathbf{X}_{(n)} &  \approx  \mathbf{A}_{n}(\mathbf{A}_N \odot  \ldots \mathbf{A}_{n+1} \odot \mathbf{A}_{n-1}\ldots \odot \mathbf{A}_{1} )^\top  %\xrightarrow{\hspace*{0.6cm}}
 = \mathbf{A}_{n}{[(\mathbf{A}_k)^{\odot_{k\neq n}} ] }^\top.
\end{aligned}
\end{equation}

Borrowing the idea from weighted least square method applied in the matrix level~\cite{kiers1997weighted},  Bro \cite{bro1998multi} demonstrates two ways of handling missing data in PARAFAC model, which might be one of the earliest approaches targeting to missing data imputation problem in multi-way data analysis. 
These two ways were further illustrated in his subsequent paper~\cite{tomasi2005parafac} with Tomasi, and the underlying ideas were explicitly or implicitly contained in almost every existing tensor completion methods as far as we know.

%Since almost every existing tensor completion approach either explicitly or implicitly borrows one of these two types of ideas, we now introduce them here respectively and use them as the one of the ways to category different imputation methods discussion in this section.

%They proposed two PARAFAC models ALS-SI and INDAFAC based on these two ideas respectively. 
%which is the most popular algorithm for calculating CP decomposition proposed by Carroll and Chang~\cite{carroll1970analysis} and Harshman~\cite{harshman1970foundations}.

The first approach is to alternatively estimate the model parameters while imputing the missing data, which could be regarded as a special expectation maximization (EM) approach under the assumption of Gaussian residuals. Hence, we call it \textbf{EM-like Approach}. This approach is usually used in alternating projection optimizations. For example,  in the work of Tomasi's and Bro's~\cite{tomasi2005parafac}, they apply this idea in the PARAFAC model and propose a model called ALS-IS leveraging the standard alternating least squares (ALS) optimization Though not explicitly described, the idea of this approach also appears in some other early works for handling missing entries in PARAFAC model~\cite{kroonenberg1983three,bro1997parafac,kiers1999parafac2}. The main idea of this approach is to conduct imputation based on the following equation iteratively:
\begin{equation}
\pmb{\mathscr{X}}=\mathcal{P}_{\mathbf{\Omega}}(\pmb{\mathscr{T}})+\mathcal{P}_{{\mathbf{\Omega}}^c } (\pmb{\mathscr{\hat{X}}} )=\pmb{\mathscr{W}} \ast \pmb{\mathscr{T}} + (\mathbf{1} -\pmb{\mathscr{W}}) \ast \pmb{\mathscr{\hat{X}}},
\label{equ:CP1}
\end{equation}
where $\pmb{\mathscr{\hat{X}}} = \llbracket \mathbf{A}_1, \ldots, \mathbf{A}_N \rrbracket $ is the interim low-rank approximation based on CP decomposition and $\pmb{\mathscr{X}}$ is the recovered tensor which is used in next iteration for decomposition, $\mathbf{\Omega}^c $ denotes the complement set of $\mathbf{\Omega}$ defined as: $\mathbf{\Omega}^c =  \{ (i_1, \ldots, i_N )| 1 \le i_i \le I_i \} \setminus \mathbf{\Omega} $ , $\pmb{\mathscr{W}}$ is the observation index tensor with the same size as $\pmb{\mathscr{T}}$:
\begin{equation}
\pmb{\mathscr{W}}(i_1,i_2,...,i_N)=\begin{cases} 
1 \quad& \text{if $\pmb{\mathscr{T}}(i_1,i_2,...,i_N)$ is observed. }\\
0 \quad& \text{if $\pmb{\mathscr{T}}(i_1,i_2,...,i_N)$ is unobserved.}\\
\end{cases}
\label{equ:Omega}
\end{equation}

This approach is easy to be conducted because it only needs to follow certain alternation projection optimization scheme such as ALS optimization~\cite{tomasi2005parafac} while performing imputation at the end of every iteration based on~\eqref{equ:CP1}. However, with the increasing percentage of missing entries, the convergence rate of the methods based on this approach may be reduced, and the risk of converging to a local minimum would be increased~\cite{tomasi2005parafac}.

The second approach mentioned by Bro~\cite{bro1998multi} is to skip the missing value and build up the model based only on the observed part, which we call \textbf{Missing-Skipping (MS) approach}. This approach is often found in gradient-based optimizations or probabilistic methods. It can also be realized by masking the missing entries during optimization scheme. The objective function when using this approach is often defined in the following format:
\begin{equation}
\begin{aligned}
\centering
%& \underset{ \pmb{\mathscr{X}}   } {\text{minimize}}
J= \sum_{ (i,j,k) \in \mathbf{\Omega} }  \mathcal{D}(    \pmb{\mathscr{X}}_{i,j,k}, \pmb{\mathscr{T}}_{i,j,k} ),
%& \text{subject to}
%&& \text{Certain Constraints}
\end{aligned}
\label{equ:MS}
\end{equation}
where $\mathcal{D}$ is an error measure similar to the one defined in Equation~\eqref{equ:TCP1}. The MS approach for tensor completion is first employed in the INDAFAC (INcomplete DAta paraFAC) model~\cite{tomasi2005parafac} to solve the aforementioned effectiveness-decay problem of the EM-like approach. This model is optimized by the Levenberg-Marquardt version of Gauss-Newton (iterative) algorithm inspired by Pentti~\cite{paatero1997weighted} and is proved to be computationally efficient when the missing ratio is high. Acar~\cite{acar2011scalable} also uses the second idea and develop an algorithm named CP-WOPT (CP Weighted OPTimization). Different from a second-order approach employed by INDAFAC, CP-WOPT is optimized based on a first-order gradient and shown to enjoy both effectiveness and higher scalability with different missing ratios. Besides, some probabilistic CP decomposition methods with Bayesian inference are also proposed for solving missing-value prediction problem~\cite{xiong2010temporal}. These methods treat $\mathcal{D}$ as the log-likelihood function and delete the terms of missing observations from likelihood functions to handle missing values and conduct imputation. To sum up, CP-based completion models, which leverage MS approach, are usually more robust compared with the models with EM-like approach when the missing ratio is large but could be hard to be optimized with existing projection-based algorithms.

%could be taken as the probabilistic format of CP decomposition method, and most of them 
%To solve these problems, a second model INDAFAC (INcomplete DAta paraFAC) was proposed based on the second idea by ignoring the missing entries for model construction. 

\subsubsection{Tucker-Based Methods}

Tucker decomposition is proposed by Tucker~\cite{tucker1966some} and further developed by Kroonenberg et al.~\cite{kroonenberg1980principal} and De Lathauwer et al.~\cite{de2000multilinear}. Given an $N^{\text{th}}$-order tensor  $\pmb{\mathscr{X}}$, its Tucker decomposition is defined as an approximation of a core tensor $\pmb{\mathscr{C}}\in \mathbb{R}^{\Pi_{n=1}^N R_n}$ multiplied by $N$ (orthogonal) factor matrices $\mathbf{A}_n \in \mathbb{R}^{I_n\times R_n}, n=1,\ldots, N$ along each mode, such that,

\begin{equation}
\begin{aligned}
\pmb{\mathscr{X}} \approx  \pmb{\mathscr{C}} \times_1  \mathbf{A}_1 \times_2  \mathbf{A}_2 \times_3  \cdots  \times_N \mathbf{A}_N=  \llbracket \pmb{\mathscr{C}} ; \mathbf{A}_1, \ldots, \mathbf{A}_N \rrbracket,
\label{equ:Tucker}
\end{aligned}
\end{equation}
where $R_n$ is a positive integer denoting an upper bound of the rank of $\pmb{\mathscr{X}}$. We can further unfold $\pmb{\mathscr{X}}$ along its $n^{\text{th}}$ mode as follows,
\begin{equation}
\begin{aligned}
\pmb{\mathscr{X}} \quad \overset{\text{unfold}}{\Longrightarrow} \quad \mathbf{X}_{(n)} &  \approx  \mathbf{A}_{n}\mathbf{C}_{(n)}(\mathbf{A}_{n-1} \otimes  \ldots \mathbf{A}_{1} \otimes \mathbf{A}_{N}\ldots \otimes \mathbf{A}_{n+1} )^\top. \\ %\xrightarrow{\hspace*{0.6cm}}
\end{aligned}
\end{equation}

Tucker decomposition is also a widely used tool for tensor completion. Similar to CP-based methods, EM-like and MS approaches are still two traditional ways of handling missing values as well as data imputation. Early works by Walczak et al.~\cite{walczak2001dealing} and Andersson et al.~\cite{andersson1998improving} have mentioned the way of using EM-like approach for handling Tucker decomposition with missing values. This method was combined with the higher-order orthogonal iteration (HOOI) algorithm~\cite{de2000best} to deal with missing values in pattern-recognition problems in~\cite{geng2009facial,geng2011face}. Some other researchers utilized MS approach such as Karatzoglou et al.~\cite{karatzoglou2010multiverse} and Chu et al.~\cite{chu2009probabilistic}. The former applies high-order SVD (HOSVD) algorithm to track low-rank Tucker subspace for complete multidimensional arrays and use the SGD algorithm for optimization with only observed values. The later one considers a probabilistic approach called ``pTucker" in which the missing part is naturally deleted in the likelihood functions.

\subsubsection{Tucker-Based Methods vs CP-Based Methods}
From the former discussion, we can see that both Tucker and CP-based methods could achieve the completion purpose with similar imputation approaches (EM-like approach or MS approach). However, two aspects may lead to different applications of different approaches. (1) \textbf{Effectiveness:} Tucker-based methods are shown to be usually more effective than CP-based methods~\cite{karatzoglou2010multiverse, Papalexakis:Tensor} under the same rank assumptions and ideal hyperparameter selections (i.e., the perfect number of latent factors based on the rank assumptions) because their core tensor is more general in capturing complex interactions among components that are not strictly trilinear. Thus, if one only concerns about the completion accuracy without caring about the uniqueness or interpretability of the decomposed latent factors, Tucker-based methods could always be a good choice. (2) \textbf{Uniqueness and Interpretability:} Comparing with Tucker decomposition, a good property of the CP decomposition is that it is often unique with the exception of elementary indeterminacies of permutation and scaling~\cite{Kolda:Tensor}. This property could also facilitate easier interpretation of factorized latent matrices with the help of domain knowledge as imposed constraints~\cite{wang2015rubik,carroll1980candelinc}. Besides, CP-based methods are usually more computationally flexible to deal with large-scale datasets in a distributed way since CP decomposition do not introduce complex core tensor as Tucker decomposition. %which is more easy to be distributed.

Recently, several methods based on hierarchical tensor representations~\cite{rauhut2015tensor,da2013hierarchical,rauhut2017low} provide a flexible generalization of classical Tucker model to deal with high order tensors. Most of these approaches are optimized using projected gradient methods such as iterative hard thresholding algorithms~\cite{rauhut2015tensor,rauhut2017low}. The key idea is to use Riemannian gradient iteration method (RGI) which contains two step in each iteration: (1) to perform a gradient step in the ambient space (2) map the result back to the low-rank tensor manifold with hierarchical singular value thresholding procedure. Another variant projected gradient method called Riemannian optimization approach~\cite{da2013hierarchical,kressner2014low,kasai2015riemannian} is also considered based on the hierarchical representations for manifold construction. For its particularity, we leave it later in Section~\ref{sec:genother} for a more detailed introduction. 

%Besides aforementioned two types of decomposition based methods, several other decomposition based methods can also be modified based on EM-like approach or MS approach to deal with missing data or for completion purpose.

%{\red This is not a realistic assumption in practice. The approach based on Tucker factorization has been used for comparison in papers Liu et al. (2013); Gandy et al. (2011) and Tomioka et al. (2011). As shown there, it is very sensitive to the rank estimation. Namely, in Gandy et al. (2011) it was demonstrated that tensor completion using the Tucker factorization fails (or doesn’t reach the desired error tolerance) if mode-ranks are not set to their true values.   requires some approximations of n-ranks of a tensor， the resulting problem is non-convex and there are no theoretical guarantees that the globally optimal solution will be found.

% the rank function is nonconvex and discrete,\cite{filipovic2015tucker}}

%The probabilistic approach of 
%~\cite{shan2011probabilistic}(only 8 cites...)

\subsection{Trace Norm Based Approaches}\label{sec:trace}

Matrix trace norm (or nuclear norm) has been popularized as a convex surrogate of non-convex rank function for solving the matrix completion problem. For matrices with bounded operator norm, it has been shown to be the tightest lower bound of matrix rank function among all possible convex approximation~\cite{recht2010guaranteed}. Generalized from this matrix relaxation, Liu et al.~\cite{liu2009tensor} first defined the tensor trace norm as the combination of the trace norms of its unfoldings and reformulated the tensor rank minimization problem as a convex optimization problem. Signoretto et al.~\cite{signoretto2010nuclear} further generalized the tensor trace norm to be a particular case of Shatten-p,q norm. In fact, a more direct generalization from matrix trace norm to tensor trace norm is to utilize the definition of tensor $n$-rank. By substituting the tensor rank with a linear combination of tensor $n$-rank, it could could be further relaxed with trace norms for defining a low-$n$-rank tensor pursuit problem~\cite{gandy2011tensor, liu2013tensor}:
\begin{equation}
\begin{aligned}
\centering
& \underset{ \pmb{\mathscr{X}} }{\text{minimize}}
&&\text{rank}_{\ast}(\pmb{\mathscr{X}})  \overset{\text{substitute}}{\Longrightarrow}  \sum_{n=1}^N \alpha_n rank(\mathbf{X}_{(n)})   \overset{\text{relax}}{\Longrightarrow}  \sum_{n=1}^N \alpha_n\, {\displaystyle \|   \mathbf{X}_{(n)}   \|_{\ast}}  \\
& \text{subject to}
&& \pmb{\mathscr{X}}_{\pmb{\Omega}}= \pmb{\mathscr{T}}_{\pmb{\Omega}} + \pmb{\mathscr{E}}_{\pmb{\Omega}},
\end{aligned}
\label{equ:relax1}
\end{equation}
where $\sum_{n=1}^N \alpha_n$ is usually defined as 1 to maintain the consistency with the matrix trace norm~\cite{liu2013tensor}. We call this model SNN (sum of the nuclear norm) model. Though these unfolding matrices cannot be optimized independently because of their multi-linear correlations, this convex relaxation allows us to solve the completion problem without predefining the tensor rank, which is more tractable in practice. An equivalent problem in the case of Gaussian noise could be formulated as: 
\begin{equation}
\begin{aligned}
\centering
& \underset{ \pmb{\mathscr{X}} }{\text{minimize}}
&& \frac{\lambda}{2} \| \mathcal{P}_{\pmb{\Omega}} (\pmb{\mathscr{X}} - \pmb{\mathscr{T}} ) \|_\text{F}^2+\sum_{n=1}^N \alpha_n\, {\displaystyle \|   \mathbf{X}_{(n)}   \|_{\ast}},  \\
\end{aligned}
\label{equ:relax2}
\end{equation}
where $\lambda>0$ is a trade-off constant. To solve problem~\eqref{equ:relax2},  Liu et al.~\cite{liu2009tensor} propose a simple algorithm optimized by the block coordinate decent\footnote{Though the formulation solved in~\cite{liu2009tensor} is kind of different from~\eqref{equ:relax}, but they are the same in essence.}. This algorithm is the first trace-norm based tensor completion algorithm, and to our best knowledge, it is the first tensor completion algorithm without fixing the rank of the tensor in advance as decomposition-based approaches do. A simplified version of it and two advanced algorithms were proposed in their subsequent journal paper~\cite{liu2013tensor}. A more popular approach for solving the above problem is to apply splitting method~\cite{signoretto2010nuclear,tomioka2010estimation,gandy2011tensor, liu2013tensor}, e.g., Alternating Direction Method of Multipliers (ADMM) which is a special case of  Douglas-Rachford splitting method\footnote{ Douglas-Rachford splitting method has also been shown to be a special case of point proximal methods, which is useful for accommodating various non-smooth constraints. Details are discussed in \cite{eckstein1992douglas}.}. As there are only slight changes in these methods, we first consider the noise exist case and introduce the integrated approach of Grandy et al.~\cite{gandy2011tensor} and Tomioka et al.~\cite{tomioka2010estimation}, and then come back to the noiseless case discussed in~\cite{signoretto2010nuclear,gandy2011tensor,liu2013tensor}.

Under the paradigm of ADMM, the completion problem defined in Equation~\eqref{equ:relax2} could be reformulated by introducing auxiliary matrices $\{\mathbf{Z}_{(n)}\}$, $n=1,\ldots, N$, as:
\begin{equation}
\begin{aligned}
\centering
& \underset{ \pmb{\mathscr{X}}, \mathbf{Z}_1,\ldots,\mathbf{Z}_N }{\text{minimize}}
&& \frac{\lambda}{2} \| \mathcal{P}_{\pmb{\Omega}} (\pmb{\mathscr{X}} - \pmb{\mathscr{T}} ) \|_\text{F}^2+\sum_{n=1}^N \alpha_n\, {\displaystyle \|   \mathbf{Z}_{n}   \|_{\ast}}  \\
& \text{subject to} \,\,
&&\mathbf{Z}_{(n)}= \mathbf{X}_{n}, n=1,\ldots,N.
\end{aligned}
\label{equ:relax3}
\end{equation}

A corresponding augmented Lagrangian objective function could be derived as:
\begin{equation}
\begin{aligned}
\centering
& \mathcal{L}_{\eta}=
\frac{\lambda}{2} \| \mathcal{P}_{\pmb{\Omega}} (\pmb{\mathscr{X}} - \pmb{\mathscr{T}} ) \|_\text{F}^2+\sum_{n=1}^N  \left(  \alpha_n\, {\displaystyle \|   \mathbf{Z}_n \|_{\ast}} +  <\mathbf{Y}_n, \mathbf{Z}_n- \mathbf{X}_{(n)}> + \frac{\eta}{2}  \lVert  \mathbf{Z}_n - \mathbf{X}_{(n)}  \rVert^2_\text{F} \right).  \\
\end{aligned}
\label{equ:Aug}
\end{equation}
where $\{\mathbf{Y}_n\}_n$ are the Lagrange multipliers, $\eta > 0$ is a penalty parameter. Following the standard updating scheme of ADMM, $\{\mathbf{Z}_{n}\}_n$ and $\{\mathbf{Y}_n\}_n$ could be iteratively updated as follows:
\begin{equation}
\begin{aligned}
&\mathbf{Z}_n \leftarrow SVT_\frac{\alpha_n}{\eta}(\mathbf{A}_n-\frac{ \mathbf{Y}_n}{\eta}), \\
&\mathbf{Y}_n\leftarrow \mathbf{Y}_n+\eta(\mathbf{Z}_n-\mathbf{A}_n), ~~ n=1,2,\ldots, N,
\label{equ:Zupdate}
\end{aligned}
\end{equation}
where $SVT_\frac{\alpha_n}{\eta}$ is the singular value thresholding operator \cite{cai2010singular} which is a shrinkage operator defined as $SVT_{\delta}(\mathbf{A}) = \mathbf{U}(diag\{\sigma_i - \delta\})_{+}\mathbf{V}^\top$, where $\mathbf{U}(diag\{\sigma_i \}_{1\le i \le r}) \mathbf{V}^\top$ represents the singular value decomposition of the matrix $\mathbf{A}$. For any matrix $\mathbf{X}$, $\mathbf{X}_{+}=\max\{\mathbf{X},0\}$, where $\max\{\cdot,\cdot\}$ denotes an element-wise operator. The update of $\pmb{\mathscr{X}}$ could have several choices, one can use the EM-like approach as we describe above~\cite{liu2013tensor} or use the exact~\cite{gandy2011tensor,tomioka2010estimation} or inexact approaches~\cite{gandy2011tensor}. The exact updating approach is describe in Algorithm 1.
To deal with the noiseless case, Signoretto et al.~\cite{signoretto2010nuclear} and Gandy et al.~\cite{gandy2011tensor} propose to decrease the value $\lambda$ in Equation~\eqref{equ:relax} while Liu et al.~\cite{liu2013tensor} directly delete the first term and prove their algorithm to be more efficient. 

%%Here is a table for comparison among different trace-norm based approaches.???
%\setlength{\textfloatsep}{1pt}
%\begin{algorithm}[H]
%\caption{ADMM for noise exist trace-norm based tensor completion}
%\begin{algorithmic}[1]
%% \REQUIRE{$\pmb{\mathscr{T}}$, $\mathbf{\Omega}$, $\{\alpha_n\}_{n=1}^N$, $\lambda$, $\eta$, $\rho$, $tol$}
%% \ENSURE{$\pmb{\mathscr{X}}$ }
%%Initialize $\mathbf{Z}_n=\mathbf{Y}_n={\bf 0}$\;
%
%%\WHILE{Certain Stop Criterion is not Satisfied}   
%%%$\eta=\min\{\eta*\rho,\eta_{\max}\}$   (accelerate optimization process)\;
%%\For{$n=1:N$}
%%Update $\mathbf{Z}_n$ using Eq.~\eqref{equ:Zupdate}\;
%%\EndFor
%%Update $\pmb{\mathscr{Y}}$ using Eq.~\eqref{equ:Zupdate}\;
%%Update $\pmb{\mathscr{X}}$ with EM-like Approach\;
%%\EndWhile
%\renewcommand{\algorithmicrequire}{\textbf{Input:}}
% \renewcommand{\algorithmicensure}{\textbf{Output:}}
% \REQUIRE $\pmb{\mathscr{T}}$, $\mathbf{\Omega}$, $\{\alpha_n\}_{n=1}^N$, $\lambda$, $\eta$, $\rho$, $tol$
% \ENSURE  $\pmb{\mathscr{X}}$
% \\ \textit{Initialization} : $\mathbf{Z}_n=\mathbf{Y}_n={\bf 0}$\\
%%  \STATE first statement
%% \\ \textit{LOOP Process}
%\WHILE{Certain Stop Criterion is not Satisfied}
%  \FOR {$n = 1$ to $N$}
%  \STATE Update $\mathbf{Z}_n$ using Eq.~\eqref{equ:Zupdate}
%%  \IF {($i \ne 0$)}
%%  \STATE statement..
%%  \ENDIF
%  \ENDFOR
% \STATE Update $\pmb{\mathscr{Y}}$ using Eq.~\eqref{equ:Zupdate}
% \STATE Update $\pmb{\mathscr{X}}$ with EM-like Approach
% \ENDWHILE
% \RETURN $\pmb{\mathscr{X}}$ 
% \label{alg:ADMM}
% \end{algorithmic}
%\end{algorithm}

\setlength{\textfloatsep}{1pt}
\begin{algorithm}[t!]
 \KwIn{$\pmb{\mathscr{T}}$, $\mathbf{\Omega}$, $\{\alpha_n\}_{n=1}^N$, $\lambda$, $\eta$, $\rho$, $tol$}
 \KwOut{$\pmb{\mathscr{X}}$ }
Initialize $\mathbf{Z}_n=\mathbf{Y}_n={\bf 0}$\;
\Repeat{Certain Stop Criterion is Satisfied}{                  %
%$\eta=\min\{\eta*\rho,\eta_{\max}\}$   (accelerate optimization process)\;
\For{$n =1:N$}{
Update $\mathbf{Z}_n$ using Eq.~\eqref{equ:Zupdate}\;
}
Update $\pmb{\mathscr{Y}}$ using Eq.~\eqref{equ:Zupdate}\;
Update $\pmb{\mathscr{X}}$ with EM-like Approach\;
}
%\Return $\pmb{\mathscr{X}}$ 
\caption{ADMM for noise exist trace-norm based tensor completion}
 \label{alg:ADMM}
\end{algorithm}

Beyond these traditional approaches focusing on minimizing the sum of nuclear-norm, several variants are proposed, which could be categorized into two types. One is to introduce variant trace norm structures, and another is to combine with decomposition based approaches by transferring the trace-norm constraints on factorized matrices. 

SNN model is neither the tightest relaxation of tensor rank nor an optimal solution considering the sample size requirement. To reduce the sample size required for completion, the first type of methods is mainly focusing on proposing better convex relaxation based on new trace norm structure. Mu et al.~\cite{mu2014square} suggest to use a single powerful regularizer rather than a combination structure as SNN model, i.e., apply trace norm on a more balanced unfolding matrix rather than using the summation of trace norm. Besides numerical experiments, the theoretical analysis of sampling bound is also conducted by exploiting the sharp properties of the Gaussian measurement. Different from matrix trace-norm structure, some other trace-norm based structures are also proposed such as tensor-nuclear-norm~~\cite{zhang2014novel,kilmer2013third} and incoherent trace norm~\cite{yuan2016incoherent}. They either target new rank structure (tensor tubal-rank) or consider the incoherence condition. As most of them are focusing on the analyzing the sampling size requirement based on different sampling approaches, we will stress their theoretical findings in the statistical analysis section.

The second type of approaches is to add trace norm constraints on factorized matrices rather than on unfolding matrices, which could be treated as mixed approaches of trace-norm based methods and decomposition based methods. These methods are mainly focusing on reducing the computation complexity since trace norm minimization methods are often costly and require SVD decomposition of potentially large matrices~\cite{kressner2014low}. Also, they are more flexible in incorporating different structures~\cite{ying2017hankel} on decomposition matrices and alleviate the requirement of the predefined ranks for decomposition based methods~\cite{liu2014factor}. Ying et al.~\cite{ying2017hankel} focus on the exponential signal completion problem. As a signal model only has $\mathcal{O}(R)$ degree of freedoms \footnote{A signal model expresses each entry as $x_{i_1,\dots.i_N}=\sum_{r=1}^R(d_r \prod_{n=1}^N z_{n,r}^{i_n-1})$. $R$ is the CP rank.}, they explore to exploit an exponential structure of the factor vectors to reduce the number of samples for exact recovery. Each factor is first transformed into a Hankel matrix to promote exponential structure, and then nuclear norms are applied to force these Hankel matrices to be low-rank. 

To address the computational complexity problem, some researchers utilize QR decomposition~\cite{liu2013efficient} while some researchers add trace-norm constraints on CP or Tucker factor matrices~\cite{liu2014factor}. Their approaches both apply the trace norm onto the factorized matrices rather than the original unfolding matrices for acceleration while the latter one also alleviates the rank pre-definition problem faced by decomposition-based approaches to some extent. Another interesting and straightforward variation of trace norm approach is to transform matrix trace norm into Frobenius norm. In~\cite{srebro2005rank}, for a matrix $\mathbf{X}$ with low-rank decomposition $\mathbf{X}=\mathbf{U}\mathbf{V}^\top$, an equivalent expression of its trace norm could be defined as:  
\begin{equation}\label{equ:sep}
{\displaystyle \|   \mathbf{X} \|_{\ast}} := min_{\mathbf{U},\mathbf{V}} \frac{1}{2} \{ \|{\mathbf U}\|_{\text{F}}^2 +  \|{\mathbf V}\|_{\text{F}}^2 \}.
\end{equation}

In a similar spirit to the matrix case, Mardani et al.~\cite{Mardani:Tensor} further extends it into the tensor level and apply it in the online situation which we will introduce in Section~\ref{sec:dy}. Other approaches, which focus on relaxing nuclear norm to new convex relaxations in order to reduce both the computational complexity and the number of sample sizes could also be be found~\cite{rauhut2015theta}.

%{\blue TENSOR THETA NORMS AND LOW RANK RECOVERY
%In this article, we introduce convex relaxations of the tensor nuclear norm which are computable in polynomial time via semidefinite programming.
%}

%The key ingredient of our approach is to define a new class of tensor nuclear norms that explicitly account for the incoherence of the linear subspaces spanned by the fibers of a tensor in defining its nuclear norm. 

%also point out the lack of SNN model but from a different perspective. 
%In a similar spirit to Mu et al.~\cite{mu2014square} f

%some other researchers directly use matrix based trace norm~\cite{??,??} 

%{\red \cite{[25] B. Recht, M. Fazel, and P. A. Parrilo, “Guaranteed minimum-rank solutions of linear matrix equations via nuclear norm
%minimization,” SIAM Rev., vol. 52, no. 3, pp. 471–501, 2010.
%[26] B. Recht and C. Re, “Parallel stochastic gradient algorithms for large-scale matrix completion,” 2011 (submitted).}}

\subsection{Other Variants}\label{sec:genother}
%Various branches of tensor completion problems are studied in existing works.  In this part, by no means to integrate all of them, we introduce several approaches. 
To provide a more comprehensive and informative introduction, we include several important variants in this section as a supplementary for general tensor completion methods.

\subsubsection{Non-negative Constrained Approaches}
Real-world applications sometimes require non-negative constraints such as image completion~\cite{xu2013block}, medical data analysis~\cite{ho2014limestone,wang2015rubik} and so on. To impute missing entries of non-negative tensors, a popular way is to add non-negative constraints on latent factor matrices based on decomposition models~\cite{cichocki2009nonnegative}. Different optimization method could be applied such as Block Coordinate Descent~\cite{xu2013block} and ADMM framework~\cite{wang2015rubik} with an iteratively non-negative threshold. For non-negative tensors with integer entries, Takeuchi and Ueda~\cite{takeuchi2016graph} use generalized Kullback-Leibler (gKL) divergence by analogy with several matrix-based approaches~\cite{cai2011graph, dhillon2005generalized,xu2012alternating}. 

%Wang et al.~\cite{wang2015rubik} add non-negative constraints on the factor matrices of the CP decomposition to enhance interpretability in health data analytics. They apply the ADMM framework and the perform a non-negative threshold iteratively on the auxiliary matrices. 

\subsubsection{Robust Tensor Completion}\label{subsubsec:rob} Robust data analysis plays an instrumental role in dealing with outliers, gross corruptions (non-Gaussian noises) in completion problem~\cite{goldfarb2014robust}. 
Drawing upon the advances in robust PCA analysis~\cite{candes2011robust}, robust tensor completion is to complete a tensor $\pmb{\mathscr{T}}$ by separating it into a low-rank part $\pmb{\mathscr{X}}$ plus a sparse part $\pmb{\mathscr{E}}$ to capture the different noise patterns. The objective function could be formulated as follows:
\begin{equation}
\begin{aligned}
\centering
& \underset{ \pmb{\mathscr{X}},\pmb{\mathscr{E}} }{\text{minimize}}
&&   \text{rank}_{\ast}(\pmb{\mathscr{X}})+ \|  \pmb{\mathscr{E}} \|_0 \\
& \text{subject to}
&& \pmb{\mathscr{X}}_{\pmb{\Omega}}+\pmb{\mathscr{E}}_{\pmb{\Omega}}= \pmb{\mathscr{T}}_{\pmb{\Omega}}.
\end{aligned}
\label{equ:RTC}
\end{equation}
Similar trace-norm strategy could be added to the low-rank part for a relaxation of the rank minimization and $l_1$ norm is often used as a convex surrogate to relax the $l_0$ norm. The completion problem could be well addressed by the joint minimization of trace norm and $l_1$ norm. Li et al.~\cite{li2010optimum} utilize a block coordinate decent method similar with~\cite{liu2009tensor} and shows promising results in separating corrupted noise and image restoration. Goldfarb and Qin~\cite{goldfarb2014robust} study the problem of robust low-rank tensor completion in a convex ADMM optimization framework and provide both theoretical and practical analysis of convex and non-convex models. Recent work of Jain~\cite{jain2017noisy} also adopts ADMM framework to complete noisy tensors with one of the CP factors being sparse. Besides ADMM framework, Zhao et al.~\cite{zhao2014robust} consider the probabilistic framework with a fully Bayesian treatment optimized by variational Bayesian inference approach. Javed et al.~\cite{javed2015stochastic} provide an online stochastic optimization method for robust low-rank and sparse error separation in addressing background subtraction problem.

\begin{table*}[t!]\scriptsize
\centering
%\begin{tabular}{|c|c|c|*{2}c|c|c|c|}\hline
%\multirow{2}{*}{Rank Assumption} & \multirow{2}{*}{ Name }& \multicolumn{2}{c|}{ Core Approach} & \multirow{2}{*}{ Regular. } & \multirow{2}{*}{ Optimization Method} & \multirow{2}{*}{ Recovery} & \multirow{2}{*}{ Time Complexity} \\
%&&Factorization& Trace Norm &&&&& \\\hline\hline
\begin{tabular}{|c|c|c|c|c|}\hline
Rank Assumption & Name & Core Objective & Optimization Method & Imputation   \\\hline\hline

\multirow{5}{*}{CP}  & ALS-SI~\cite{tomasi2005parafac,bro1998multi} & CPD  & ALS & EM-like   \\\cline{2-5}

 & INDAFAC~\cite{tomasi2005parafac} & CPD & Levenberg-Marquadt &  MS   \\\cline{2-5}

& CP-WOPT~\cite{acar2011scalable} &  CPD & nonlinear conjugate gradient &  MS    \\\cline{2-5}

& BPTF~\cite{xiong2010temporal} & Probablistic CPD & Bayesian (MAP) & MS    \\\cline{2-5}
 % & \cite{papalexakis2013k} &   & $\frac{\lambda}{2} \sum \vert A \vert$ &  F \& $l_1$  & ALS \\\cline{2-6}

%  & n-NTF\cite{shashua2005non} &  & $\frac{\lambda}{2} \sum \lVert A \rVert_{2}^{2}$  &F \& Non-Neg  & ALS \\\cline{2-6}

& STC~\cite{krishnamurthy2013low} & CPD+Adaptive Sampling & RLS & MS \\\cline{2-5}

   & Tensor L.S.~\cite{bhojanapalli2015new} & CPD+Adaptive Sampling  &ALS & MS \\\cline{2-5}

 & \cite{jain2014provable} & CPD  & ALS & MS  \\\hline

 & TNCP~\cite{liu2015trace} & SNN$+$CPD & ADMM  & EM-like   \\\cline{2-5}

\multirow{15}{*}{Tucker}

% hayashi2010exponential 
% & Tucker3\cite{kroonenberg1980principal}\cite{kapteyn1986approach}&   &  &F &ALS \\\cline{2-6}
% Walczak et al.~\cite{andersson1998improving} and Andersson et al.~\cite{walczak2001dealing}
 
 & ALS/IA (Tucker3/IA)~\cite{walczak2001dealing,andersson1998improving}& TD (HOOI)  & ALS & EM-like  \\\cline{2-5}

 & pTucker~\cite{chu2009probabilistic}& Probablistic TD  & EM/Bayesian (MAP) & EM/MS  \\\cline{2-5}
 & MRTF~\cite{karatzoglou2010multiverse}&  TD (HOSVD) & SGD & MS  \\\cline{2-5}
  & \cite{filipovic2015tucker} & TD  & NCGM~\cite{acar2011scalable}  &MS   \\\cline{2-5}
% & HOOI\cite{de2000best}&   & &  &  \\\cline{2-6}
& geomCG~\cite{kressner2014low} &  Riemannian Opt.  & Riemannian NCG & MS  \\\cline{2-5}
& Riemannian Preconditioning~\cite{kasai2015riemannian} & Riemannian Opt.  & Riemannian Preconditioned NCG& MS  \\\cline{2-5}
%Nonlinear Conjugate Gradient

 & SiLRTC~\cite{liu2009tensor, liu2013tensor} & SNN &Block CD   &EM-like \\\cline{2-5}
  & FaLRTC~\cite{liu2013tensor} & SNN &Smoothing Scheme Scheme  &EM-like  \\\cline{2-5}
 & HaLRTC~\cite{liu2013tensor} & SNN & ADMM   &EM-like \\\cline{2-5}

 & \cite{tomioka2010estimation}& SNN & ADMM &EM-like   \\\cline{2-5}
  & (A/E)-CMLE~\cite{signoretto2010nuclear}& SNN & ADMM &EM-like   \\\cline{2-5}
& ADM-TR (E)~\cite{gandy2011tensor} & SNN &ADMM  &EM-like  \\\cline{2-5}
& Square Deal~\cite{mu2014square} & Single NN & matrix ALM~\cite{lin2010augmented} &EM-like  \\\cline{2-5}

& \cite{romera2013new} & Alternative Relaxation & ADMM &EM-like  \\\cline{2-5}
   & \cite{yuan2016incoherent} & Incoherent Tensor Norms  & Gradient based  & MS \\\hline

 Hierarchical & HTTC~\cite{da2013hierarchical} & Hierarchical TD & Steepest Descent/CG & MS \\\cline{2-5}
Tucker & TIHT~\cite{rauhut2015tensor,rauhut2017low}  & Hierarchical TD &  Iterative Hard Thresholding & MS   \\\hline

Tubal-rank & t-SVD~\cite{zhang2014novel} & Tensor-nuclear-norm & ADMM &  EM-like \\\hline

% {\red	 & STC~\cite{krishnamurthy2013low}  & Adaptive Sampling &  Partial Incoherent  &    $\mathcal{O}(Ir^{\frac{5}{2}} N^{2} log(r))$          \\ 
  
%   &Parallel WALS \cite{bhojanapalli2015new}   & Adaptive Sampling  &   orthogonal CPD, Symmetric  &     $\mathcal{O}( (\sum_{i=1}^I \| (\mathbf{U}^\ast)^i \|^\frac{3}{2}  )^2  Ir^3\kappa^4 log^2(I))$             } \\ \hline      
			
 \end{tabular}
\caption{\label{table:basic}Comparison between general tensor completion algorithms. CPD: CP decomposition. TD: Tucker Decomposition. SNN: Sum of nuclear norms. MAP: Maximum a posteriori estimation.}
\vspace{-0pt}
\end{table*}

%{\blue
%\subsubsection{Sparse Tensor Completion}
%Bayesian sparse Tucker models for dimension reduction and tensor completion
%%\subsubsection{Tensor Completion for Binary Values}
%}

\subsubsection{Riemannian Optimization}
Riemannian optimization technique has gained increasing popularity in recent years~\cite{da2013hierarchical}. The idea of it lies in an alternative treatment focusing on the fixed-rank tensor completion problem. By embedding the rank constraint into the search space, an unconstrained problem on a smooth manifold $\mathcal{M}_\mathbf{r}=\{\pmb{\mathscr{X}}\in  \mathbb{R}^{I_1\times I_2 \times \ldots \times I_N}  |\text{rank}_\text{tc}(\pmb{\mathscr{X}})=\mathbf{r}=(r_1,\ldots,r_N)\}$ is defined as follows:
\begin{equation}
\begin{aligned}
\centering
& \underset{ \pmb{\mathscr{X}}\in \mathcal{M}_\mathbf{r} }{\text{minimize}}
&& \frac{1}{2} \| \mathcal{P}_{\pmb{\Omega}} (\pmb{\mathscr{X}} - \pmb{\mathscr{T}} ) \|_\text{F}^2.
\end{aligned}
\label{equ:riemannian}
\end{equation}
Riemannian optimization is an iteratively gradient projection approach based on a two-step procedure: the projected gradient step and a retraction step. A smooth manifold and its tangent space, a proper definition of Riemannian metric for gradient projection and the retraction map are three essential settings for Riemannian optimization.  As the Tucker decomposition gives an efficient representation of tensors in $\mathcal{M}_\mathbf{r}$, the factorized orthogonal matrices belonging to Stiefel manifold of matrices~\footnote{Stiefel manifold of matrices is defined as $\{{\bf W} \in \mathbb{R}^{n\times p}| {\bf W}^\top {\bf W} = {\bf I}_p  \},~n\ge p$.~\cite{nishimori2005learning} } are usually used for the parametrization of manifold $\mathcal{M}_\mathbf{r}$ and its tangent space~\cite{kressner2014low,kasai2015riemannian}. Besides the Tucker manifold $\mathcal{M}_\textbf{r}$, other choices of the smooth manifold are also used such as hierarchical Tucker space for handling high-dimensional applications~\cite{da2013hierarchical}. After defining the manifold and its tangent space, the most important ingredients in Riemannian approach is to set the Riemannian metric and gradient. Kressner~\cite{kressner2014low} focused on the search space and exploited the differential geometry of the rank constraint based on Tucker decomposition. Kasai et al.\cite{kasai2015riemannian} laid emphasis on the cost function $\| \mathcal{P}_{\pmb{\Omega}} (\pmb{\mathscr{X}} - \pmb{\mathscr{T}} ) \|_\text{F}^2$ by introducing the block diagonal approximation of its Hessian and defined a novel metric on the tangent space of $\mathcal{M}_\mathbf{r}$ and its quotient manifold. They further generalized a preconditioned conjugate gradient algorithm proposed in~\cite{mishra2014r3mc} with a total computational cost $\mathcal{O}(|\Omega|r_1r_2r_3)$ for a three-order tensor. Finally, the retraction map is used map the updated tensor back to the low-rank tensor manifold. Maps such as HOSVD method~\cite{kressner2014low} and hierarchical singular value thresholding~\cite{rauhut2015tensor} which satisfy the necessary properties of a retraction mentioned in~\cite{kressner2014low} could be used.

There are still various of interesting methods for solving the completion problem such as alternative convex relaxation approach for Tucker ranks~\cite{romera2013new}, adaptive sampling methods~\cite{krishnamurthy2013low, bhojanapalli2015new} or various matrix-based methods~\cite{ji2016tensor}. As it is unpractical to cover all the bases, readers interested in these approaches are encouraged to explore them in the original paper. Finally, we list several representative algorithms we mentioned in Table~\ref{table:basic} for summing up and comparison.

% tensor-SVD~\cite{zhang2014novel,kilmer2013third} and 

% (1): $\sum_{i=1}^3 \max_{\lVert A\rVert \le 1} (\alpha_i<\mathcal{X},A>-\frac{\mu_i}{2}\lVert A \rVert_F^2)$
% (2): $\mathcal{X}=\llbracket \mathcal{G};A,B,C\rrbracket$
%{\red other
%tensor-SVD~\cite{zhang2014novel}
%A New Convex Relaxation for Tensor Completion~\cite{romera2013new}
%\subsubsection{Matrix based approach}
%,  all matrix completion method could be used on tensor or its unfolding matrices, we
%tensor completion via a multi-linear low-n-rank factorization model~\cite{tan2014tensor}
%comparison? \cite{Tensor versus matrix completion: a comparison with application to spectral data}
%\cite{liu2013efficient} An efficient matrix factorization method for tensor completion, QR decomposition

\subsection{Statistical Assumption and Theoretical Analysis}\label{sec:stat}

%{\blue
%The noiseless case is also considered by some researchers, e.g.,~\cite{liu2013tensor}. 

%\noindent \textbf{Q (3): In what condition, we could expect to achieve a unique and exact completion?} 
%In the matrix case, 

%If the number of measurements is sufficiently large, and if the entries are sufficiently uniformly distributed as above, one might hope that there is only one low-rank matrix with these entries. 

%Noisy tensor completion via the sum-of-squares hierarchy~\cite{barak2016noisy}

In this section, we introduce some primary statistical assumptions and theoretical analysis of the completion feasibility. The goal is to elaborate more theoretical foundations of previously mentioned methods before delving into some advanced tensor completion techniques in big data analytics. These assumptions also provide the answers to the question we leave in Section~\ref{subsec:TCP}, i.e., In what condition, we could expect to achieve a unique and exact completion?\footnote{In the matrix completion problem described in~\cite{candes2012exact}, this question is answered as: ``If the number of measurements is sufficiently large, and if the entries are sufficiently uniformly distributed as above, one might hope that there is only one low-rank matrix with these entries.''} It should be noted that, in general, one cannot expect to fully recover every low-rank tensor from a sample of its entries. For example, similar to the matrix situation described in~\cite{candes2012exact}, if a tensor has only one non-zero entry, it is not guaranteed to recover it even we are able to randomly select $90\%$ of its entries, since it is with $10\%$ probability to obtain only zero samples. Thus, a more appropriate question for the completion feasibility should be: In what condition, could we expect to achieve a unique and exact completion with high probability?

%{\blue The intuition of this assumption comes from the fact that: suppose a large tensor has only one non-zero element, it cannot be recovered can not be recovered as described in the matrix case in~\cite{candes2012exact}.} 

\subsubsection{Sampling Assumption}
Tensor completion is usually based on random sampling assumption in which we assume the partially observed entries are uniformly random sampled from the original tensor. In matrix case, Bernoulli sampling~\cite{candes2012exact} or independent sampling with replacement~\cite{recht2011simpler} are always assumed to simply the random sampling approach. Generalizing these assumptions such as Bernoulli sampling to the tensor case is straightforward~\cite{huang2014provable}. Several other sampling assumptions are also used for either the convenience of theoretical demonstration or the practicability in real-world applications, e.g., Gaussian measurements~\cite{mu2014square} , Fourier measurements~\cite{rauhut2017low} and Adaptive Sampling~\cite{krishnamurthy2013low, bhojanapalli2015new}. Gaussian random measurement is a Gaussian linear map $\mathcal{G}: \mathbb{R}^{I_1\times \ldots \times I_N} \rightarrow \mathbb{R}^m$ defined by $m$ tensors $\pmb{\mathscr{G}}_i \in \mathbb{R}^{I_1\times \ldots \times I_N}$ via $[\mathcal{G}(\pmb{\mathscr{X}})](i)=<\pmb{\mathscr{G}}_i, \pmb{\mathscr{X}}>$, where each tensor $\pmb{\mathscr{G}}_i$ has i.i.d. standard normal entries. It enjoys sharp theoretical tools~\cite{mu2014square} and has been well-studied in signal processing field~\cite{tropp2007signal} especially in the compressed sensing area~\cite{donoho2006compressed}. Another widely studied sampling measurement is random Fourier measurement constructed by Fourier transformation: 
\begin{equation}
\mathcal{G}: \mathbb{R}^{I_1\times \ldots \times I_N} \rightarrow \mathbb{R}^m, ~~\mathcal{G}(\pmb{\mathscr{X}})=\frac{1}{\sqrt{m}}\mathcal{P}_\Omega \mathcal{F}_N(\pmb{\mathscr{X}}),
\end{equation}
where $\mathcal{P}_\Omega$ is the random subsampling operator, $\mathcal{F}_N$ is the N-dimensional Fourier transformation, which is formally defined as: 
\begin{equation}
\mathcal{F}_N(\pmb{\mathscr{X}})(j_1,\ldots,j_N)=\sum_{k_1=1}^{I_1}\cdots\sum_{k_N=1}^{I_N}\pmb{\mathscr{X}}(k_1,\ldots,k_N)e^{-2\pi i \sum_{n=1}^N \frac{k_nj_n}{I_n} }.
\end{equation}

Comparing with subgaussian measurements which usually serve as the benchmark guarantees of required observations in exact low-rank recovery, Fourier measurement offers advantages on its more structured sampling property and available fast multiplication routines~\cite{rauhut2017low}.

In some real-world applications, sampling populations are sparse but clustered. Under these circumstances, non-adaptive sampling methods may not be able to achieve exact recovery using a similar size of samples as usual. Instead, varies adaptive sampling approaches~\cite{krishnamurthy2013low, bhojanapalli2015new} are derived to relax the incoherence assumptions which is another important assumption introduced next.

\subsubsection{Incoherence Assumption}
Tensor incoherence assumption is a generalized conception of matrix incoherence arisen in compressed sensing. It is closely correlated with passive uniformly samplings where each entry should have a comparable amount of information to guarantee the recovery tractable. The opposite situation is ``coherent'' which means most of the mass of a tensor concentrate in a small number of elements and these informative elements has to be completely sampled. It may lead to a missing in mass when sampling methods are uniformly at random and independent of the underlying distribution of the tensor. Mathematically, the coherence of an r-dimensional linear subspace $\mathbf{U} $ of $\mathbb{R}^N$ is defined as:
\begin{equation}
\mu(\mathbf{U})=\frac{N}{r} \max_{1\le i \le N} \| \mathbf{P}_{\mathbf{U}} \mathbf{e}_i \|^2_{2} 
= \frac{  \max_{1\le i \le N}  \| \mathbf{P}_{\mathbf{U}} \mathbf{e}_i \|^2_{2}    }{    N^{-1} \sum_{i=1}^N  \| \mathbf{P}_{\mathbf{U}} \mathbf{e}_i \|^2_{2}    },
\end{equation}
where $\mathbf{P}_{\mathbf{U}}$ is the orthogonal projection onto subspace $\mathbf{U}$ and $\mathbf{e}_i$'s are the canonical basis for $\mathbb{R}^N$. Based on this definition, the tensor $\mu$-incoherent is defined as:
\begin{equation}
\mu_i(\pmb{\mathscr{X}}):= \mu(\mathcal{L}_i(\pmb{\mathscr{X}}))=\mu(\mathbf{X}_{(i)}) \le \mu,~~(i=1,\ldots, N),
\end{equation}
where $\mathcal{L}_i(\pmb{\mathscr{X}})$ is the linear subspace spanned by the mode-i fibers. In~\cite{huang2014provable}, besides this N-mode incoherence, the tensor incoherence conditions are further extended to a new ``mutual incoherence" based on the singular value decompositions of the matricizations $\mathbf{X}_{(n)}$'s. Based on different sampling strategies and the incoherence assumptions, now we discuss the theoretical bounds of the observed entries we need to recover the tensors using different algorithms.

\subsubsection{Number of Observed Entries}
To guarantee a unique reconstruction of the original tensor, a theoretic lower bound of the number of observed entries is often required. In a matrix case,  a given incoherent $m\times n$ matrix could be recovered with high probability if the sample size is larger than $Cnr\dot polylog(n)$ for some constant $C$, where $r$ is the matrix rank~\cite{gross2011recovering}. Generalizing this result to the high-order situation is not straightforward~\cite{mu2014square}. 
Tomioka et al.~\cite{tomioka2011statistical} first conducted the statistical analysis of low-rank tensor recovery. They provided the first reliable recovery bound for the SNN (sum of nuclear norms) model~\eqref{equ:relax1} under the assumption of Gaussian measurement. In their analysis, for a given $N^{\mbox{th}}$-order tensor $\pmb{\mathscr{X}}\in \mathbb{R}^{I\times \ldots \times I}$ of Tucker rank $(r,r,\ldots,r)$, $\pmb{\mathscr{X}}$ could be completed with high probability if the number of observed entries is larger than $CrI^{N-1}$, where $C$ is a constant. This corollary is further proved by Mu et al.~\cite{mu2014square} to be a necessary condition for SNN model and the probability is shown to be $1-4exp(-\frac{(\kappa-m-2)^2}{16(\kappa-2)})$, where $\kappa=min_n\{  {\displaystyle \|   \mathbf{X}_{(n)}   \|_{\ast}} /  \|  \pmb{ \mathscr{X} } \|_{\text{F}}  \} \times I^{N-1}$. 

However, this necessary bound for SNN model is far from satisfactory because one only needs no more than $r^N+rIN$ parameters to describe the tensor $\pmb{\mathscr{X}}$ defined above based on the Tucker decomposition in Equation~\eqref{equ:Tucker}. Furthermore, Mu et al. prove that $(2r)^N+2rIN+1$ measurements are sufficient to complete $\pmb{ \mathscr{X}}$ almost surely based on the non-convex model formulated in Equation~\eqref{equ:TCP3}. In fact, the freedom of a tensor $\pmb{\mathscr{X}} \in \mathbf{R}^{I_1\times \ldots \times I_N}$ with Tucker rank $(r_1, \ldots, r_N)$ is only $\prod_{n=1}^N r_n+ \sum_{n=1}^N (r_n I_n -r_n^2)$~\cite{kressner2014low}. This result can be calculated based on an easy two-step construction: (1) The number of entries of the core tensor is randomly valued which demonstrate its freedom is $\prod_{n=1}^N r_n$. (2) To construct the rest part, we can iteratively unfold the core tensor along each mode to get the mode-$n$ unfolding matrix of size $r_n\times C$, where $C=I_1\times \ldots \times I_{n-1} \times r_{n+1}\times \ldots \times r_N$ (n=1,\ldots,N). Each of this matrix represents $r_n$ uncorrelated vectors which could be used to construct the rest $I_n-r_n$ vectors with freedom $r_n(I_n-r_n)$.  Thus, the total freedom is $\prod_{n=1}^N r_n+ \sum_{n=1}^N (r_n I_n -r_n^2)$. This analysis motivates researchers to find lower bounds as well as advanced methods for tensor completion. In~\cite{mu2014square}, the authors improve the SNN model into a better convexification. Rather than using a combination of all mode-$n$ trace norms, they suggest using an individual trace norm of a more balanced matrix defined by $\pmb{\mathscr{X}}_{[j]} = reshape(\mathbf{X}_{(1)}, \prod_{n=1}^j I_n, \prod_{n=j+1}^N I_n)$, where $j\in [N] := \{1,\ldots, N \}$.  Under this square deal setting, $CrI^{[\frac{N}{2}]}$ entires are sufficient for recovery when $rank_{CP} \pmb{\mathscr{X}}=r$, where $\pmb{\mathscr{X}} \in \mathbf{R}^{I_1\times \ldots \times I_N}$. Only $Cr^{[\frac{N}{2}]}I^{[\frac{N}{2}]}$ observations are needed to recover a tensor $\pmb{\mathscr{X}}$ of Tucker rank $(r,\ldots,r)$ with hight probability.
%The fact of using single regularization rather th is early discovered by Oymak et al.~\cite{oymak2015simultaneously}. 
  
Jain et al.~\cite{jain2014provable} stress the setting of random sampling (rather than Gaussian random sampling in~\cite{mu2014square}) and consider that the square deal approach requires each observation to be a ``dense random projection" rather than a single observed entry. They generate a matrix alternating minimization technique to the tensor level with a CP decomposition approach and prove that a symmetric third order tensor $\pmb{\mathscr{X}}\in \mathscr{R}^{I\times I\times I}$ with CP rank-$r$ could be correctly reconstructed from $\mathcal{O}(I^{\frac{3}{2}}r^5 log^4 I)$ randomly sampled entries. In a similar spirit, Barak and Moitra~\cite{barak2015tensor} give an algorithm based on sum-of-squares relaxation for prediction of incoherent tensors. Instead of exact recovery, their main result shows that $\mathcal{O}(I^{\frac{3}{2}}r^2 log^4 I)$ observations could guarantee an approximation with an explicit upper bound on error. However, this sum-of-squares approach is point out not to be scale well to large tensors and is substituted by a spectral approach proposed by Montanari et al.~\cite{montanari2016spectral} with matching statistical guarantee (i.e., the required sample size). Recently, Yuan and Zhang~\cite{yuan2016incoherent} explore the correlations between tensor norms and different coherence conditions. They prove that an $N^{\text{th}}$-order tensor $\pmb{\mathscr{X}}\in \mathbb{R}^{I\times \ldots \times I}$ with CP rank-$r$ could be completed entirely from $\mathcal{O}((r^{\frac{N-1}{2}}I^{\frac{3}{2}}+r^{N-1}I)(log(I))^2)$ uniformly sampled entries through a proper incoherent nuclear norm minimization.

\begin{table*}\tiny
 \centering
% \setlength{\tabcolsep}{6.3pt}
%\vspace{-3pt}
%\begin{small}
    \begin{tabular}{|c|c|c|c|c|}  \hline 
    % \multicolumn{2}{c||}{Tensor Type}      &    \multicolumn{6}{c|}{General Multi-Aspect Streaming Tensor}      &    \multicolumn{6}{c}{Temporal Multi-Aspect Streaming Tensor}           \\  \cline{1-14}   
Rank Assumption & Method  &   Sampling Method & Incoherent and other conditions & Requirement for exact recovery (w.h.p) \\  \hline     \hline                                                
 \multirow{4}{*}{$\text{rank}_\text{cp}=r$} & Provable TF~\cite{jain2014provable}  & Random Sampling & Incoherent, orth. CPD, Symmetric  &$\mathcal{O}(r^5I^{\frac{3}{2}} log^4(I))$      \\                                                     
                                                 
 & ITN~\cite{yuan2016incoherent}  & Uniformly Random Sample &  Incoherent  &     $\mathcal{O}((rI^{\frac{3}{2}}+r^{2}I)log^2(I))$           \\ 
 
  & NTC~\cite{barak2016noisy}  &  Uniformly Random Sample  &  Incoherent  &     $\mathcal{O}((rI^{\frac{3}{2}})log^4(I))$           \\

  & STC~\cite{krishnamurthy2013low}  & Adaptive Sampling &  Partial Incoherent  &    $\mathcal{O}(Ir^{\frac{5}{2}} log(r))$          \\ 
  
   &Parallel WALS~\cite{bhojanapalli2015new}   & Adaptive Sampling  &   orthogonal CPD, Symmetric  &     $\mathcal{O}( (\sum_{i=1}^I \| (\mathbf{U}^\ast)^i \|^\frac{3}{2}  )^2  Ir^3\kappa^4 log^2(I))$          \\

    &M-norm~\cite{ghadermarzy2017near}  & Uniformly Random Sample  & Incoherent   &   $O(r^{6} I)$       \\ \hline

%                    Near-optimal sample complexity for convex tensor completion
%``our recent work on tensor completion using M-norm and max-qnorm  which needs $O(r^{6} I)$ measurements for approximate recovery (using notation of Table 3 in your paper).''

 \multirow{2}{*}{$\text{rank}_\text{tc}=(r,r,r)$} & SNN~\cite{tomioka2011statistical,mu2014square} & Gaussian measurements &    N.A.      &  $\mathcal{O}(rI^2) $        \\
                                   
 & Square Deal~\cite{mu2014square}      & Gaussian measurements &    N.A.      & $\mathcal{O}(r^{[\frac{3}{2}]}I^{[\frac{3}{2}]})$      \\

  & GoG~\cite{xia2017polynomial}      & Uniformly Random Sample  &   Incoherent     & $\mathcal{O}(r^{\frac{7}{2}}I^{\frac{3}{2}}log^{\frac{7}{2}} (I) +r^{7}Ilog^6(I) )$    \\\hline

tubal-rank r & t-SVD~\cite{zhang2017exact}   &  Uniformly Random Sample  & Incoherent &$\mathcal{O}(rI^2 log(I))$       \\  \hline

    \end{tabular}
     \vspace{10pt}
    \caption{Statistical assumptions \& bounds (third-order tensor of size $I\times I \times I$). $\text{rank}_\text{tc}$ means Tucker rank.}
\label{table:Bounds}
 \vspace{-0pt}
\end{table*}

Recently, the theme of adaptive sensing emerges as an efficient alternative to random sampling approach in large-scale data obtaining and processing. By exploiting adaptivity to identify highly informative entries, the required number of observed entries could be substantially reduced. Krishnamurthy et al.~\cite{krishnamurthy2013low} propose an adaptive sampling method with an estimation algorithm which could provably complete an $N^{\mbox{th}}$-order rank-$r$ tensor with $\mathcal{O}(Ir^{N-\frac{1}{2}} N^{2} log(r))$ entries. Subsequently, Bhojanapalli and Sanghavi~\cite{bhojanapalli2015new} give a new sampling approach based on a parallel weighted alternating least square algorithm. They show that a symmetric rank-$r$ third order tensor $\pmb{\mathscr{X}}=\sum_{i=1}^r \sigma_i^\ast \mathbf{U}_i^\ast \otimes  \mathbf{U}_i^\ast \otimes  \mathbf{U}_i^\ast $, where $\mathbf{U}^\ast \in \mathbb{R}^{I\times r}$ is an orthonormal matrix, could be proved to be exactly recovered from $\mathcal{O}( (\sum_{i=1}^I \| (\mathbf{U}^\ast)^i \|^\frac{3}{2}  )^2  Ir^3\kappa^4 log^2(I))$, where $(\mathbf{U}^\ast)^i$ are n rows of $\mathbf{U}^\ast$, $\kappa$ is a restricted condition number. Although both algorithm crucially relies on the adaptive sampling technique which does not generalize to random samples, they achieve the completion purpose with little or no incoherence assumptions on the underlying tensor. 
In the end, we summarize the sufficient sampling lower bounds of several existing approaches for exact recovery as well as their sampling methods and incoherent assumptions in Table~\ref{table:Bounds}.
%To make the completion tractable, the is often assumed leading to 
%Provable Tensor Factorization with Missing Data~\cite{jain2014provable} {\red related work }
%A new sampling technique for tensors~\cite{bhojanapalli2015new}
%%%%%%%~\cite{huang2014provable}
%~\cite{signoretto2014learning}
%~\cite{yuan2016tensor}
%~\cite{yuan2016incoherent}
%\subsubsection{Time Complexity?}

%Tensor completion methods usually require low-rank assumption and uniformly sampling. 

\section{Tensor Completion with Auxiliary Information}\label{sec:aux}

In the previous section, tensor completion methods are introduced along with some key statistical assumptions according to the completion feasibility. Though the techniques varied, both experimental results and theoretical analysis reflect a natural and intuitive phenomenon that: with the increasing ratio of missing entries, the prediction accuracy tends to be significantly decreased. In real-world data-driven applications, besides the target tensor object, additional side information such as spatial and temporal similarities among objects or auxiliary coupled matrices/tensors may also exist~\cite{wang2014low, Ge:Tensor,ge2016taper, Narita:Tensor, xiong2018field}. These heterogeneous data sources are usually bonded and compensate with each other and could serve as potential supplements to improve the completion quality especially when the missing ratio of the target tensor is high. In this section, we provide an overview of related approaches and mainly introduce two ways of incorporating variety auxiliary information in existing literatures of tensor completion -- similarity based approaches and coupled matrices/tensors factorization. 

\subsection{Similarity Based Approaches}
Inspired by relation regularized matrix factorization proposed by Li et al.~\cite{li2009relation}, Narita et al.~\cite{Narita:Tensor} use two regularization methods called ``within-mode regularization" and ``cross-mode regularization" respectively to incorporate auxiliary similarity among modes for tensor factorization. These methods are all based on EM-like approach combined with Tucker or CP decomposition. The key idea is to construct within-mode or cross-mode similarity matrices and incorporate them as following regularization terms:
\begin{equation}
\begin{aligned}
\centering
& R_\text{within}(\pmb{\mathscr{X}};\mathbf{A_1},\ldots,\mathbf{A_N}) =\sum_{n=1}^N \sum_{i,j=1}^{I_n} \mathbf{S}_n(i,j) \| \mathbf{A}_n(i,:) - \mathbf{A}_n(j,:) \|_\text{F}^2  =\sum_{n=1}^N \text{tr}(\mathbf{A}_n^\top \mathbf{L}_n \mathbf{A}_n) \\
& R_\text{cross} (\pmb{\mathscr{X}};\mathbf{A_1},\ldots,\mathbf{A_N}) = \text{tr}((\mathbf{A}_1 \otimes \ldots \otimes \mathbf{A}_N  )^\top \mathbf{L} (\mathbf{A}_1 \otimes \ldots \otimes \mathbf{A}_N) ),
\end{aligned}
\label{equ:relax}
\end{equation}
where $\mathbf{A}_n$ is the mode-$n$ latent matrix of CP or Tucker decomposition, $\mathbf{S}_n$ is the mode-$n$ auxiliary similarity matrix of size $I_n \times I_n$. $\mathbf{L}_n$ is the Laplacian matrix of $\mathbf{S}_n$ and $\mathbf{L}$ is defined as the Laplacian matrix of $\mathbf{S}_1 \otimes \ldots \otimes \mathbf{S}_N$ which is also called Kronecker product similarity in the early work of Kashima et al.~\cite{kashima2009link}. By combining them with standard decomposition approaches and applying to tensor completion problem, the auxiliary information was shown to improve completion accuracy especially when observations are sparse. In fact, an early work~\cite{zheng2010collaborative} has taken advantage of the ``within-mode regularization" and combine it with the coupled matrix method for the mobile recommendation problem. Similar idea of these two approaches have also been used in~\cite{Ge:Tensor,bahadori2014fast} and~\cite{Chen:Tensor} respectively. 

\begin{figure}
\centering
\vspace{-0cm}
\includegraphics[height=4cm, width=12.5cm]{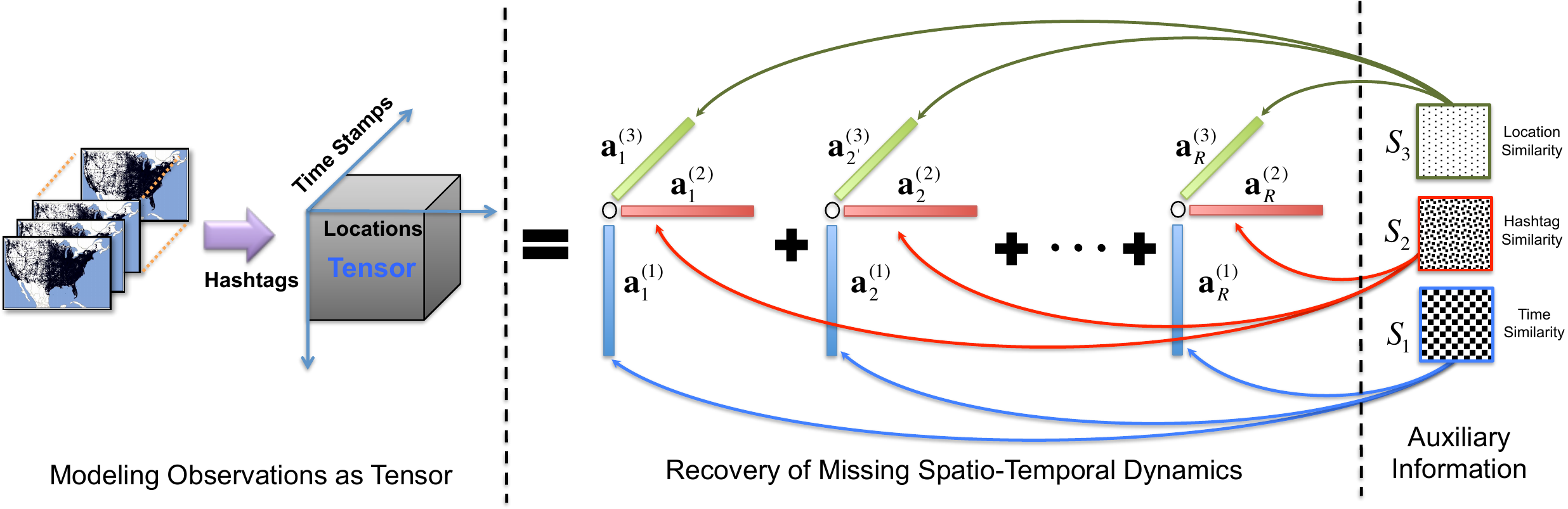} %\vspace{-.4cm} 
\vspace{-0.2cm} 
\caption{Illustration of the AirCP Framework.~\cite{Ge:Tensor}}
\label{fig:aircp}  
\vspace{0.2cm} 
\end{figure}

%(i.e., the number of twitter hashtags)

To effectively incorporate auxiliary information, the main concentration of these similarity-based methods is to define the within-mode or cross-mode similarity matrices. As an example, we briefly introduce one state-of-the-art model leveraging the within-mode auxiliary similarity for tensor completion. Figure~\ref{fig:aircp} illustrates the general framework of the AirCP (Auxiliary Information Regularized CANDECOMP/PARAFAC) model proposed by Ge et al.~\cite{Ge:Tensor} to impute the missing online memes by incorporating their auxiliary spatio-temporal information. By modeling the observed data as a third-order tensor (left), the authors seek to recover the missing entries with a CP-based method (center) by integrating auxiliary information like the relationships between locations, memes and times (right). Three similarity matrices $\{\mathbf{S}_i \}_{i=1,2,3}$ are defined for three modes adopting the geographical distance, temporal relationships or the fusions of multiple similarity measures. For example, the spatial relationships ($\textbf{S}_2$) and the temporal relationships ($\textbf{S}_3$) are modeled as the fusion of two similarity measures and the tri-diagonal matrix respectively as follows:

\begin{minipage}{.57\textwidth}
\vspace{4pt}
\begin{equation}
\begin{aligned}
	\textbf{S}_2(i,j) =  \tau \bm{\textbf{S}_{GD}}(i, j), 
	    + (1-\tau)\bm{\textbf{S}_{AS}}(i, j) \\ 
	    (1\le i,j \le I_2),
\end{aligned}
\end{equation}
\vspace{4pt}
\end{minipage}
\begin{minipage}{.36\textwidth}
\vspace{4pt}
\begin{equation}
	\textbf{S}_3 = 
	\begin{bmatrix}
		0 & 1 & 0 & \dots\\
		1 & 0 & 1 & \dots\\
		0 & 1 & 0 & \dots\\
		\vdots & \vdots & \vdots & \ddots\\
	\end{bmatrix},
\end{equation}
\vspace{0pt}
\end{minipage}

\noindent where $\textbf{S}_{GD}$ is the geographical distance matrix defined as $\textbf{S}_{GD}(l_i,l_j) = exp\left({-\frac{Dist(l_i, l_j)^2}{2\alpha^2}}\right)$ in which $\alpha$ is a predefined dispersion constant and $Dist(l_i, l_j)$ is the distance between location $l_i$ and location $l_j$ based on their GPS coordinates; $\textbf{S}_{AS}$ is a heuristics similarity called adoption similarity~\cite{Ge:Tensor}, which we do not specify it here for the ease of presentation; $\tau$ is a balancing hyperparameter. Similar to these similarity matrices defined in AirCP, there are also other ways to define the similarity matrices, which are usually guided by different domain knowledge varying from applications to applications~\cite{bahadori2014fast,Chen:Tensor}. From the perspective of probabilistic approaches~\cite{bazerque2013rank, lamba2016incorporating,zhao2015bayesian,zhao2015bayesianTucker}, these similarity matrices among each mode could all be treated as the inverse of kernel matrices in Gaussian prior distributions defined for each factor matrices.

 %As a natual extension of matrix factorization with auxiliary regularization proposed by
%Lack: predefined similarity matrix and fixed rank as we described

\subsection{Coupled Matrices/Tensors Factorization}

\begin{figure}
\centering
\vspace{-0cm}
\includegraphics[height=3.3cm, width=9cm]{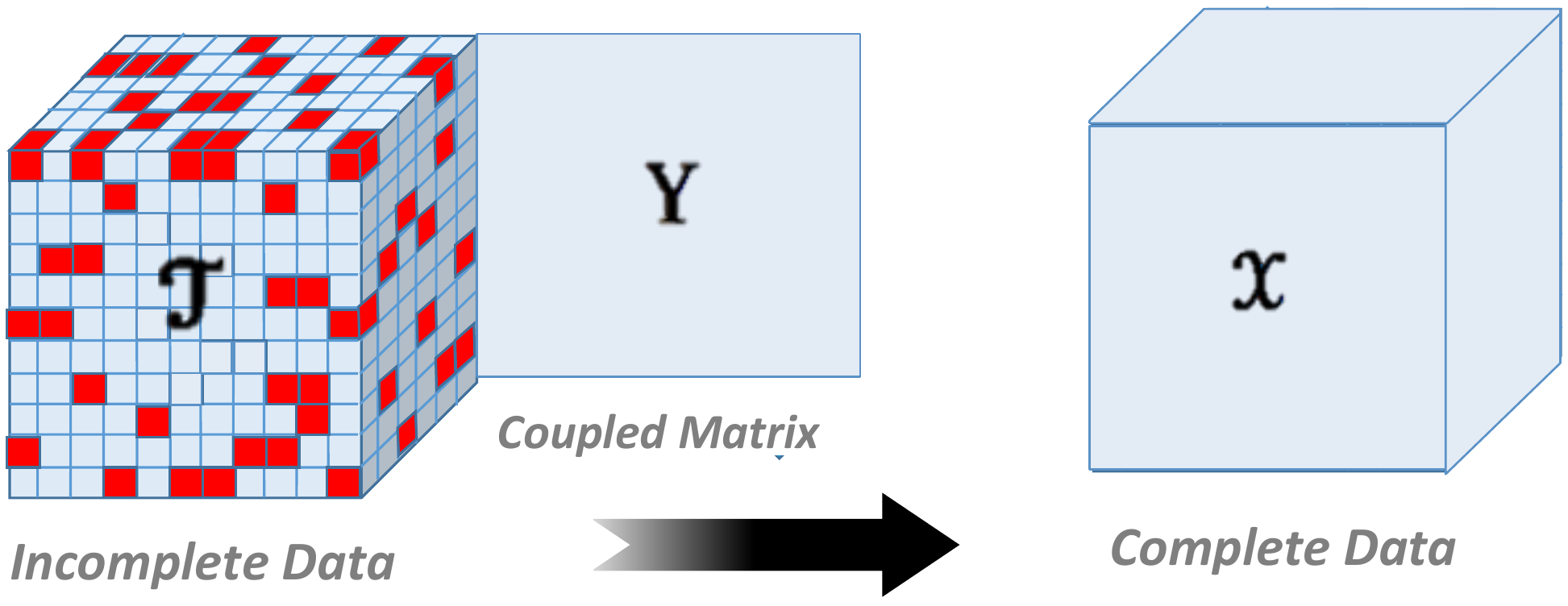} %\vspace{-.4cm} 
\vspace{-0.4cm} 
\caption{Illustration of the Coupled Matrices/Tensors Factorization for Tensor Completion.~\cite{acar2011all}}
\label{fig:couple}  
\vspace{0.3cm} 
\end{figure}

Another popular way of incorporating auxiliary information is to couple tensors or matrices together for jointly factorization and imputation. Deriving from coupled matrix factorization~\cite{smilde2000multiway,singh2008relational}, the way of coupled side information has also been widely developed in multi-way data analysis~\cite{banerjee2007multi,lin2009metafac, acar2011all, zheng2010collaborative, wang2014travel, beutel2014flexifact, ermics2015link}. One of the most popular methods applied to tensor completion is coupled matrix and tensor factorization (CMTF) proposed by Acar et al.~\cite{acar2011all}. Based on the idea of sharing the factorization matrices among matrices and tensors constructed from heterogeneous datasets, they propose an all-at-once algorithm to impute an incomplete tensor coupled with one or more matrices. Mathematically, it is translated to:
\begin{equation}
\begin{aligned}
\centering
 \underset{ \mathbf{A}_1,\cdots,\mathbf{A}_N, \mathbf{V} }{\text{minimize}}
 \,\, \frac{1}{2} \| \mathcal{P}_{\pmb{\Omega}} (\pmb{\mathscr{X}}- \llbracket  \mathbf{A}_1, \ldots, \mathbf{A}_N  \rrbracket   ) \|_\text{F}^2 + \frac{1}{2} \| \mathbf{Y} - \mathbf{A}_{n} \mathbf{V}^\top  \|_\text{F}^2,
% \quad \text{subject to} \,\,
% \pmb{\Omega} \circledast \pmb{\mathscr{X}}=\pmb{\mathscr{T}},
\end{aligned}
\label{equ:coupled}
\end{equation}
where $\mathbf{Y}$ is the coupled auxiliary matrix for the $n^\text{th}$ mode. An intuitive illustration is depicted in Figure~\ref{fig:couple}. Both the target tensor $\pmb{\mathscr{X}}$ and auxiliary matrix $\textbf{Y}$ have low-rank structure, and they are assumed to share common latent factors in their coupled mode.  Acar et al.~\cite{acar2011all} also show that though CP decomposition achieves comparable recovery accuracy when the missing ratio is lower than $80\%$, its recovery error sharply increased when the error rate is further increased. By contrast, CMTF could keep acquiring low recovery accuracy until the error rate is over $90\%$.
Similar to the CP-based methods~\cite{zheng2010collaborative,beutel2014flexifact}, factor matrices of Tucker decomposition could also be coupled with auxiliary matrices or tensors to benefit the completion tasks~\cite{wang2014travel,ermics2015link, yilmaz2011generalised}. Moreover, recent work by Zhou et al.~\cite{zhou2017tensor} provides a Riemannian manifold approach to incorporate auxiliary coupled matrices for tensor completion. By adding coupled-matrix regularizers\footnote{Similar to the one defined in Equation~\eqref{equ:coupled}.} into the vanilla Riemannian tensor completion model described in Equation~\eqref{equ:riemannian}, the model could be optimized with Riemannian Conjugate Gradient descent method. An interesting property of this model is that: the coupled-matrix regularizers could be equivalently transformed as the same format as the with-in mode regularizer defined in Equation~\eqref{equ:relax}, with the difference being that the Laplacian matrices are substituted by the projection matrices constructed by the coupled auxiliary matrices.

\subsection{Other Approaches}

Besides the above two approaches, an increasing number of research are conducted recently seeking for new ways to exploit various types of auxiliary information. We list some of them here and give a brief introduction.
%(1) \emph{Hybrid approach} of coupled are the most intuitive one to explore. For example, Ge et al.~\cite{ge2016taper} use . 

\subsubsection{Hybrid Models} The hybrid model is one of the most intuitive approaches based on the aforementioned two approaches. For example, Ge et al.~\cite{ge2016taper} propose a tensor-based recommendation framework TAPER to tackle the personalized expert recommendation problem. In addition to the similarity matrices utilized in its groundwork model AirCP, TAPER also incorporates the cross-mode features as coupled matrices and examines the effectiveness of different combinations of the heterogeneous contextual information.

\subsubsection{Coupled Trace Norm}  This method is a recent model by Wimalawarne et al.~\cite{wimalawarne2017convex}. Different from couple factorization models, the authors defined three different coupled trace norms by calculating the trace norm of coupled matrices concatenated with the latent factorized matrices or the matricization of the targeted tensor. For example, for a third-order tensor, the coupled overlapped trace norm for its first mode with the corresponding coupled matrix is defined as:
\begin{equation}
\|\pmb{\mathscr{X}}, \mathbf{Y}\|^1 := \| [\mathbf{X}_{(1)}; \mathbf{Y}]  \|_{\ast} + \sum_{i=2}^3 \| \mathbf{X}_{(i)}  \|_{\ast},
\end{equation}
where $[\mathbf{X}_{(1)}; \mathbf{Y}]$ denotes the concatenation of the mode-1 unfolding of the target tensor and the corresponding coupled matrix.

\subsubsection{Other Constraints}

Since all of the methods above can be treated as adding different types of regularization constraints based on the auxiliary information, in the end, we describe several other constraints leveraging different side information here for better coverage of related advances. In~\cite{yokota2016smooth}, the authors proposed a smooth constrained completion method, which imposes linear constraints on individual components in the CP factorized matrices. It could be treated as an extension to the symmetric similarity matrix in Equation~\ref{equ:relax} into non-symmetric equations using prior knowledge. The method also considers other two types of smoothness constraints, i.e., total variation and quadratic variation, and supports automatic rank search by gradually increasing the number of components during the optimization until the optimal rank is achieved. Rather than putting the concerns on single components, Vervliet et al. consider anther linear constraint in~\cite{vervliet2017canonical}. They assume each factorized matrix ($\mathbf{A}_{n}$) could be further decomposed into a known matrix $\mathbf{B}_{n}$ and an unknown coefficient matrix $\mathbf{C}_{n}$, i.e., $\mathbf{A}_{n}=\mathbf{B}_{n}\mathbf{C}_{n}$. It could also be treated as a ``reversed'' version of the coupled matrices/tensors factorization (since we are not decomposing the auxiliary matrices $\mathbf{B}_{n}$ but decomposing the latent matrices $\mathbf{A}_{n}$) and could be optimized with a novel nonlinear least squares algorithm faster and more accurately comparing with the traditional CP decomposition methods when few entries are observed or when missing entries are structured.

%~propose another linear constraints on the decomposed CP factors. 

%{\blue

%xiong2018field
%\cite{Field-of-Experts Filters Guided Tensor Completion}

%}

%it is closely related to the coupled matrices/tensors factorization methods, which we call it a ``reversed'' version of the since the constraints they are highly correlated with each

%increases the number of components gradually until the optimal rank 

%Simultaneous tensor decomposition and completion using factor priors.

%\textbf{smooth constraints} 
%CANONICAL POLYADIC DECOMPOSITION OF INCOMPLETE TENSORS WITH LINEARLY CONSTRAINED FACTORS∗
%and leave the rest for interested readers to further explore. 

%{Evrim Acar, Go ̈zde Gurdeniz, Morten A. Rasmussen, Daniela Rago, Lars O. Dragsted, and Rasmus Bro. 2012. Coupled matrix factorization with sparse factors to identify potential biomarkers in metabolomics. In IEEE International Conference on Data Mining Workshops (ICDMW). IEEE, 1–8. DOI:http://dx.doi.org/10.1109/icdmw.2012.17}

%{\red Instead of a CP model, we can have CMTF that assumes a Tucker model [Ermis ̧ et al. 2015]:
%min ∥X−G×3 U3 ×2 U2 ×1 U1∥2F +∥Y−U1DT∥2F. (6) U1 ,U2 ,U3 ,G,D
%An important issue when we have two or more heterogeneous pieces of data cou- pled with each other is the one we informally call “drowning,” that is, when one of the datasets is vastly denser or larger than the other(s) and thus dominates the ap- proximation error. In order to alleviate such problems, we have to weigh the different datasets accordingly [Wilderjans et al. 2009].}

\section{Scalable Tensor Completion}
\label{sec:scalable}

%Bro and Andersson~\cite{bro1998multi} proposed the earliest method to accelerate CP decomposition with missing data.
%With the rapid growth of relational data and network data that can be naturally expressed as a high-order tensor, 

With the rapid growth in the volume of datasets with the high-order tensor structure, it is very challenging for tensor completion approaches to handle large-scale datasets with billions of elements in each mode due to the high computational costs and space. In this section, we introduce existing efforts in tackling the scalable tensor completion problem and elaborate these methods with the key challenge they focus on. Since almost all of the current advances are concentrating on CP- and Tucker-based approaches, we split them into a separate section and emphasize them first, and then summarize other related methods with a unified view. It should be noted that although some methods (especially decomposition based methods) neither explicitly conduct the completion tasks nor claim that they are focusing on the completion problem, as they leverage the sparsity property to tackle the scalable challenges, we still include some popular ones here for the sake of completeness.

%{\red As most of the related paper are tensor decomposition based approach }

\subsection{Scalable CP and Tucker-Based Methods}

As described in Section~\ref{sec:general}, decomposition based methods are one of the most widely used types of methods in tensor completion problem. Due to increasingly amount of data volume, they often face with several challenges such as the intermediate data explosion problem~\cite{kang2012gigatensor} where the amount of intermediate data of an operation exceeds the capacity of a single machine or even a cluster, and the data decentralization problem, where the multilinear property, as well as  various types of the regularization terms, affect the scalability and parallelism of the optimization. Notably, these challenges also appear in the scalable tensor decomposition problem, but we mainly target on the methods, which are proposed for the tensor completion problem or sparse tensor decomposition
~\footnote{The sparse tensor decomposition here is different from decomposing a tensor into sparse factors. While former one assumes the original tensor to be sparse, the latter one usually requires the decomposed latent factors to be sparse. }  for three reasons: (1) Our main focus is to impute the missing entries rather than distilling the tensors into possibly interpretable soft clusterings. (2) Different from the dense tensors, which are usually assumed in the tensor decomposition problem, the missing ratios of the incomplete tensor or the sparsity property may highly affect the time and space complexities of the tensor completion algorithms besides the size of the tensors. (3) Although decomposing sparse tensors does not require the process of data imputation, the sparsity property it assumes causes it to share high similarities and challenges with the tensor completion problem. Since the methods in both problems focus on the optimization schemes to address the scalability issues, most of these optimization algorithms could be interchangeably used thus benefit both tasks.

% as while updating latent matrices by ALS approaches

\subsubsection{Intermediate Data Explosion Problem}

Prior research studies have put their main focus on the intermediate data explosion problem occurring in the operations for updating factor matrices in alternating least squares algorithm~\cite{bader2006algorithm, yu2012scalable, beutel2014flexifact, choi2014dfacto, jeon2015haten2, shin2017fully}. With the increase of the size of input data, the size of the intermediate data in the operations could become huge (with hundreds of millions of nonzero elements~\cite{kang2012gigatensor}).

Alternating Least Squares (ALS)~\cite{smilde2005multi} is one of the most popular algorithms to decompose either the fully observed tensor data or the partially observed tensor data into its CP format by conditionally updating each factor matrix given others. However, previous ALS approaches usually have scalability issues. It is inevitable to materialize matrices, calculated by Khatri-Rao Product, that are very large in sizes and cannot fit in memory. For instance, we review the definition of Khatri-Rao product of $\mathbf{A} \in \mathbb{R}^{I \times R}$ and $\mathbf{B} \in \mathbb{R}^{J \times R}$ with the same number of columns denoted as $R$:
\begin{equation}
        \mathbf{A} \odot \mathbf{B} = [\mathbf{A}(:,1) \otimes \mathbf{B}(:,1), \dots, \mathbf{A}(:,R) \otimes \mathbf{B}(:,R)],
\end{equation}
where the size of $\mathbf{A} \odot \mathbf{B}$ is $IJ \times R$. Apparently, if $I=1 \  million$, $J=1 \  million$ and $R=10$, the Khatri-Rao Product $\mathbf{A} \odot \mathbf{B}$ is of size $1 \  trillion \times 10$ which is normally very large and dense, and obviously cannot fit in memory. Therefore, how to solve this intermediate data explosion problem plays an important role in scaling tensor decomposition algorithms. Since the sparsity is a common property of tensors in the completion task, meaning that most elements in the tensor are zeros (or more generally, unobserved), a naive solution to the intermediate data explosion problem is to adopt the coordinate format to only store non-zero elements for tensor completion and further avoid the materialization of huge, unnecessary intermediate Khatri-Rao products~\cite{bader2006algorithm}.

A recent ALS approach for partially observed matrices~\cite{zhou2008large} is extensible and parallelizable for the CP-based tensor completion, which updates each latent matrix $\mathbf{A}_n$ row by row as follows:
\begin{equation}
        \mathbf{A}_n(i,:) = (\mathbf{B}^{(n)}_i + \lambda\bm{I}_R)^{-1}\mathbf{C}^{(n)}_i,
\end{equation}
where $\mathbf{B}^{(n)}_i \in \mathbb{R}^{R \times R}$ is a matrix whose entries are:
\begin{equation}
        \mathbf{B}^{(n)}_i(j_1,j_2) = \sum_{(i_1,i_2,\dots,i_N) \in \mathbf{\Omega}_i^{(n)}}(\prod_{k \neq n}\mathbf{A}_k(i,j_1)\prod_{k \neq n}\mathbf{A}_k(i,j_2)).
\end{equation}
$\mathbf{C}^{(n)}_i \in \mathbb{R}^{R \times 1}$ is a vector whose entries are:
\begin{equation}
        \mathbf{C}^{(n)}_i(j,1) = \sum_{(i_1,i_2,\dots,i_N) \in \mathbf{\Omega}_i^{(n)}}(x_{i_1,i_2,\dots,i_N} \prod_{k \neq n}\mathbf{A}_k(i,j)).
\end{equation}
$\bm{I}_R$ denotes an identity matrix with the size of $R \times R$, $\mathbf{\Omega_i^{(n)}}$ is the set of indices of observed entries in $\pmb{\mathscr{X}}$ whose $n^{\text{th}}$ mode's index is $i$. Therefore, updating each latent matrix can be parallelized by distributing the rows of the latent matrix across machines and performing the update simultaneously without affecting the correctness of ALS. However, since this ALS approach requires that each machine should exchange its latent matrix with others, its communication and memory costs are relatively high with the complexity of $O(R\sum_{n=1}^NI_n)$ per iteration. Hence, each machine has to load all the fixed matrices into its memory as a scalability bottleneck noted in~\cite{gemulla2011large, yu2012scalable}.

\begin{figure}
\centering
\vspace{-0cm}
\includegraphics[height=5cm, width=12cm]{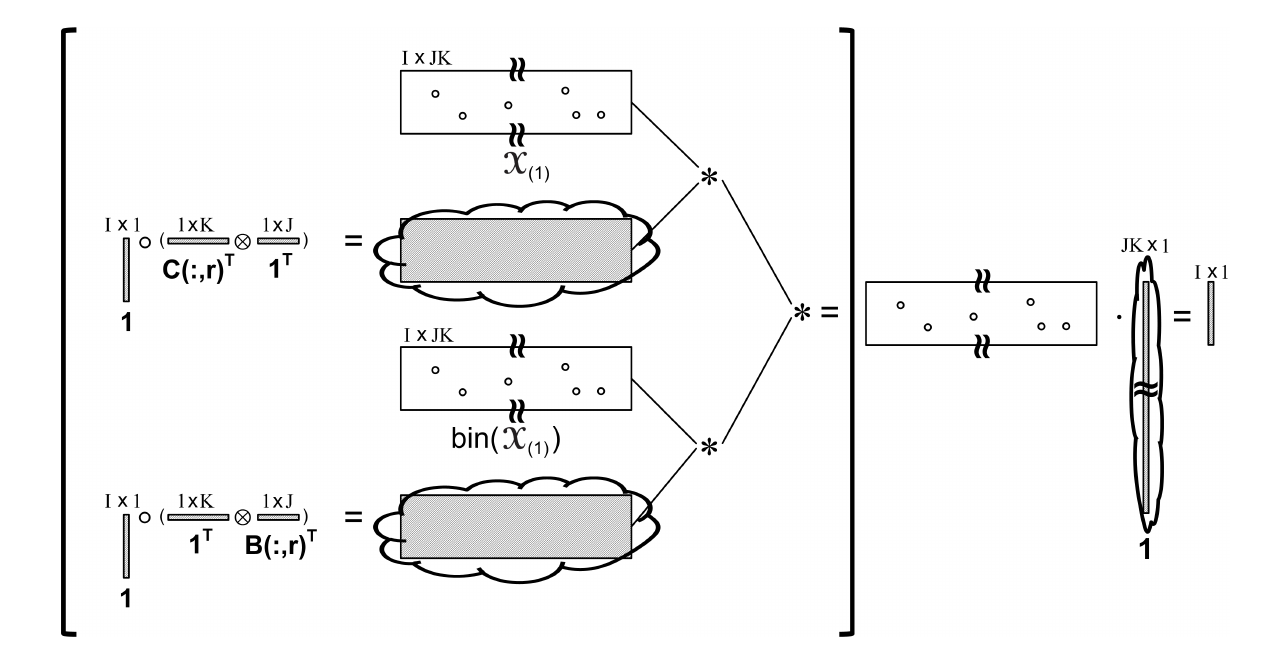} %\vspace{-.4cm} 
\vspace{-0.4cm} 
\caption{Illustration of the key idea of GigaTensor to avoid the intermediate data explosion problem.~\cite{kang2012gigatensor}}
\label{fig:giga}  
\vspace{0.3cm} 
\end{figure}

Similarly, Zinkevich et al.~\cite{zinkevich2010parallelized} extend stochastic gradient descent (SGD) to a distributed version as parallelized stochastic gradient descent (PSGD). Concretely, PSGD randomly splits the observed tensor data into $M$ machines and runs SGD on each machine independently. The updated parameters (latent matrices) are averaged after each iteration. Each element $A_n(i,j)$ in the latent matrix is updated by the following rule:
\begin{equation}
        \mathbf{A}_n(i,j) \leftarrow \mathbf{A}_n(i,j) - 2\eta(\frac{\lambda {\mathbf A}_n(i,j)}{\|\Omega_i^{(n)}\|}-r_{i_1,i_2,\dots,i_N}(\prod_{k \neq n}\mathbf{A}_k(i,j))),
\end{equation}
where $r_{i_1,i_2,\dots,i_N} = x_{i_1,i_2,\dots,i_N} - \sum_{s=1}^R\prod_{k \neq n}\mathbf{A}_k(i,s)$. Apparently, in PSGD, the memory requirements cannot be distributed since all latent parameters $O(R\sum_{n=1}^N I_n)$ need to be exchanged by each machine after each iteration. Also, PSGD has a slow convergence rate.

Another optional way to avoid intermediate data explosion in tensor completion is to reduce the operations that may be too large to fit in memory like the tensor-matrix multiplication. For instance, Kolda and Sun~\cite{kolda2008scalable} work on the Tucker decomposition method for sparse data and solve the intermediate data explosion problem on the target:
\begin{equation}
\pmb{\mathscr{Y}}_n\rightarrow\pmb{\mathscr{X}} \times_1 \mathbf{A}_1 \times_2  \mathbf{A}_2 \times_{n-1} \mathbf{A}_{n-1} \times_{n+1} \mathbf{A}_{n+1}\cdots  \times_N \mathbf{A}_N,
\end{equation}
where $\pmb{\mathscr{Y}}_n$ is the multiplication of the tensor-matrix multiplication for the $n^{\text{th}}$ dimension, i.e., computing the product of a sparse tensor times a series of dense matrices, which may be too large to fit in memory in the process. In order to maximize the computational speed while optimally utilizing the limited memory, the computation is handled in a \emph{piecemeal fashion} by adaptively selecting the order of operations. Concretely, the tensor data is stored with the coordinate format as proposed in~\cite{bader2006algorithm}. Such tensor-matrix multiplication can be calculated one slice or fiber at a time instead of directly multiplying a matrix since the tensor-vector multiplication will produce the results with much smaller size that can easily fit in the limited memory.

Though methods stated above are capable of efficiently decompose and impute an incomplete tensor when the tensor data is very large and sparse, these methods still operate in the limited memory of a single machine. Hence, many researchers utilize the parallel computing system like MPI (Message Passing Interface)~\cite{choi2014dfacto} and the distributed computing system such as MAPREDUCE~\cite{beutel2014flexifact, jeon2015haten2, shin2017fully} to effectively and efficiently perform tensor decomposition and completion for large-scale datasets. In order to address the intermediate data explosion problem in a distributed/parallel fashion, the main ideas include carefully selecting the order of computations in the update of latent matrices for minimizing flops (floating point operations), exploiting the sparsity of the tensor data to avoid the intermediate data explosion, and utilizing the distributed cache multiplication to further minimize the intermediate data by exploiting the skewness in matrix multiplications in a distributed/parallel computing system.

One of the most popular scalable distributed algorithm for PARAFAC sparse tensor decomposition is GigaTensor proposed by Kang et al.~\cite{kang2012gigatensor}, which solves the intermediate data explosion problem by reordering computation and exploiting the sparsity of the tensor data in MAPREDUCE. Specifically, GigaTensor targets the intermediate data explosion problem occurred on updating latent matrices based upon ALS as follows (taking a $3^\text{rd}$-order tensor as an example):
\begin{equation}
        \mathbf{A} \leftarrow \mathbf{X}_{(1)}(\mathbf{C} \odot \mathbf{B})(\mathbf{C}^\top\mathbf{C}*\mathbf{B}^\top\mathbf{B})^\dagger,
\end{equation}
where $\mathbf{X}_{(1)}$ is the unfolded matrix of the tensor $\pmb{\mathscr{X}}$ on the first mode, $\mathbf{A}$, $\mathbf{B}$ and $\mathbf{C}$ are latent facotrized matrices for three dimensions , respectively, and $\dagger$ denotes the pseudo inverse. Inspired by the concept of divide-conquer, the main idea (Figure~\ref{fig:giga}) is to take advantage of the sparsity in tensor data to decouple the two terms in the Khatri-Rao product $\mathbf{C} \odot \mathbf{B}$ and sequentially performing algebraic operations with the unfolded tensor $\mathbf{X}_{(1)}$ instead of directly computing the Khatri-Rao product.
Concretely, computing $\mathbf{X}_{(1)}(\mathbf{C} \odot \mathbf{B})$ can be equally treated as calculating $(\mathbf{N}_1 * \mathbf{N}_2) \cdot \mathbf{1}_{JK}$ where $\mathbf{N}_1=\mathbf{X}_{(1)}*(\mathbf{1}_I \circ (\mathbf{C}(:,r)^\top \otimes \mathbf{1}_J^\top))$, $\mathbf{N}_2=bin(\mathbf{X}_{(1)})*(\mathbf{1}_K^\top \otimes \mathbf{1}_I \circ (\mathbf{B}(:,r)^\top))$, $bin()$ function converts any non-zero value into one, preserving sparsity, and $\mathbf{1}_{JK}$ is an all-$1$ vector of size $JK$. By this way, the size of intermediate data will reduce from $O(JKR+m)$ to $O(max(J+m, K+m))$, where $m$ is the number of non-zero elements. All these operations are implemented in MAPREDUCE by designing mapper and reducer. As a similar fashion, Jeon et al.~\cite{jeon2015haten2} propose HaTen2 that improves on GigaTensor by unifying Tucker and PARAFAC decompositions for sparse tensors into a general framework. Both GigaTensor and HaTen2 can be easily extended to solve the completion problem by adding an indication tensor into the objective function.

Recently, DFacTo, an efficient, scalable and distributed algorithm proposed by Choi et al.~\cite{choi2014dfacto}, also addresses the intermediate data explosion problem by particularly focusing on scaling $\mathbf{\mathcal{F}} = \mathbf{X}_{(1)}(\mathbf{C} \odot \mathbf{B})$. Concretely, DFacTo computes $\mathbf{\mathcal{F}}$ in a column-wise manner as follows:
\begin{equation}
\begin{aligned}
        & \mathcal{H} = \mathbf{X}_{(2)}^\top\mathbf{B}(:,r) \\
        & \mathbf{\mathcal{F}}(:,r) = unvec_{(I,K)}(\mathcal{H})\mathbf{C}(:,r),
\end{aligned}
\end{equation}
where $unvec_{(I,K)}(\cdot)$ is the operator of reshaping a $IK$ dimensional vector into a $I \times K$ matrix. All these operations in DFacTo are very efficient assuming that $\pmb{\mathscr{X}}$ is a sparse tensor represented by the Compressed Sparse Row (CSR) format. DFacTo is implemented by applying a master-slave architecture of MPI (Message Passing Interface). The communication cost of DFacTo is relatively high since the master has to transmits all latent matrices to the slaves at every iteration.

\subsubsection{Decentralization of the Tensor Data Problem}

From a different perspective, some researchers focus on the way of decentralizing the original observed tensor data based on divide-and-conquer rather than the intermediate products. Though it could implicitly solve the intermediate data explosion problem, the decentralization itself could become a new headache since the multilinear properties may prevent the decentralization of the tensor data to be easily done.

$FLEXIFACT$~\cite{beutel2014flexifact} is one of recent approaches targeting on this problem. It splits the observed tensor data into $M^N$ blocks in which each $M$ dis-joint blocks are treated as a stratum without sharing common fibers. $FLEXIFACT$ processes a stratum at a time by decentralizing $M$ dis-joint blocks into $M$ machines and processes them independently and simultaneously. In each machine, $FLEXIFACT$ applies stochastic gradient descent to solve the corresponding block. By this way, the updated parameters in this machine are disjoint with the ones updated by others. Hence, the memory requirements of $FLEXIFACT$ are distributed among $M$ machines. Nevertheless, its communication cost is high since after processing a stratum, $FLEXIFACT$ needs to aggregate all parameters updated by each machine into the master machine that updates them and decentralizes them for the next iteration.

% that solves an optimization problem by iterating over parameters for updating each parameter at a time while all others are fixed. 

\begin{figure}
\centering
\vspace{-0cm}
\includegraphics[height=3cm, width=12cm]{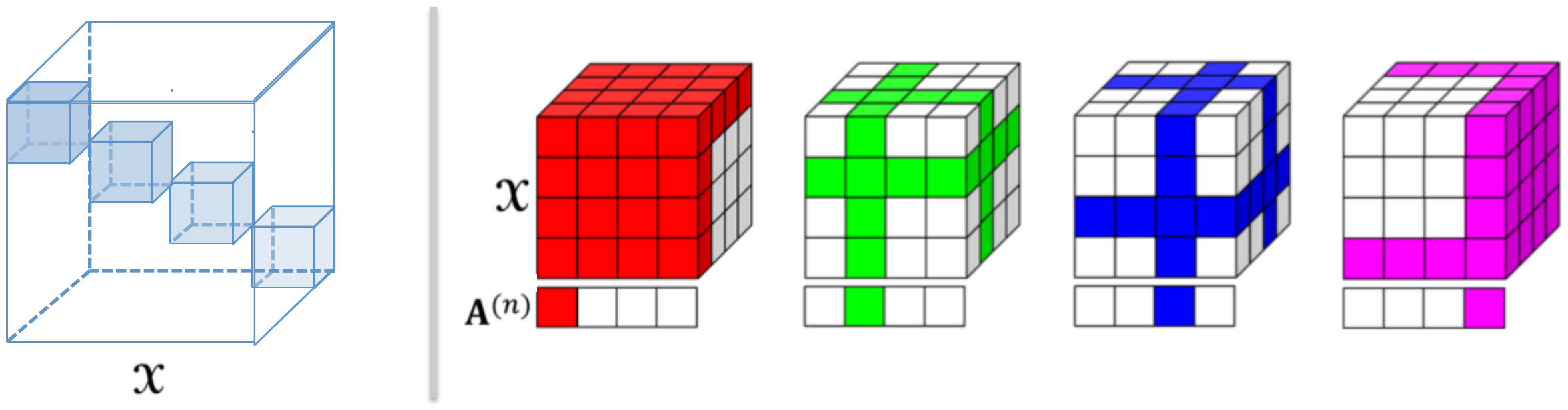} %\vspace{-.4cm} 
\vspace{-0.4cm} 
\caption{Comparison of the decentralization manner between $FLEXIFACT$ (left)~\cite{beutel2014flexifact} and CDTF/SALS (right)~\cite{shin2017fully} on a third-order tensor with four machines. For $FLEXIFACT$, we only show four of the 16 blocks, which are separated in four machines in one stratum. For CDTF/SALS, different colored blocks, which are decentralized in different machines, represent the data for optimizing different latent factors. One block could be delivered to different machines.}
\label{fig:compare}  
\vspace{0.3cm} 
\end{figure}

To overcome the problem of high communication cost in algorithms like $FLEXIFACT$, a scalable tensor factorization CDTF (Coordinate Descent for Tensor Factorization) is developed by Shin et al.~\cite{shin2017fully}, which extends CCD++~\cite{yu2012scalable} to higher orders based on coordinate descent. Concretely, each parameter can be updated by the following rule:
\begin{equation}
        {\mathbf A}_n(i,j) = \frac{\sum_{(i_1,i_2,\dots,i_N) \in \mathbf{\Omega}_i^{(n)}}(\hat{r}_{i_1,i_2,\dots,i_N}\prod_{k \neq n}{\mathbf A}_k(i,j))}{\lambda + \sum_{(i_1,i_2,\dots,i_N) \in \mathbf{\Omega}_i^{(n)}}\prod_{k \neq n}{\mathbf A}_k(i,j)},
\end{equation}
where $\hat{r}_{i_1,i_2,\dots,i_N}$ is the rank-one factorization of tensor $\hat{\pmb{\mathscr{R}}}$ and can be updated with the complexity of $O(\|\mathbf{\Omega}_i^{(n)}\|N)$ by the following rule:
\begin{equation}
        \hat{r}_{i_1,i_2,\dots,i_N} \leftarrow r_{i_1,i_2,\dots,i_N} + \prod_{n=1}^N{\mathbf A}_n(i_n,k),
\end{equation}
where $r_{i_1,i_2,\dots,i_N} = x_{i_1,i_2,\dots,i_N} - \sum_{s=1}^R\prod_{k \neq n}\mathbf{A}_k(i,s)$ that can be updated with the complexity of $O(\|\mathbf{\Omega}_i^{(n)}\|N)$ by the following rule:
\begin{equation}
        r_{i_1,i_2,\dots,i_N} \leftarrow r_{i_1,i_2,\dots,i_N} + ({\mathbf A}_n^{old}(i_n, k)-{\mathbf A}_n(i_n, k))\prod_{l \neq n}{\mathbf A}_l(i_l,k).
\end{equation}
The update sequence of parameters in CDTF adopts the column-wise order by updating the $k^\text{th}$ column of a latent matrix and then moving on to the $k^{\text th}$ column of the next latent matrix. After updating the $k^\text{th}$ columns of all latent matrices, the $(k+1)^{\text th}$ columns of latent matrices get started to be updated. Extensively, Shin et al. propose another scalable tensor factorization algorithm based on subset alternating least square (SALS) which updates each $C (1 \leq C \leq R)$ columns of latent matrices row by row. On the contrary, CDTF updates each column entry by entry; ALS updates all $R$ columns row by row. Both CDTF and SALS are implemented in MapReduce~\cite{shin2017fully}. The differences between the decentralization strategies used in $FLEXIFACT$ and CDTF/SALS are compared in Figure~\ref{fig:compare}.

Besides the aforementioned approaches, there are also other scalable tensor completion or sparse tensor decomposition approaches focusing on either CP format~\cite{rai2014scalable, karlsson2016parallel,kaya2016parallel,papastergiou2017distributed,li2017model,smith2017constrained,ge2018distenc} or Tucker format~\cite{liu2014generalized, smith2017accelerating,oh2017scalable}. As the end of this subsection, we compare several the state-of-the-art distributed/parallel scalable tensor completion / sparse tensor decomposition algorithms in Table~\ref{CompScalableMethods}, from the aspects of scalability, platform, and complexity.

%{\blue

%\cite{karlsson2016parallel}Parallel Algorithms for Tensor Completion in the CP Format

%A Distributed Proximal Gradient Descent Method for Tensor Completion

%Generalized Higher-Order Orthogonal Iteration for Tensor Decomposition and Completion

%Parallel CP decomposition of sparse tensors using dimension trees

%Accelerating the tucker decomposition with compressed sparse tensors

%}

\begin{table*}[t!]\tiny
        \vspace{-0pt}
        \centering
        \scalebox{0.8}{
        \begin{tabular}{|c| c| c| c| c| c| c| c| c|} \hline
                & Parallel ALS~(generalized from \cite{zhou2008large}) & PGSD~\cite{zinkevich2010parallelized} & DFacTo \cite{choi2014dfacto} & FLEXIFACT~\cite{beutel2014flexifact} & GigaTensor~\cite{kang2012gigatensor} & HaTen2~\cite{jeon2015haten2} & CDTF \cite{shin2017fully}\\ \hline\hline
                                            \rowcolor[gray]{0.7}
                \textbf{Scalability} & & & & & & &\\
                Dimension & \checkmark & \checkmark & \checkmark & & \checkmark & \checkmark & \checkmark \\
                No. of Parameters & & & \checkmark & \checkmark & & & \checkmark \\
                No. of Machines & \checkmark & & \checkmark &  & \checkmark & \checkmark & \checkmark \\
                Rank & & & \checkmark & \checkmark & \checkmark & \checkmark & \checkmark \\ \hline
                \rowcolor[gray]{0.7}
                \textbf{Platform} & & & & & & & \\
                MPI & & & \checkmark & & & & \\
                MapReduce & & & & \checkmark &  \checkmark & \checkmark &  \checkmark\\
                Parallel & \checkmark & \checkmark & & & & &\\ \hline
               \rowcolor[gray]{0.7}
                \textbf{Complexity} & & & & & & & \\
                Time   & $O( (|\Omega|(N + K) + IK^2)NK/M)$ & $O(|\Omega|NK/M)$ & $O((|\Omega|+nnz( \mathbf{X}_{(2)}))K/M)$ & $O(|\Omega|NK/M)$ & $O(IK^2)$ & $O(|\Omega|NK/M)$ & $O(|\Omega|N^2K/M)$ \\
                Space & $O(NIK)$ & $O(NIK)$ & $O(nnz( \mathbf{X}_{(2)}))$ & $O(NIK/M)$ & $O(MK^2)$ & $O(|\Omega|K)$ & $O(NIK)$ \\
                Communication & $O(NIK)$ & $O(NIK)$ & $O(nnz( \mathbf{X}_{(2)}))$ & $O(M^{N-2}NIK)$ & $O(MK^2)$& $O(|\Omega|K)$ & $O(NI)$ \\\hline
        \end{tabular}
        }
        \caption{Comparison between distributed/parallel scalable tensor completion/decomposition algorithms}
        \label{CompScalableMethods}
        \vspace{-0pt}
\end{table*}

\subsection{Other Approaches}

%As far as we know, there hasn't been a lot of methods focus on the scalable tensor completion problem with other methods except tensor completion. However, the 

Although CP and Tucker-based methods are the main focus in existing  literatures, more and more researches are trying to exploring other methods to tackle the scalability issue in tensor completion problem. In the context of CP rank, Cheng et al.~\cite{cheng2016scalable} put their focus on approximating the tensor trace norms to achieve computational efficiency. By constructing a polytopal approximation of the tensor dual spectral norm, the completion problem regularized with tensor spectral norm could be solved by the generalized conditional gradient algorithm~\cite{harchaoui2015conditional}. Another recent approach focusing on scalability issue of the trace norm regularizer in the LRTC problem is described in~\cite{guo2017efficient}. The authors utilize the Frank-Wolfe (FW) algorithm, which has been successfully applied for matrix completion problem with nuclear norm regularization and propose a time and space-efficient algorithm for the sparse LRTC problem. Since the scalability issue appeared in multiway datasets is not only caused by the size of each dimension but also caused by the number dimensions, some researchers focus on the developing scalable algorithm to address the high-order challenge in the scalability tensor completion problem.  A recent model by  Imaizumi et al.~\cite{imaizumi2017tensor} leverages the tensor train (TT) decomposition to address the tensor completion problem for higher-order tensors. To achieve the theoretical veracity from the statistical perspective, they first propose a convex relaxation of the TT decomposition and derive the completion error bound. By developing a randomized alternating least square optimization approach, a scalable completion algorithm is then proposed to achieve both time efficiency and space efficiency.

% Efficient Sparse Low-Rank Tensor Completion Using the Frank-Wolfe Algorithm

%First, we introduce a convex relaxation of the TT decomposition problem and derive its error bound for the tensor completion task. Next, we develop an alternating optimization method with a randomization technique, in which the time complexity is as efficient as the space complexity is. I
% the space-efficient 
% introduce trying to use with tensor-train decompositions.
% On Tensor Train Rank Minimization: Statistical Efficiency and Scalable Algorithm
%Scalable and sound low-rank tensor learning

%Although some of them neither possess distributed characters nor , they have such as solving the higher-order 

\section{Dynamic Tensor Completion}
\label{sec:dy}
Beyond traditional static or batch setting, with the increasing amount of high-velocity streaming data, more concerns have been addressed on dynamic tensor analysis~\cite{Nion:Tensor,sun2006beyond, sun2008incremental, huang2013fast}. As most of the existing surveys are lack of detailed introduction of dynamic tensor analysis, we first give a brief investigation of some existing works on dynamic tensor analysis especially focusing on dynamic tensor factorization, and then introduce several dynamic tensor completion algorithms.

As for CP decomposition, Nion and Sidiropoulos~\cite{Nion:Tensor} introduce two adaptive PARAFAC algorithms adopting the recursive least square and simultaneous diagonalization tracking approaches to address the online third-order tensor factorization problem. Recent work by Zhou et al.~\cite{zhou2016accelerating} develops an accelerated online CP algorithm that can incrementally track the decompositions of one-mode-change $N^\text{th}$ tensors. Phan et al.~\cite{phan2011parafac} partition a large-scale tensor into some small grids and develop a grid-based scalable tensor factorization method (gTF), which could also be easily applied to address the streaming tensor factorization problem.

Besides CP decomposition, some Tucker decomposition methods were also proposed. One of the earliest work is conducted by Sun et al.~\cite{sun2006beyond, sun2008incremental}. They propose two dynamic Tucker factorization methods  based on the incremental update of covariance matrices for dynamic tensor analysis. Subsequently, Yu et al.~\cite{Rose:Tucker} raised an accelerated online tensor learning algorithm (ALTO) to conduct streaming Tucker factorization. Several other online Tucker decomposition methods are conducted recently not only targeting on the one-mode-increase pattern~\cite{hu2011incremental, sobral2014incremental} but also deriving possible solutions for multi-aspect streaming patterns drawing upon matrix-based online-learning methods such as incremental SVD~\cite{ma2009dynamic}. Furthermore, a histogram-based approach~\cite{fanaee2015multi} conducted on multi-aspect streaming tensor analysis could be regarded as the pioneering research on the multi-aspect streaming analysis.

%{\blue Fast and Guaranteed Tensor Decomposition via Sketching}

\subsection{Streaming Tensor Completion}
% The outline of the algorithm is provided in Algorithm 2.

%\setlength{\textfloatsep}{10pt}
%\begin{algorithm}[t!]
%\setstretch{1.5}
% \KwIn{$\{ \mathbf{Y}_t, \mathbf{\Omega}_t \}_{t=1}^\infty$, $\{\bar{\mu}[t]\}_{t=1}^\infty$, $R$, $\lambda_t$}
% \KwOut{$\{\mathbf{X}_t, \mathbf{A}[t],\mathbf{B}[t],\gamma[t]   \}_{t=1}^\infty$ }
% %Construct the similarity matrix $\mathbf{S}$ \;
%Initialize $\mathbf{A}[0],\mathbf{B}[0]$ at random, and $\bar{\mu}[0]>0$.\\
%\For{$t =0,1,2,\ldots$}{
%$\mathbf{A}^{\top}[t] \triangleq [\mathbf{\alpha}_1[t],\ldots,\mathbf{\alpha}_M[t]]$ and $\mathbf{B}^\top[t] \triangleq [\mathbf{\beta}_1[t],\ldots,\mathbf{\beta}_N[t]]$ \;
%$\gamma[t]=[\lambda \mathbf{I}_R + \sum_{(m,n)\in \Omega_t} (\mathbf{\alpha}_m[t]\odot\mathbf{\beta}_n[t])(\mathbf{\alpha}_m[t]\odot\mathbf{\beta}_n[t])^\top]^{-1} \sum_{(m,n)\in \Omega_t} \mathbf{Y}_t(m,n)(\mathbf{\alpha}_m[t]\odot\mathbf{\beta}_n[t]) $\;
%$\mathbf{A}[t+1]=(1-\frac{\lambda_t}{t\bar{\mu}[t]})\mathbf{A}[t]+\frac{1}{\bar{\mu}[t]}[\mathbf{\Omega}_t\odot(\mathbf{Y}_t-\mathbf{A}[t]\text{diag}(\gamma[t])\mathbf{B}^\top[t])]\mathbf{B}[t]\text{diag}(\gamma[t])$\;
%$\mathbf{B}[t+1]=(1-\frac{\lambda_t}{t\bar{\mu}[t]})\mathbf{B}[t]+\frac{1}{\bar{\mu}[t]}[\mathbf{\Omega}_t\odot(\mathbf{Y}_t-\mathbf{A}[t]\text{diag}(\gamma[t])\mathbf{B}^\top[t])]^\top\mathbf{A}[t]\text{diag}(\gamma[t])$
%%Update $\mathbf{A}_n^{(0)}$ and $\mathbf{A}_n^{(1)}$ using Eq.~\eqref{equ:MTCupdate}\;
%%Perform a gradient step on $\mathbf{h}^t_i$ using Eq.~\eqref{equ:SANRH}\;
%}
%\caption{Streaming Tensor Completion}
% \label{alg:Dynamic}
%\end{algorithm}

Although various researches have been conducted on dynamic tensor analysis, there hasn't been much work focusing on the tensor completion problem with dynamic patterns. Early work of Meng et al.~\cite{meng2003line} use PARAFAC models to in-filling missing future data and have shown to overcome the challenge of signal monitoring. Matsubara~\cite{matsubara2012fast} use topic modeling with Gibbs sampling method to get the posterior latent factors of each mode to solve the future traffic forecasting problem, e.g., how many clicks a user may generate tomorrow. However, this approach is more close to being a pure future prediction problem rather than a completion problem. A more proper dynamic completion problem is addressed in a recent approach proposed by Mardani et al.~\cite{mardani2014imputation, Mardani:Tensor}. {They focus on streaming tensor completion problem and propose an online Stochastic Gradient Decent algorithm for tensor decomposition and imputation. Figure~\ref{fig:stc}~\cite{mardani2014imputation} illustrates the difference between batch tensor completion and streaming tensor completion problem. For a third-order tensor of size $M\times N\times T$, where its third mode is the temporal mode, all of the incomplete slices $\{\mathbf{\Omega}_t \ast \mathbf{Y}_t\}_{t=1,\ldots,T}$ are assumed to be collected in advance in the batch setting while in streaming tensor completion, each slice $\mathbf{\Omega}_t \ast \mathbf{Y}_t$ becomes available until the corresponding time $t$.

To address this problem, Mardani et al.~\cite{Mardani:Tensor} propose a streaming version of CP decomposition and achieve the completion purpose by adding the separable nuclear-norm regularizations describe in Equation~\eqref{equ:sep}. Using a three-order tensor as an example, the main objective function is a time-wise exponentially-weighted least-squares loss defined as follows:
\begin{equation}
\vspace{-0cm}
\begin{aligned}
\mathcal{L}= \frac{1}{2}\sum_{t=1}^T \theta^{T-t} \big[  \pmb{\Omega} \ast (\mathbf{Y}_t-\mathbf{A}\text{diag}(\pmb{\gamma}_t )\mathbf{B}^\top )  
 +\frac{\lambda_t}{\sum_{t=1}^T \theta^{T-t} }(\|\mathbf{A}\|_\text{F}^2 + \|\mathbf{B}\|_\text{F}^2)+ \lambda_t \| \pmb{\gamma}_t \|^2   \big],
\end{aligned}
\vspace{-0cm}
\label{equ:stc}
\end{equation}
where $\pmb{\gamma}_t^\top$ represents the $t$-th row of the CP factorized matrix $\mathbf{C}\in \mathbb{R}^{T \times R} $, $0 < \theta \le 1 $ is called the forgetting factor to decay the influence of the past data so that facilitating the subspace tracking in non-stationary environments~\cite{solo1994adaptive,Mardani:Tensor}. $\{ \lambda_t\}_{t=1,\ldots,T}$ are the regularization hyperparameters controlling the regularization terms.

The optimization is performed with an online updating scheme, which is a generalization of matrix online PCA algorithm~\cite{feng2013online}, to factorize and impute the tensor ``on the fly''. The main idea is that: when a new data slice $\mathbf{\Omega}_t \ast \mathbf{Y}_t$ arrives at time $t$, we first fix the factors of the non-incremental modes (i.e., $\mathbf{A}$ and $\mathbf{B}$) and calculate the new temporal factor $\gamma_t$ with an close-formed solution, which is derived based on the ridge-regression problem of $\pmb{\gamma}_t$ using Equation~\eqref{equ:stc}. And then update the non-incremental factors $\mathbf{A}$ and $\mathbf{B}$ with the stochastic gradient descent.  Similar with online PCA, the latent assumption of these online SGD methods should be a rarely and smoothly changed streaming tensor subspace, such as surveillance video streaming, which allows them to pursue the real subspace iteratively. In some real-world situation, the processing speed could be faster than data acquiring speed. Kasai~\cite{kasai2016online} proposes another online CP decomposition algorithm OLSTEC while conducting data imputation based on recursive least squares algorithm. They consider the situations where the data processing speed is much faster than data acquiring speed and propose a more rapid convergence algorithm for tracking dramatically changed subspace. By storing auxiliary matrices of non-increasing modes, these size fixed modes could be efficiently updated by considering a parallel set of smaller least square problems. The incremental subspace of the time mode is obtained with a one-time pursuit during each timestamp. Although sacrificing time complexity to some extent (increased from $\mathcal{O}(|\Omega_t|R^2)$ to $\mathcal{O}(|\Omega_t|R^2+(I_1+I_2)R^3 )$ for a third-order tensor $\pmb{\mathscr{X}}\in \mathbb{R}^{I_1\times I_2\times I_3}$) , OLSTEC is able to converge faster for tracking dramatically changed low-rank subspace comparing to SGD based method.

\begin{figure}
\centering
\vspace{-0cm}
\includegraphics[height=3cm, width=12.5cm]{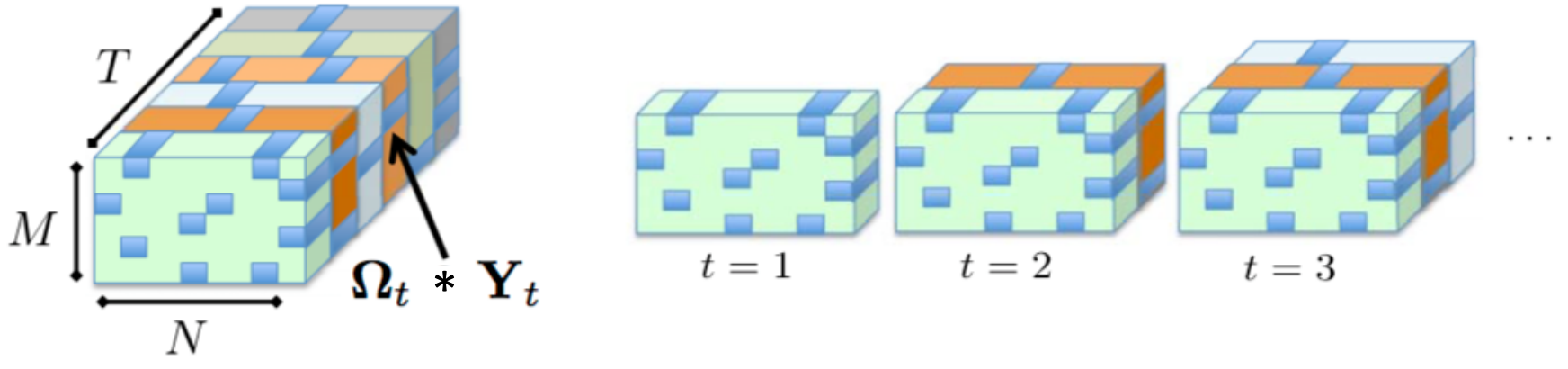} %\vspace{-.4cm} 
\vspace{-0.4cm} 
\caption{Batch Tensor Completion (left) Versus Steaming Tensor Completion (right).~\cite{Mardani:Tensor}}
\label{fig:stc}  
\vspace{0cm} 
\end{figure}

\subsection{Multi-Aspect Streaming Tensor Completion}

\begin{figure}
\centering
\vspace{-0cm}
\includegraphics[height=3.2cm, width=13.5cm]{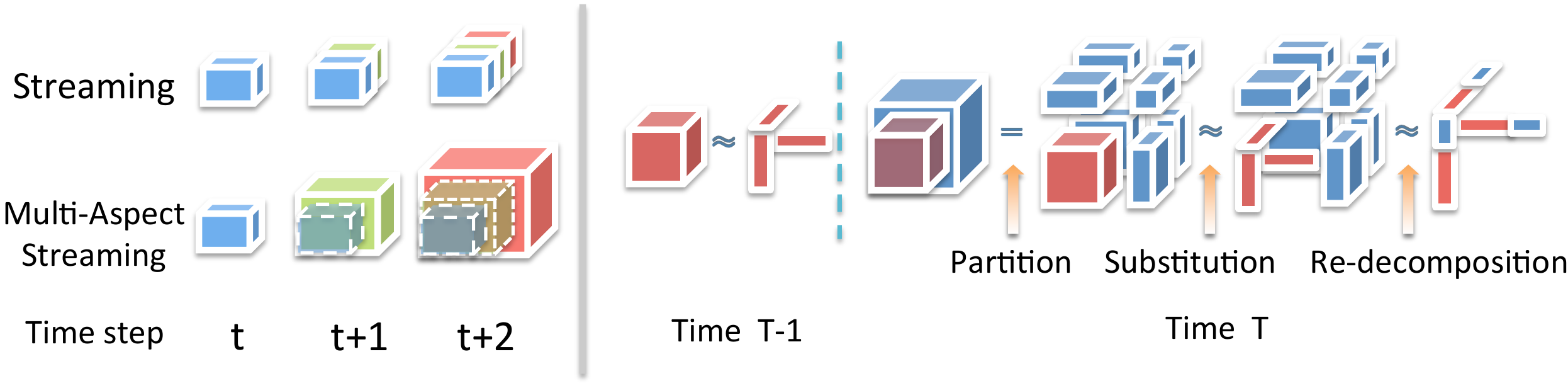} %\vspace{-.4cm} 
\vspace{-0cm} 
\caption{Streaming Tensor Completion Versus  Multi-Aspect Steaming Tensor Completion(left), Multi-Aspect Streaming Tensor Decomposition (right).~\cite{song2017multi}}
\label{fig:mastc}  
\vspace{0cm} 
\end{figure}

Recently, several researchers focus on a more general situation in tensor related problems, which is called multi-aspect streaming~\cite{song2017multi, fanaee2015multi, nimishakavi2018inductive}. As shown in the left figure in Figure~\ref{fig:mastc}, different from the streaming tensor setting, in the multi-aspect streaming setting, a tensor could increase on any of its modes, which imposes greater challenge in coordinating multidimensional dynamics, handling the higher time and space complexity as well as imputing the incremental missing entries, etc.~\cite{song2017multi}.

Song et al.~\cite{song2017multi} propose the first Multi-Aspect Streaming Tensor Completion method (MAST), which could handle the all-mode-change tensor dynamics while imputing the missing entries. The core model is a CP-based dynamic tensor decomposition model consisting of three operations, i.e., partition, substitution and re-decomposition. The key idea of its updating scheme at each pair of consecutive timestamps is shown in Figure~\ref{fig:mastc}. Suppose we obtain a small tensor along with its CP decomposition at time $T-1$, and it increases into a larger tensor at time $T$. Since the old tensor is a sub-tensor of the new one, we could partition the large tensor into smaller blocks based on the size of the old tensor and approximately substitute the old tensor with its CP decomposition. This substitution would highly improve the optimization speed especially when the ratio $\mathbf{t} = \frac{\text{size(old tensor)}}{ \text{size(new tensor)} }$ is large. Similar with Equation~\eqref{equ:stc}, a forgetting factor is also used during the optimization to alleviate the error induced by the substitution step. The authors also describe a special case, which is called temporal multi-aspect streaming, and provide a corresponding algorithm to further reduce the completion error based on this special pattern.

In addition to CP-based models, Nimishakavi et al.~\cite{nimishakavi2018inductive} propose a Tucker-based model called SIITA which could handle the multi-aspect streaming tensor completion while incorporating auxiliary information. It is an online inductive framework extended from the matrix completion framework proposed by~\cite{natarajan2014inductive,si2016goal} and achieves superior performance in both batch and streaming settings. As the dynamic tensor completion are still in its early stage, there are still vast opportunities and challenges that are valuable and attracted to delve and pursue.

%By substituting the partitioned small tensor
%It also provides another possible vision of the dynamic tensor decomposition/completion problem for future research.

}

%online, process sequentially to limit latency
%4D focal stack light field cameras20
%reconstruction: 2D->4D

%applications and influence of Tensor Completion to Methodologies (such as Signal Processing, Compressed Sensing, Numerical Analysis) and Application Domains

\section{Applications }
\label{sec:app}
The tensor completion problem arises across many application scenarios. In this section, we address several domains including social sciences, computer vision and signal processing, healthcare and bioinformatics, and others. For each application, three questions are mainly discussed: (1) what the concrete problem is; (2) how it is formulated as a tensor completion problem; (3) and what kind of methods are applied.

%To give a better distinguish among different applications, we first introduce the tensor completion problem appeared in several application domains from~\Crefrange{subsec:social}{subsec:chem} and then describe two more application fields, i.e. computer vision and signal processing, influence of tensor completion to methodologies

%Social science is a major category of academic disciplines, concerned with society and the relationships among individuals within a society. It in turn has many branches, 

\subsection{Social Computing}\label{subsec:social}

Social computing concerns with applying computation methods to explore individual activities and relationships within societies. It contains various interdisciplinary branches and includes multiple problems that could be modeled as tensor completion problems.

\subsubsection{Link Prediction}
Link prediction aims at predicting missing edges in a graph or the probabilities of future node connections. Traditionally, this  problem is treated as a matrix completion problem aiming at imputing the missing or observed links based on the existing node connections. However, dynamic interactions over time~\cite{acar2009link}, as well as multiple types of node interactions~\cite{liu2014factor}, introduce additional dimensions, leading the tensor completion algorithms to be naturally used to handle the multidimensional dataset.

There are two commonly used approaches, which extend the traditional graph topology matrix into tensors based on different data structures. (1) By adding the temporal mode into a 2-D bipartite graph, it natural derives a binary completion problem of a three-order tensor. Acar et al.~\cite{acar2009link} exploit CP decomposition with a heuristic definition of CP similarity score for future link prediction. Their subsequent work in~\cite{dunlavy2011temporal} gives another definition of CP score using the Holt-Winters forecasting method~\cite{chatfield1988holt} and proves the effectiveness of tensor-based methods in forecasting the varying periodic patterns among links. (2) Different type of interactions among nodes could also be defined as the third mode~\cite{liu2014factor,kashima2009link}. Based on this construction, Kashima et al.~\cite{kashima2009link} first utilize auxiliary similarity information of nodes for tensor link prediction. Ermi{\c{s}} et al.~\cite{ermics2015link} address the link prediction problem by jointly analyzing multiple sources via coupled factorization. They apply the proposed generalized framework on two different scale real-world datasets UCLAF (user-location-activity) and Digg (user-story-comment) and compare the affect of different loss function and factorization models in practice. See also Ermi{\c{s}}~ et al.~\cite{ermics2012link}.

\subsubsection{Recommender Systems}
Recommender systems aim at predicting the users' preferences from partial knowledge for personalized item recommendation. It naturally derives a completion problem if we treat unobserved ``ratings'' of users to items as missing entries of data matrices or tensors. Due to the ternary relational nature of data in various situations (e.g., Social Tagging Systems (STSs)), many recommendation algorithms, which are originally designed to operate on matrices, cannot be easily applied on these multidimensional datasets~\cite{symeonidis2016matrix}. Early works by Rendle et al.~\cite{rendle2009learning,rendle2010pairwise,rendle2010factorization} illustrate the power of tensor-based methods for personalized tag recommendation. They proposed three different methods to recommend new items a user would like to tag. Their first approach focuses on learning the best ranking by minimizing Area Under Curve score rather than an element-wise error. This method is shown to outperform both tensor-based method HOSVD and ranking methods such as PageRank in quality and runtime. Their second approach is a special case of CP decomposition. This approach separates the trinity interaction among user-item-tag into the summation of pairwise interactions (user-tag, user-item, item-tag) and is proved to achieve higher prediction accuracy faster. Finally, they propose a mixture approach called factorization machines (FMs) by combining factorization models with Support Vector Machines (SVM). Benefited from factorization models, the model could nest interactions among variables to overcome the sparsity scenarios where SVMs fail. Furthermore, factorization machines are general predictors, which could mimic both matrix and tensor-based factorization models such as SVD++~\cite{koren2008factorization} and their second approach PITF. Other attempts such as HOSVD-based dimensionality reduction~\cite{symeonidis2008tag, peng2010collaborative} have also been considered for personalized tag recommendations or tag-involved recommender systems. 

Another way of transforming a user-item matrix into a tensor object is to incorporate temporal information. One of the early approaches is a Bayesian Probabilistic Tensor Factorization model proposed by Xiong et al.~\cite{xiong2010temporal}. They extend the matrix-based probabilistic approach as well as the MCMC optimization method proposed in~\cite{salakhutdinov2008bayesian} and demonstrates the advantage of a high-order temporal model over matrix-based static models. A more general case of this model is considered by Karatzoglou et al.~\cite{karatzoglou2010multiverse}. They take advantage of different types of context information by considering them as additional dimensions besides traditional 2-D user-item matrix.  They exploit Tucker factorization optimized by SGD method in which missing entries are skipped during the decomposition process and reconstructed in the end. This multiverse recommendation approach outperforms traditional matrix-based systems by a wider margin which demonstrates the superiority of multidimensional analysis when more contextual information is available. 

In some situations, side information of users, items as well as features are available~\cite{singh2008relational,zheng2010collaborative,ge2016taper,bahadori2014fast}, which motivates us to combine auxiliary information for collective learning. Borrowing the idea from matrix case~\cite{singh2008relational}, Zheng et al.~\cite{zheng2010collaborative} target the problem of location-based activity recommendation system. They couple four kinds of side information including location features, GPS visiting information (user-location) as well as two similarity relationships of users and activities respectively. These auxiliary regularizers effectively alleviate the sparsity problem while mining knowledge from GPS trajectory data for mobile recommendations on both locations and activities. 

Recently, Ge et al.~\cite{ge2016taper} propose a model called ``Taper'' which is a contextual tensor-based approach tailored towards the personalized expert recommendation. They work on recommending personalized topics produced by experts (high-quality content producers) in social media in which dataset is constructed as a third-order tensor (user-expert-topic). Through modeling the homogeneous similarities between homogeneous entities (e.g., users and users) and heterogeneous similarities between pair-wise entities (e.g., users and experts), the proposed framework can achieve high-quality recommendation measured by precision and recall. 

Besides methods mentioned above, other models (e.g., Parafac2 model~\cite{pantraki2015automatic}) or optimization techniques (e.g., Riemannian optimization~\cite{kasai2015riemannian}) are also applied in recommender systems. In general, most of the applications on recommender systems are focused on decomposition-based approaches. This is because: (1) Factorization models are easy to implement and effective in learning latent factors among users or items by incorporating interactions among different types of modes. (2) They have advantages in dealing with sparsity or cold start problem which are usually encountered in recommender systems as described in~\cite{rendle2010factorization}.

\subsubsection{Urban Computing}
% Hayashi et al.~\cite{hayashi2010exponential} consider both missing value prediction problem and anomaly detection. The authors study the probabilistic factorization model with heterogeneously attributed arrays and developed an generalized exponential family factorization (ETF) approach. Their proposed framework is applied on a heterogeneous office-logging dataset which includes six types of human behaviors in C\&C Innovation Research Laboratories (CCIL), NEC Corporation. The factor matrices of user modes are served as the inputs of outlier detection algorithms. By comparing with pTucker method~\cite{chu2009probabilistic}, ETF achieves more robust detection performance under appropriate assumption of exponential-family distributions.

Urban computing deals with the human behaviors and mobilities with the help of computing technology to benefit living quality in urban areas. ``Traffic and mobility'' is one of the most significant topics in this field, which is correlated with tensor completion problem. For example, in intelligent transportation systems, outlier data may be generated due to the malfunctions in collection procedure and record systems. Some researchers connect this problem with tensor completion problem with a variant setting, which they assume the positions of missing parts are not known beforehand~\cite{tan2013traffic}. In another word, partial existed entries might entirely be some noise data (outliers), and their actual values are still unknown~\cite{tan2013traffic}. This corrupted data recovery problem is addressed by robust tensor recovery framework~\cite{tan2013traffic}, which is proved to outperform a traditional trace-norm approach proposed by Gandy et al.~\cite{gandy2011tensor}. Wang et al.~\cite{wang2014travel} focus on the problem of estimating the travel time of paths in a city. They construct a third-order tensor (road segments $\times$ drivers $\times$ time slots) based on the real-time GPS trajectories and the road network data. To address the high sparsity in this tensor, the authors build another denser tensor based on the historical trajectories and two auxiliary matrices storing the geographic similarity between different road segments and the correlation among the various time slots. A Tucker-based Coupled Matrix Tensor Factorization algorithm is used for jointly decomposition, recovery and estimation. Beyond the batch setting, Tan et al.~\cite{tan2013new} propose a Windows-Based Tensor Completion algorithm to tackle the short-term traffic prediction problem. They form the problem as a third-order tensor (week$\times$day$\times$point) to leverage the strong similarity between the same day in different weeks. The algorithm handles the time dependency through a sliding window. Each sliding window consists of a subset of matrix slices in the matrix stream, which is decomposed and imputed with Tucker-based completion methods.

%Besides Several other related areas including 

%\cite{tan2013new}

\subsubsection{Information Diffusion} 
Information diffusion refers to a process by which a piece of information or knowledge is spread initiated by a sender and reaches receivers through medium interactions~\cite{zafarani2014social}. Ge et al.~\cite{Ge:Tensor}  linked the problem of modeling and predicting the spatiotemporal dynamics of online memes with tensor completion. The meme data is constructed as a third-order tensor (meme $\times$ locations $\times$ time). To uncover the full (underlying) distribution, they incorporate the latent auxiliary relationships among locations, memes, and times into a novel recovery framework. Experiments on the real-world twitter hashtag dataset with three kinds of data missing scenarios (random missing, missing entire memes at specific locations, missing entire locations) validate the effectiveness of their tensor-based framework and the auxiliary spatiotemporal information.

\subsubsection{Computer Network}
Network traffic matrix records the amount of information exchanged between the source and destination pairs such as computers and routers. A third-tensor object could be constructed since the matrix evolves with time.  Acar et al.~\cite{acar2011scalable} address the completion problem in computer network traffic where missing data arises due to the high expense of the data collection process. The dataset is formed as a third-order tensor denoted as source routers $\times$ destination routers $\times$ time, where entries indicate the amount of traffic flow sent from sources to destinations during specific time intervals. Their proposed method-CP Weighted OPTimization-enjoys both recovery accuracy and scalability.

There are also various other types of applications in social computing domain, which are correlated with tensor completion problems or leverage related approaches such as social-aware Image tag refinement~\cite{tang2017tri}, truth discovery~\cite{xiao2015believe}, and so on. It is reasonable to believe that tensor completion problem would affects and promotes more and more applications in this area.

%Reasoning With Neural Tensor Networks for Knowledge Base Completion

\subsection{Healthcare, Bioinformatics and Medical Applications}

Medical images such as Magnetic Resonance Imaging (MRI) or Computed tomography(CT) scans are one of the most powerful tools for medical diagnosis. Due to the fast acquisition process, these image datasets are often incomplete. Early applications by Gandy et al.~\cite{gandy2011tensor} were performed on three medical MRI imaging (KNIX: MRI scans of a human knee, INCISIX: MRI dental scans, BRAINIX: MRI brain scans). Bazerque~\cite{bazerque2013rank} apply a probability PARAFAC approach on corrupted brain MRI images. By incorporating prior information under the Bayesian framework, their method is able to enhance the smoothing and prediction capabilities. Similarly, Liu et al.~\cite{liu2015trace} also use Brain MRI dataset for testing their completion method which is a combination of trace-norm based and decomposition-based approach. Furthermore, the dynamic algorithm proposed by Mardani et al.~\cite{Mardani:Tensor} has also been applied to streaming cardiac MRI data for fast online subspace tracking.

Besides the applications on medical images, several other types of datasets and analytics are also conducted in the realm of bioinformatics and healthcare. Dauwels et al.~\cite{dauwels2011handling} apply CP decomposition to handle missing data in medical questionnaires. Bazerque et al.~\cite{bazerque2013rank} perform experiments on RNA sequencing data which represents the gene expression levels in yeast. Want et al.~\cite{wang2015rubik} focus on the problem of Computational phenotyping. By incorporating two types of knowledge-guided constraints-guidance constraint and pairwise constraint-into tensor completion process, the proposed framework ``Rubik'' can convert heterogeneous electronic health records (EHRs) into meaningful concepts for clinical analysis while proceeding completion and denoising tasks simultaneously. Finally, in an interdisciplinary field related to brain analysis and signal processing, Acar et al.~\cite{acar2011scalable} use multi-channel EEG (electroencephalogram) signals where missing data is encountered due to disconnections of electrodes and demonstrate the usefulness of scalable CP factorization method in capturing the underlying brain dynamics for incomplete data. 

%Wang et al.~\cite{wang2015rubik} add non-negative constraints on the factor matrices of the CP decomposition to enhance interpretability in health data analytics. They apply the ADMM framework and the perform a non-negative threshold iteratively on the auxiliary matrices. 

\subsection{Chemometrics}\label{subsec:chem}
Chemometrics is an area of using mathematical and statistical tools to improve the chemical analysis. Appellof and Davidson~\cite{appellof1981strategies} are credited as the first one who applies tensor factorization in chemometrics~\cite{Kolda:Tensor}. Since then, tensor analysis has become actively researched in this field including low-rank approximation as well as missing data imputation, etc. A famous dataset for tensor factorization and completion is the  semi-realistic amino acid fluorescence data contributed by Bro and Andersson~\cite{bro1997parafac}. It consists of 5 laboratory-made solutions of three amino acids~\cite{tomioka2010estimation}. Tomasi et al.~\cite{tomasi2005parafac} first consider missing data problem using this dataset. They carry out experiments with three different types of missing elements: randomly missing values, randomly missing spectra/vectors, and systematically missing spectra/vectors. In a similar spirit, Narita et al.~\cite{Narita:Tensor} test two kinds of assumptions of missing patterns (element-wise missing and slice-wise missing) using a different chemical benchmark dataset: flow injection. This dataset is a chemical substances dataset represented as a third-order tensor $substances \times wavelengths \times reaction~times$. By incorporating auxiliary similarity information of each mode, their approach greatly improves the completion accuracy comparing to existing methods especially when observations are sparse. Other explorations of tensor completion problem in chemometrics can also be found in~\cite{smilde2005multi,chu2009probabilistic}

% and signal processing

\subsection{Computer Vision}

In the area of computer vision, various problems could be formulated as tensor completion problems, such as image inpainting~\cite{gandy2011tensor}, video decoding~\cite{liu2009tensor}, compressed sensing~\cite{duarte2012kronecker} and spectral data analysis~\cite{signoretto2011tensor}.

\subsubsection{Image Completion}
 
Image completion or image inpainting problem has been actively discussed in computer vision field for its practicability in many real-world applications, such as visual surveillance, human-machine interaction, image retrieval and biometric identification. As images are in nature multidimensional dataset (e.g., an RGB image is a third-order tensor with the color channels (red, green, blue) as its third dimension), tensor-based algorithms are well-suited in coordinating these multiple dimensions and leveraging the multilinear correlations.

Facial image analysis is one of the important research topics for multilinear image analysis~\cite{vasilescu2002multilinear}.  Signoretto et al.~\cite{signoretto2014learning} consider the problem of facial image completion using a benchmark Olivetti face dataset. They called the pure completion problem ``hard completion problem''. The corresponding soft completion problem is a semi-supervised problem which informs partial label information, i.e., simultaneously complete the images and a label matrix of images. Their spectral regularized framework tackles both pure completion problems and multi-task cases. Earlier works conducted by Geng et al.~\cite{geng2009facial,geng2011face} consider the ``hard completion problem''. They leverage the HOOI method to model face images for face completion, recognition, and facial age estimation. To address the missing value problem owing to data collection, they apply the EM-like approach to impute the missing entries and demonstrate the extraordinary performance of their multilinear subspace analysis method in age estimation. 

%Numerical analysis demonstrates the merits of the proposed approaches. 

Hyperspectral data completion is also widely discussed in image completion work. An early approach of using hyperspectral images is proposed by Gandy et al.~\cite{gandy2011tensor}. They use an URBAN hyperspectral dataset in which each image represents a different band of collected light wavelengths with a spatial resolution of $200 \times 200$ pixels. Another benchmark dataset is called ``Ribeira''\footnote{\url{http://personalpages.manchester.ac.uk/staff/d.h.foster/Hyperspectral_images_of_natural_scenes_04.html}} hyperspectral image dataset, which was first used in~\cite{foster2004information} and further utilized by Signoretto et al.~\cite{signoretto2011tensor} for tensor completion. The datasets consist of collections of raw reflectances sensed from contiguous spectral bands. One could apply tensor-based approaches via treating ``scene'' as the third mode. See also~\cite{kressner2014low, kasai2015riemannian,liu2015trace} for more completion approaches on this dataset. 

Besides these two kinds of images, many other types of images are considered in tensor completion problems, such as building facade image~\cite{tan2014tensor, liu2015trace}, reflectance data~\cite{liu2009tensor,liu2013tensor} and CT\/MRI images~\cite{tan2013low}. Either of them is natural RGB images or reconstructed datasets with an additional channel mode or scene mode. We leave the discussion of medical image analysis (mainly focus on MRI images) in the section of healthcare and medical application.

\subsubsection{Video Completion}
Videos can be naturally represented as multidimensional arrays by combing the streaming of scenes. Besides the application of video inpainting, video compression could also be treated as a completion problem during the uncompress process. A user can remove unimportant or unwanted pixels from each frame of original videos and recover it using completion tools. The earliest application of tensor completion methods on video datasets may be the trace norm approach proposed by Liu et al.~\cite{liu2009tensor}. Mu et al.~\cite{mu2014square} also apply their completion methods on color videos, which are naturally represented as four-mode tensors (length$\times$width$\times$channels$\times$frames). Kasai~\cite{kasai2016online} focuses on streaming subspace tracking from incomplete data and exploits the recursive least square (RLS) method for online CP decomposition. They evaluate the tracking performances using an airport hall surveillance video where frames of the video arrive in a sequential order. Furthermore, to test the performance of the proposed method in tracking dynamic moving background. They reconstruct the input video stream by reordering the cropped incomplete frames and show the rapid adaptivity of their method OLSTEC to the changed background. Other video completion examples could be found in~\cite{tan2013low,zhang2014novel}.

%In computer vision and graphics, many problems can be formulated as a missing value estimation problem, e.g. image in-painting [4], [22], video decoding, video in-painting [23], scan completion, and appearance acquisition completion.

\subsection{Compressed Sensing}
Compressed sensing initiated by~\cite{candes2006robust, donoho2006compressed} in 2006 allows the recovery of a sparse signal from a small number of nonadaptive, linear random measurements through efficient convex algorithms such as $\ell_1$-minimization and iterative hard thresholding~\cite{eldar2012compressed}. It does exact recoveries of high-dimensional sparse vectors after the dimensionality reduction. It is one of the most important applications in signal processing and  embedded with multitudinous multilinear datasets. For example, if we treat videos and images as signals, both of them are multidimensional data, and their compressions could be treated as the special cases of compressed sensing.

As a further extension of compressed sensing, the Kronecker-based compressed sensing model~\cite{duarte2012kronecker} has been proposed and explored in order to offer a practical implementation of compressed sensing for high-order tensors. Some researchers~\cite{caiafa2012block, caiafa2013computing, li2013generalized} focus on building generalized tensor compressed sensing algorithms by exploiting the Kronecker structure with various approaches such as $\ell_1$-minimization. Sidiropoulos et al.~\cite{sidiropoulos2012multi} recovers sparse, low-rank tensors from few measurements by leveraging compressed sensing for multilinear models of tensor data, and shows that the identifiability properties of the tensor CP decomposition model can be used to recover a low-rank tensor from Kronecker measurements (by compressing each mode separately). Concretely, it evolves into two steps: fitting a low-rank model in the compressed domain, and performing per-mode decompression. Rauhut~\cite{rauhut2015tensor} reconstructs a low-rank tensor from a relatively small number of measurements by utilizing the framework of hierarchical tensor (HT) formats based on a Hierarchical SVD-based approach followed by a similar work~\cite{da2015optimization}. However, it should be noted that this method is hard to scale tensors with large mode sizes as it strongly relies on computing SVDs of large matrices.  Friedland et al.~\cite{friedland2014compressive} exploit a unified framework of compressed sensing for high-order tensors with preserving the intrinsic structure of tensor data, which provides an efficient representation for multi-dimensional with simultaneous acquisition and compression from all tensor modes. Li et al.~\cite{li2016compressive} focus on incorporating structured sparsity into tensor representation to compress/recover the multi-dimensional signals with the varying non-stationary statistics, and inheriting the merit from tensor-based compressed sensing to alleviate the computational and storage burden in sampling and recovery. Caiafa and Cichocki~\cite{caiafa2015stable} study a fast non-iterative tensor compressed sensing method based on a low multilinear-rank model without assuming certain sparsity patterns, which is suitable for large-scale problems.
%hierarchical tucker~\cite{da2013hierarchical}???
%{\red Low Rank Tensor Recovery via Iterative Hard Thresholding} Tensor-train+hierarchical
%iterative hard thresholding method for fitting the factors of a Tucker decomposition~\cite{rauhut2017low}
%{\red In this section, we illustrate a real data application of our approach, where we reconstruct missing parts of an audio spectrogram X(f, t), that represents the STFT coefficient magnitude at frequency bin f and time frame t of a piano piece, see top left panel of Fig.3. This is a difficult matrix completion problem: as entire time frames (columns of X) are missing, low rank reconstruction techniques are likely to be ineffective. Yet such missing data patterns arise often in practice, e.g., when packets are dropped during digital communication. We will develop here a novel approach, expressed as a coupled TF model. In particular, the reconstruction will be aided by an approximate musical score, not necessarily belonging to the played piece, and spectra of isolated piano sounds.} musical audio restoration problem. ~\cite{yilmaz2011generalised}

%PARAFAC and missing values

%Since the “factors” are known to be the three amino acids, this is a perfect data for testing whether the proposed method can automatically find those factors.~\cite{tomioka2010estimation}

\subsection{Other Applications}
There are still many interdisciplinary applications which are hard to categorize into the categories above. By no means to integrate them all, we introduce two of them and leave others for interested readers to further explore. 

\subsubsection{Climate Data Analysis} How to fill in the missing entries of the climatological data is also a direction for applying tensor-based methods. Silva et al.~\cite{da2013hierarchical} apply their hierarchical Tucker algorithms with careful choice of dimension tree to interpolate frequency slices from two generated seismic data. Bahadori et al.~\cite{bahadori2014fast} analyze two datasets - the U.S. Historical Climatology Network Monthly and the Comprehensive Climate Datasets, which is a collection of the climate records of North America. Their completion method with spatial-temporal auxiliary information was tested with both cokriging and forecasting tasks and proved to be not only more efficient but also more effective in achieving lower estimation error. 

\subsubsection{Numerical Analysis}
Some applications are directly related to numerical analysis such as the reconstruction of function data and solving the parameterized linear system. One example of the former one is to recover the compress tensors related to functions with singularities~\cite{kressner2014low}. An example of the latter one could be found in the same paper where they complete the solution tensor of a parametrized linear system obtained from a discretized stochastic elliptic PDE with Karhunen-Lo{\`e}ve expansion.

\section{Experimental Setup}\label{sec:exp}
To facilitate the employment of these methods for practitioners and researchers, in this section, we introduce several important components in experimental settings including ways of synthetic data constructions and evaluation metrics.%, and available repositories.

\subsection{Synthetic Data Construction}
An ideal real-world tensor dataset with the known rank and sampling assumption are usually hard to acquire. Based on different assumptions of tensor rank, outside noises, sampling methods and auxiliary information, researchers have provided different ways to construct synthetic tensors. We introduce three ways of constructions including how to generate a rank-$R$ CP tensor, a rank $(R_1,R_2,R_3)$ Tucker tensor and a rank-$R$ CP tensor with auxiliary similarity information on each mode. For convenience, we use third-order tensors as examples.

\subsubsection{Rank-$R$ CP Tensor} For CP rank, a commonly used way could be found in~\cite{acar2011scalable}, they construct a third-order tensor $\pmb{\mathscr{X}}$ with CP rank-$R$ by randomly sampling entries of the factor matrices $\mathbf{A}\in \mathbb{R}^{I\times R}, \mathbf{B}\in \mathbb{R}^{J\times R}, \mathbf{C}\in \mathbb{R}^{K\times R}$ from the standard normal distribution $\mathcal{N}(0,1)$. Each column of the factor matrices is normalized to unit length, and the constructed tensor is denoted as:
\begin{equation}
\pmb{\mathscr{X}} = \llbracket  \mathbf{A},\mathbf{B},\mathbf{C}  \rrbracket + \eta \frac{\| \pmb{\mathscr{X}} \| }{\| \pmb{\mathscr{N}} \|} \pmb{\mathscr{N}}
\end{equation}
where $\pmb{\mathcal{N}}$ is a Gaussian noise tensor, and $\eta$ represents the noise parameter which is sometimes described as Signal-to-Noise Ratio(SNR)~\cite{liu2013tensor} or Signal-to-Interference Rate (SIR)~\cite{zhou2016accelerating}. Acar el al.~\cite{acar2011scalable} also mentioned that this way of construction ensures the uniqueness of the CP decomposition with probability one since the columns of each factor matrix are linearly independent with probability one, which satisfies the necessary conditions for uniqueness defined in~\cite{kruskal1977three}. Missing entries could be denoted by a binary tensor which is uniformly sampled from Bernoulli distribution. Some structured missing entries in the form of missing fibers, slices can also be considered. Concrete missing ratio and sampling methods should base on the incoherence assumptions described in Section~\ref{sec:stat} and specific completion methods we use.
%calculated by $\eta=10^(-SNR/20))$

\subsubsection{Rank $(R_1,R_2,R_3)$ Tucker Tensor}
Similarly, we can construct a tenor with Tucker rank $(R_1,R_2,R_3)$ utilizing the Tucker decomposition $\pmb{\mathscr{X}} =  \pmb{\mathscr{C}} \times_1  \mathbf{A}_1 \times \mathbf{A}_2 \times_3  \mathbf{A}_3$. Entries of the core tensor $\pmb{\mathscr{C}}\in \mathbb{R}^{R_1\times R_2\times R_3}$ and factorized matrices could be randomly sampled from $\mathcal{N}(0,1)$. Factorized matrices are usually orthogonalized. Ways to generate missing entries are the same as we mentioned above.

\subsubsection{Rank-$R$ CP Tensor with Auxiliary Information} We have introduced several methods for tensor completion with auxiliary similarity information in Section~\ref{sec:aux}. By utilizing the auxiliary information, one might be able to perform exact recovery even for situations of missing slices. Narita et al.~\cite{Narita:Tensor} propose a way of constructing synthetic CP tensors with tri-diagonal similarity matrices. The factor matrices ${\bf A} \in {I \times R}$, ${\bf B} \in {J \times R}$, and ${\bf C} \in {K \times R}$ are generated by the following linear formula~\cite{Narita:Tensor}:
\begin{equation*}
	\begin{aligned}
		& {\bf A}(i,r)=i\varepsilon_r+\varepsilon'_r, & i=1,2,\dots,100,r=1,2,\dots,R\\
		& {\bf B}(j,r)=j\zeta_r+\zeta'_r, & j=1,2,\dots,100,r=1,2,\dots,R\\
		& {\bf C}(k,r)=k\eta_r+\eta'_r, & k=1,2,\dots,100,r=1,2,\dots,R\\
	\end{aligned}
\end{equation*}
where $\{\varepsilon_r,\varepsilon_r',\zeta_r,\zeta_r',\eta_r,\eta_r'\}_{r=1,2,\dots,R}$ are constants generated from $N(0,1)$. The synthetic tensor is defined by $\pmb{\mathscr{X}}=\llbracket {\bf A},{\bf B}, {\bf C}\rrbracket$. Since each factor matrix is linear constructed column by column, the neighboring rows are similar to each other, the similarity matrix of each mode is a tri-diagonal matrix defined as follows:
\begin{equation}
	\Theta_i = 
	\begin{bmatrix}
		0 & 1 & 0 & \dots\\
		1 & 0 & 1 & \dots\\
		0 & 1 & 0 & \dots\\
		\vdots & \vdots & \vdots & \ddots\\
	\end{bmatrix}
\end{equation}
\vspace{-15pt}

\subsection{Evaluation Metrics}
%and which is suitable in could used based on different applications 
As completion problem could be generalized as prediction problem, metrics are usually selected based on concrete applications such as RMSE in recommender systems, AUC in binary classification, nDCG in information retrieval and so on. We introduce several commonly used metrics below. One should not be restricted to them and could select more suitable metrics based on specific problems and situations.

\subsubsection{Relative Standard Error / Percentage of Fitness / Tensor Completion Score} Relative Error is the most commonly used evaluation metric, it is formally defined as: 
\begin{displaymath}
\mbox{Relative Standard Error}= \frac{\lVert \pmb{\mathscr{X}}_\text{rec}-\pmb{\mathscr{X}}_\text{real} \rVert_F}{\lVert  \pmb{\mathscr{X}}_\text{real} \rVert_F}, \quad \mbox{Percentage of Fitness} = 1-\mbox{Relative Standard Error}
\end{displaymath}
where $\pmb{\mathscr{X}}_\text{rec}$ is the recovered tensor, $\pmb{\mathscr{X}}_\text{real}$ is the real tensor. Percentage of Fitness is a similar metric measuring the fitness of the reconstructed tensor and the ground-truth tensor. A refined relative error, called Tensor Completion Score by Acar et al.~\cite{acar2011scalable}, is also used defined as:
\begin{displaymath}
\mbox{TCS}= \frac{\lVert (1-\pmb{\mathscr{W}})\ast (\pmb{\mathscr{X}}_\text{rec}-\pmb{\mathscr{X}}_\text{real}) \rVert_F}{\lVert (1-\pmb{\mathscr{W}})\ast \pmb{\mathscr{X}}_\text{real} \rVert_F}\\
\end{displaymath}
where $\pmb{\mathscr{W}}$ is the indication tensor of the observations.

\subsubsection{MSE / RMSE / MAE} Mean-square error and root-mean-square deviation, and mean absolute error are frequently-used metrics especially in recommendation problems~\cite{karatzoglou2010multiverse,zheng2010collaborative,xiong2010temporal}. Their formal definitions are:
\begin{equation}
\begin{aligned}
& \mbox{MSE}= \frac{\lVert \pmb{\mathscr{W}} \ast ( \pmb{\mathscr{X}}_\text{rec}- \pmb{\mathscr{X}}_\text{real}) \rVert^2_F}{ \lVert  \pmb{\mathscr{W}} \rVert_F^2  } 
\quad  \mbox{RMSE}= \frac{\lVert   \pmb{\mathscr{W}} \ast ( \pmb{\mathscr{X}}_\text{rec}-\pmb{\mathscr{X}}_\text{real} ) \rVert_F}{\lVert \pmb{\mathscr{W}} \rVert_F } \\ 
& \qquad\qquad\qquad \mbox{MAE}= \frac{1}{\sum | \pmb{\mathscr{W}} |  } 
 \sum \pmb{\mathscr{W}} \ast | \pmb{\mathscr{X}}_\text{rec}-\pmb{\mathscr{X}}_\text{real} | \end{aligned}
\end{equation}

\subsubsection{Precision / Recall / Area Under Curve(AUC)} Some applications could be treated as classification problems such as link predictions or recommendations. Thus, several metrics for classification problems such as precision (or precision@k), recall, F-measure and AUC score could be found in some tensor completion papers~\cite{symeonidis2008tag,peng2010collaborative,rendle2010pairwise,ge2016taper,Ge:Tensor}. To calculate these metrics such as AUC score, one could either extract all missing entries or average these scores on each slice based on their requirements in real-world applications.

\section{Conclusion and Future Outlook}

Tensor completion problem which permeates through a wide range of real-world applications has become an actively studied field attracting increasing attention. Recent advances in both theory and practice provide us versatile and potent methods to apply on various problems even with auxiliary information, large-scale data, and dynamical patterns. However, the deeper the delving, the more challenges we will confront. Reviewing the past offers us a better vision to look forward to future directions:

\noindent\textbf{Volume: Tensor Completion with Large-Scale and High-order Data.} Large-scale and high-order properties are usually involved in real-world tensors. The increasing volume of tensor-structured data puts higher demands of the completion algorithms from both time and space perspectives. As we have described in Section~\ref{sec:scalable}, two main challenges underlying in the large-scale tensor completion problem are intermediate data explosion and data decentralization. Though multiple approaches have been proposed to address the scalability problem, there still exists a vast room for innovation and promotion. First, space and time complexities still highly limit the practical application of tensor decomposition/completion methods. Second, most of the existing work only focuses on low-dimensional tensor completion problem. Although methods based on hierarchical Tucker representations have been utilized to address high-order problems~\cite{rauhut2015tensor}, their generalizability and robustness remain unclear. Third, some methods such as sampling methods~\cite{krishnamurthy2013low, bhojanapalli2015new} require specific assumptions (e.g., incoherence assumptions), which are hard to verify and guarantee for real-world complex tensors. Fourth, as deep learning models are usually data-hungry,  is it possible to combine tensor-based algorithms with deep learning techniques to address the volume challenge as well as improving completion accuracy?

%Two, one could be by accommodating higher data volume 

\noindent\textbf{Velocity: Tensor Completion in Dynamic Data Analytics.}  With the rapid velocity of data growth, dynamic tensor completion problem has been paid more and more attention recently as we have outlined in Section~\ref{sec:dy}.  It is still riddled with various open problems in line with the intricate multi-aspect dynamic patterns. First, how to handle dynamically changing tensors? While some recent attempts have been made on solving the streaming tensor completion problem~\cite{Mardani:Tensor, song2017multi}, efforts are still needed for coordinating multidimensional dynamics induced by the uncertainty of tensor size and modes. Second, what are the influences of different dynamical patterns on the model construction, parameter selection, and imputation effectiveness? For example, dynamically changed tensors may cause the fixed rank assumption inapplicable. Arbitrarily adopting static strategies might be inappropriate and reduce the completion effectiveness. Third, whether there exist theoretical guarantees for the proposed algorithm under dynamic settings or not? To our best knowledge, bare of existing efforts have been put to provide theoretical analysis such as convergence or sufficient conditions and statistical assumptions for dynamic tensor completion.  

%remaining to be solved

\noindent\textbf{Variety: Tensor Completion with Heterogeneous Data Sources.} In practice, tensor structured data may exhibit heterogeneous properties when viewed obtained via different routes or viewed from different perspectives. It is critical, yet challenging, to consider heterogeneous data sources to benefit the tensor completion problem. First, as introduced in Section~\ref{sec:aux}, the heterogeneous data sources could serve as auxiliary information to mitigate statistical assumptions while enhancing the completion effectiveness. Thus, exploring potential ways to effectively incorporate these heterogeneous data source could be a valuable direction. For example, combining the heterogeneous deep learning techniques with tensor completion algorithms could be one possible approach. Second, in a wide variety of real-world data-driven applications, domain knowledge and expertise are beneficial for modeling, analysis, understanding, and completion. Several recent works have tried to incorporate human knowledge into completion tasks and prove to acquire favorable performance such as in the area of healthcare~\cite{wang2015rubik}. How to combine domain knowledge with tensor completion problems could be another promising future direction for exploitation and exploration. Finally, incorporating the heterogeneous data under large-scale and dynamic settings may also be interesting and meaningful to pursue.

\noindent\textbf{Other Aspects.} Besides the aforementioned ``3V'' challenges, as the big data research may also concern about the ``veracity'' and ``value'', it is intriguing to explore the tensor completion problems from these two perspectives. From the ``veracity'' perspective,  the uncertainties of the high-order datasets, the tensor completion algorithms, and completion results could be one meaningful and promising direction to pursue. From the  ``value'' perspective, whether the data is valuable to utilize or not and if the completion assumptions,  methods, and results are valuable and practical for real-world applications should also be concerned in the future.

\begin{acks}

This work is, in part, supported by DARPA (\#W$911$NF-$16$-$1$-$0565$) and NSF (\#IIS-$1657196$). The views, opinions, and/or findings expressed are those of the author(s) and should not be interpreted as representing the official views or policies of the Department of Defense or the U.S. Government.

%The following people were kind enough to read earlier versions of this manuscript and offer helpful advice on many fronts: Evrim Acar Ataman, Rasmus Bro, Peter Chew, Lieven De Lathauwer, Lars Eld ́en, Philip Kegelmeyer, Morten Mørup, Teresa Selee, and Alwin Stegeman. We are also indebted to the anonymous referees, whose feedback was extremely useful in improving the manuscript. Finally, we would like to acknowledge all of our colleagues who have engaged in helpful discussion and provided pointers to related literature, especially Dario Bini, Pierre Comon, Henk Kiers, Pieter Kroonenberg, J. M. Landsberg, Lek-Heng Lim, Carmeliza Navasca, Berkant Savas, Nikos Sidiropoulos, Jos Ten Berge, and Alex Vasilescu. Finally, we would like to pay special tribute to two colleagues who passed away while this article was in review: Gene Golub and Richard Harshman have directly and profoundly shaped our work and the work of many others. They will be dearly missed.
% * <song_3134@tamu.edu> 2018-04-25T16:24:48.129Z:
%
% ^.

\end{acks}

% Bibliography
\bibliographystyle{ACM-Reference-Format}
\bibliography{Tensor}


\begin{thebibliography}{228}

%%% the Web address.

\ifx \showCODEN    \undefined \def \showCODEN     #1{\unskip}     \fi
\ifx \showDOI      \undefined \def \showDOI       #1{#1}\fi
\ifx \showISBNx    \undefined \def \showISBNx     #1{\unskip}     \fi
\ifx \showISBNxiii \undefined \def \showISBNxiii  #1{\unskip}     \fi
\ifx \showISSN     \undefined \def \showISSN      #1{\unskip}     \fi
\ifx \showLCCN     \undefined \def \showLCCN      #1{\unskip}     \fi
\ifx \shownote     \undefined \def \shownote      #1{#1}          \fi
\ifx \showarticletitle \undefined \def \showarticletitle #1{#1}   \fi
\ifx \showURL      \undefined \def \showURL       {\relax}        \fi
% invisible to TeX
\providecommand\bibfield[2]{#2}
\providecommand\bibinfo[2]{#2}
\providecommand\natexlab[1]{#1}
\providecommand\showeprint[2][]{arXiv:#2}

\bibitem[\protect\citeauthoryear{Acar, Dunlavy, and Kolda}{Acar
  et~al\mbox{.}}{2009}]%
        {acar2009link}
\bibfield{author}{\bibinfo{person}{Evrim Acar}, \bibinfo{person}{Daniel~M
  Dunlavy}, {and} \bibinfo{person}{Tamara~G Kolda}.}
  \bibinfo{year}{2009}\natexlab{}.
\newblock \showarticletitle{Link prediction on evolving data using matrix and
  tensor factorizations}.
\newblock \bibinfo{journal}{{\em Data Mining Workshops, 2009. ICDMW'09. IEEE
  International Conference on\/}} (\bibinfo{year}{2009}).
\newblock


\bibitem[\protect\citeauthoryear{Acar, Dunlavy, Kolda, and M{\o}rup}{Acar
  et~al\mbox{.}}{2011a}]%
        {acar2011scalable}
\bibfield{author}{\bibinfo{person}{Evrim Acar}, \bibinfo{person}{Daniel~M
  Dunlavy}, \bibinfo{person}{Tamara~G Kolda}, {and} \bibinfo{person}{Morten
  M{\o}rup}.} \bibinfo{year}{2011}\natexlab{a}.
\newblock \showarticletitle{Scalable tensor factorizations for incomplete
  data}.
\newblock \bibinfo{journal}{{\em Chemometrics and Intelligent Laboratory
  Systems\/}} (\bibinfo{year}{2011}).
\newblock


\bibitem[\protect\citeauthoryear{Acar, Kolda, and Dunlavy}{Acar
  et~al\mbox{.}}{2011b}]%
        {acar2011all}
\bibfield{author}{\bibinfo{person}{Evrim Acar}, \bibinfo{person}{Tamara~G
  Kolda}, {and} \bibinfo{person}{Daniel~M Dunlavy}.}
  \bibinfo{year}{2011}\natexlab{b}.
\newblock \showarticletitle{All-at-once optimization for coupled matrix and
  tensor factorizations}.
\newblock \bibinfo{journal}{{\em arXiv preprint arXiv:1105.3422\/}}
  (\bibinfo{year}{2011}).
\newblock


\bibitem[\protect\citeauthoryear{Acar and Yener}{Acar and Yener}{2009}]%
        {acar2009unsupervised}
\bibfield{author}{\bibinfo{person}{Evrim Acar} {and}
  \bibinfo{person}{B{\"u}lent Yener}.} \bibinfo{year}{2009}\natexlab{}.
\newblock \showarticletitle{Unsupervised multiway data analysis: A literature
  survey}.
\newblock \bibinfo{journal}{{\em IEEE transactions on knowledge and data
  engineering\/}} (\bibinfo{year}{2009}).
\newblock


\bibitem[\protect\citeauthoryear{Andersson and Bro}{Andersson and Bro}{1998}]%
        {andersson1998improving}
\bibfield{author}{\bibinfo{person}{Claus~A Andersson} {and}
  \bibinfo{person}{Rasmus Bro}.} \bibinfo{year}{1998}\natexlab{}.
\newblock \showarticletitle{Improving the speed of multi-way algorithms:: Part
  I. Tucker3}.
\newblock \bibinfo{journal}{{\em Chemometrics and intelligent laboratory
  systems\/}} (\bibinfo{year}{1998}).
\newblock


\bibitem[\protect\citeauthoryear{Appellof and Davidson}{Appellof and
  Davidson}{1981}]%
        {appellof1981strategies}
\bibfield{author}{\bibinfo{person}{Carl~J Appellof} {and} \bibinfo{person}{ER
  Davidson}.} \bibinfo{year}{1981}\natexlab{}.
\newblock \showarticletitle{Strategies for analyzing data from video
  fluorometric monitoring of liquid chromatographic effluents}.
\newblock \bibinfo{journal}{{\em Analytical Chemistry\/}}
  (\bibinfo{year}{1981}).
\newblock


\bibitem[\protect\citeauthoryear{Ashraphijuo and Wang}{Ashraphijuo and
  Wang}{2017}]%
        {ashraphijuo2017fundamental}
\bibfield{author}{\bibinfo{person}{Morteza Ashraphijuo} {and}
  \bibinfo{person}{Xiaodong Wang}.} \bibinfo{year}{2017}\natexlab{}.
\newblock \showarticletitle{Fundamental conditions for low-CP-rank tensor
  completion}.
\newblock \bibinfo{journal}{{\em JMLR\/}} (\bibinfo{year}{2017}).
\newblock


\bibitem[\protect\citeauthoryear{Bader and Kolda}{Bader and Kolda}{2006}]%
        {bader2006algorithm}
\bibfield{author}{\bibinfo{person}{Brett~W Bader} {and}
  \bibinfo{person}{Tamara~G Kolda}.} \bibinfo{year}{2006}\natexlab{}.
\newblock \showarticletitle{Algorithm 862: MATLAB tensor classes for fast
  algorithm prototyping}.
\newblock \bibinfo{journal}{{\em ACM TOMS\/}} (\bibinfo{year}{2006}).
\newblock


\bibitem[\protect\citeauthoryear{Bahadori, Yu, and Liu}{Bahadori
  et~al\mbox{.}}{2014}]%
        {bahadori2014fast}
\bibfield{author}{\bibinfo{person}{Mohammad~Taha Bahadori},
  \bibinfo{person}{Qi~Rose Yu}, {and} \bibinfo{person}{Yan Liu}.}
  \bibinfo{year}{2014}\natexlab{}.
\newblock \bibinfo{booktitle}{{\em Fast multivariate spatio-temporal analysis
  via low rank tensor learning}}.
\newblock \bibinfo{publisher}{NIPS}. 3491--3499 pages.
\newblock


\bibitem[\protect\citeauthoryear{Banerjee, Basu, and Merugu}{Banerjee
  et~al\mbox{.}}{2007}]%
        {banerjee2007multi}
\bibfield{author}{\bibinfo{person}{Arindam Banerjee}, \bibinfo{person}{Sugato
  Basu}, {and} \bibinfo{person}{Srujana Merugu}.}
  \bibinfo{year}{2007}\natexlab{}.
\newblock \showarticletitle{Multi-way clustering on relation graphs}.
\newblock \bibinfo{journal}{{\em ICDM\/}} (\bibinfo{year}{2007}).
\newblock


\bibitem[\protect\citeauthoryear{Barak and Moitra}{Barak and Moitra}{2015}]%
        {barak2015tensor}
\bibfield{author}{\bibinfo{person}{Boaz Barak} {and} \bibinfo{person}{Ankur
  Moitra}.} \bibinfo{year}{2015}\natexlab{}.
\newblock \showarticletitle{Tensor prediction, rademacher complexity and random
  3-xor}.
\newblock \bibinfo{journal}{{\em CoRR, abs/1501.06521\/}}
  (\bibinfo{year}{2015}).
\newblock


\bibitem[\protect\citeauthoryear{Barak and Moitra}{Barak and Moitra}{2016}]%
        {barak2016noisy}
\bibfield{author}{\bibinfo{person}{Boaz Barak} {and} \bibinfo{person}{Ankur
  Moitra}.} \bibinfo{year}{2016}\natexlab{}.
\newblock \showarticletitle{Noisy tensor completion via the sum-of-squares
  hierarchy}. In \bibinfo{booktitle}{{\em Conference on Learning Theory}}.
\newblock


\bibitem[\protect\citeauthoryear{Bazerque, Mateos, and Giannakis}{Bazerque
  et~al\mbox{.}}{2013}]%
        {bazerque2013rank}
\bibfield{author}{\bibinfo{person}{Juan~Andr{\'e}s Bazerque},
  \bibinfo{person}{Gonzalo Mateos}, {and} \bibinfo{person}{Georgios~B
  Giannakis}.} \bibinfo{year}{2013}\natexlab{}.
\newblock \showarticletitle{Rank regularization and Bayesian inference for
  tensor completion and extrapolation}.
\newblock \bibinfo{journal}{{\em IEEE transactions on signal processing\/}}
  (\bibinfo{year}{2013}).
\newblock


\bibitem[\protect\citeauthoryear{Beutel, Talukdar, Kumar, Faloutsos,
  Papalexakis, and Xing}{Beutel et~al\mbox{.}}{2014}]%
        {beutel2014flexifact}
\bibfield{author}{\bibinfo{person}{Alex Beutel}, \bibinfo{person}{Partha~Pratim
  Talukdar}, \bibinfo{person}{Abhimanu Kumar}, \bibinfo{person}{Christos
  Faloutsos}, \bibinfo{person}{Evangelos~E Papalexakis}, {and}
  \bibinfo{person}{Eric~P Xing}.} \bibinfo{year}{2014}\natexlab{}.
\newblock \showarticletitle{Flexifact: scalable flexible factorization of
  coupled tensors on hadoop}.
\newblock \bibinfo{journal}{{\em ICDM\/}} (\bibinfo{year}{2014}).
\newblock


\bibitem[\protect\citeauthoryear{Beyer}{Beyer}{2011}]%
        {beyer2011gartner}
\bibfield{author}{\bibinfo{person}{Mark Beyer}.}
  \bibinfo{year}{2011}\natexlab{}.
\newblock \showarticletitle{Gartner Says Solving'Big Data'Challenge Involves
  More Than Just Managing Volumes of Data}.
\newblock \bibinfo{journal}{{\em Gartner. Archived from the original on\/}}
  (\bibinfo{year}{2011}).
\newblock


\bibitem[\protect\citeauthoryear{Bhojanapalli and Sanghavi}{Bhojanapalli and
  Sanghavi}{2015}]%
        {bhojanapalli2015new}
\bibfield{author}{\bibinfo{person}{Srinadh Bhojanapalli} {and}
  \bibinfo{person}{Sujay Sanghavi}.} \bibinfo{year}{2015}\natexlab{}.
\newblock \showarticletitle{A new sampling technique for tensors}.
\newblock \bibinfo{journal}{{\em arXiv preprint arXiv:1502.05023\/}}
  (\bibinfo{year}{2015}).
\newblock


\bibitem[\protect\citeauthoryear{Bro}{Bro}{1997}]%
        {bro1997parafac}
\bibfield{author}{\bibinfo{person}{Rasmus Bro}.}
  \bibinfo{year}{1997}\natexlab{}.
\newblock \showarticletitle{PARAFAC. Tutorial and applications}.
\newblock \bibinfo{journal}{{\em Chemometrics and intelligent laboratory
  systems\/}} (\bibinfo{year}{1997}).
\newblock


\bibitem[\protect\citeauthoryear{Bro}{Bro}{1998}]%
        {bro1998multi}
\bibfield{author}{\bibinfo{person}{Rasmus Bro}.}
  \bibinfo{year}{1998}\natexlab{}.
\newblock \showarticletitle{Multi-way analysis in the food industry: models,
  algorithms, and applications}.
\newblock \bibinfo{journal}{{\em MRI, EPG and EMA,'' Proc ICSLP 2000\/}}
  (\bibinfo{year}{1998}).
\newblock


\bibitem[\protect\citeauthoryear{Cai, He, Han, and Huang}{Cai
  et~al\mbox{.}}{2011}]%
        {cai2011graph}
\bibfield{author}{\bibinfo{person}{Deng Cai}, \bibinfo{person}{Xiaofei He},
  \bibinfo{person}{Jiawei Han}, {and} \bibinfo{person}{Thomas~S Huang}.}
  \bibinfo{year}{2011}\natexlab{}.
\newblock \showarticletitle{Graph regularized nonnegative matrix factorization
  for data representation}.
\newblock \bibinfo{journal}{{\em IEEE Transactions on Pattern Analysis and
  Machine Intelligence\/}} (\bibinfo{year}{2011}).
\newblock


\bibitem[\protect\citeauthoryear{Cai, Cand{\`e}s, and Shen}{Cai
  et~al\mbox{.}}{2010}]%
        {cai2010singular}
\bibfield{author}{\bibinfo{person}{Jian-Feng Cai}, \bibinfo{person}{Emmanuel~J
  Cand{\`e}s}, {and} \bibinfo{person}{Zuowei Shen}.}
  \bibinfo{year}{2010}\natexlab{}.
\newblock \showarticletitle{A singular value thresholding algorithm for matrix
  completion}.
\newblock \bibinfo{journal}{{\em SIOPT\/}} (\bibinfo{year}{2010}).
\newblock


\bibitem[\protect\citeauthoryear{Caiafa and Cichocki}{Caiafa and
  Cichocki}{2012}]%
        {caiafa2012block}
\bibfield{author}{\bibinfo{person}{Cesar~F Caiafa} {and}
  \bibinfo{person}{Andrzej Cichocki}.} \bibinfo{year}{2012}\natexlab{}.
\newblock \showarticletitle{Block sparse representations of tensors using
  Kronecker bases}.
\newblock \bibinfo{journal}{{\em IEEE ICASSP\/}} (\bibinfo{year}{2012}).
\newblock


\bibitem[\protect\citeauthoryear{Caiafa and Cichocki}{Caiafa and
  Cichocki}{2013}]%
        {caiafa2013computing}
\bibfield{author}{\bibinfo{person}{Cesar~F Caiafa} {and}
  \bibinfo{person}{Andrzej Cichocki}.} \bibinfo{year}{2013}\natexlab{}.
\newblock \showarticletitle{Computing sparse representations of
  multidimensional signals using kronecker bases}.
\newblock \bibinfo{journal}{{\em Neural computation\/}} (\bibinfo{year}{2013}).
\newblock


\bibitem[\protect\citeauthoryear{Caiafa and Cichocki}{Caiafa and
  Cichocki}{2015}]%
        {caiafa2015stable}
\bibfield{author}{\bibinfo{person}{Cesar~F Caiafa} {and}
  \bibinfo{person}{Andrzej Cichocki}.} \bibinfo{year}{2015}\natexlab{}.
\newblock \showarticletitle{Stable, robust, and super fast reconstruction of
  tensors using multi-way projections}.
\newblock \bibinfo{journal}{{\em IEEE Transactions on Signal Processing\/}}
  (\bibinfo{year}{2015}).
\newblock


\bibitem[\protect\citeauthoryear{Candes and Recht}{Candes and Recht}{2012}]%
        {candes2012exact}
\bibfield{author}{\bibinfo{person}{Emmanuel Candes} {and}
  \bibinfo{person}{Benjamin Recht}.} \bibinfo{year}{2012}\natexlab{}.
\newblock \showarticletitle{Exact matrix completion via convex optimization}.
\newblock \bibinfo{journal}{{\it Commun. ACM}} (\bibinfo{year}{2012}).
\newblock


\bibitem[\protect\citeauthoryear{Cand{\`e}s, Li, Ma, and Wright}{Cand{\`e}s
  et~al\mbox{.}}{2011}]%
        {candes2011robust}
\bibfield{author}{\bibinfo{person}{Emmanuel~J Cand{\`e}s},
  \bibinfo{person}{Xiaodong Li}, \bibinfo{person}{Yi Ma}, {and}
  \bibinfo{person}{John Wright}.} \bibinfo{year}{2011}\natexlab{}.
\newblock \showarticletitle{Robust principal component analysis?}
\newblock \bibinfo{journal}{{\em Journal of the ACM (JACM)\/}}
  (\bibinfo{year}{2011}).
\newblock


\bibitem[\protect\citeauthoryear{Candes and Plan}{Candes and Plan}{2010}]%
        {candes2010matrix}
\bibfield{author}{\bibinfo{person}{Emmanuel~J Candes} {and}
  \bibinfo{person}{Yaniv Plan}.} \bibinfo{year}{2010}\natexlab{}.
\newblock \showarticletitle{Matrix completion with noise}.
\newblock \bibinfo{journal}{{\it Proc. IEEE}} (\bibinfo{year}{2010}).
\newblock


\bibitem[\protect\citeauthoryear{Cand{\`e}s, Romberg, and Tao}{Cand{\`e}s
  et~al\mbox{.}}{2006}]%
        {candes2006robust}
\bibfield{author}{\bibinfo{person}{Emmanuel~J Cand{\`e}s},
  \bibinfo{person}{Justin Romberg}, {and} \bibinfo{person}{Terence Tao}.}
  \bibinfo{year}{2006}\natexlab{}.
\newblock \showarticletitle{Robust uncertainty principles: Exact signal
  reconstruction from highly incomplete frequency information}.
\newblock \bibinfo{journal}{{\em IEEE Transactions on information theory\/}}
  (\bibinfo{year}{2006}).
\newblock


\bibitem[\protect\citeauthoryear{Carroll and Chang}{Carroll and Chang}{1970}]%
        {carroll1970analysis}
\bibfield{author}{\bibinfo{person}{J~Douglas Carroll} {and}
  \bibinfo{person}{Jih-Jie Chang}.} \bibinfo{year}{1970}\natexlab{}.
\newblock \showarticletitle{Analysis of individual differences in
  multidimensional scaling via an N-way generalization of ``Eckart-Young''
  decomposition}.
\newblock \bibinfo{journal}{{\em Psychometrika\/}} (\bibinfo{year}{1970}).
\newblock


\bibitem[\protect\citeauthoryear{Carroll, Pruzansky, and Kruskal}{Carroll
  et~al\mbox{.}}{1980}]%
        {carroll1980candelinc}
\bibfield{author}{\bibinfo{person}{J~Douglas Carroll}, \bibinfo{person}{Sandra
  Pruzansky}, {and} \bibinfo{person}{Joseph~B Kruskal}.}
  \bibinfo{year}{1980}\natexlab{}.
\newblock \showarticletitle{CANDELINC: A general approach to multidimensional
  analysis of many-way arrays with linear constraints on parameters}.
\newblock \bibinfo{journal}{{\em Psychometrika\/}} (\bibinfo{year}{1980}).
\newblock


\bibitem[\protect\citeauthoryear{Chatfield and Yar}{Chatfield and Yar}{1988}]%
        {chatfield1988holt}
\bibfield{author}{\bibinfo{person}{Chris Chatfield} {and}
  \bibinfo{person}{Mohammad Yar}.} \bibinfo{year}{1988}\natexlab{}.
\newblock \showarticletitle{Holt-Winters forecasting: some practical issues}.
\newblock \bibinfo{journal}{{\em The Statistician\/}} (\bibinfo{year}{1988}).
\newblock


\bibitem[\protect\citeauthoryear{Chen, Hsu, and Liao}{Chen
  et~al\mbox{.}}{2014}]%
        {Chen:Tensor}
\bibfield{author}{\bibinfo{person}{Yi-Lei Chen}, \bibinfo{person}{Chiou-Ting
  Hsu}, {and} \bibinfo{person}{Hong-Yuan~Mark Liao}.}
  \bibinfo{year}{2014}\natexlab{}.
\newblock \showarticletitle{Simultaneous tensor decomposition and completion
  using factor priors}.
\newblock \bibinfo{journal}{{\em TPAMI\/}} (\bibinfo{year}{2014}).
\newblock


\bibitem[\protect\citeauthoryear{Cheng, Yu, Zhang, Xing, and Schuurmans}{Cheng
  et~al\mbox{.}}{2016}]%
        {cheng2016scalable}
\bibfield{author}{\bibinfo{person}{Hao Cheng}, \bibinfo{person}{Yaoliang Yu},
  \bibinfo{person}{Xinhua Zhang}, \bibinfo{person}{Eric Xing}, {and}
  \bibinfo{person}{Dale Schuurmans}.} \bibinfo{year}{2016}\natexlab{}.
\newblock \showarticletitle{Scalable and sound low-rank tensor learning}. In
  \bibinfo{booktitle}{{\em Artificial Intelligence and Statistics}}.
\newblock


\bibitem[\protect\citeauthoryear{Choi and Vishwanathan}{Choi and
  Vishwanathan}{2014}]%
        {choi2014dfacto}
\bibfield{author}{\bibinfo{person}{Joon~Hee Choi} {and} \bibinfo{person}{S
  Vishwanathan}.} \bibinfo{year}{2014}\natexlab{}.
\newblock \showarticletitle{DFacTo: Distributed factorization of tensors}. In
  \bibinfo{booktitle}{{\em Advances in Neural Information Processing Systems}}.
  \bibinfo{pages}{1296--1304}.
\newblock


\bibitem[\protect\citeauthoryear{Chu and Ghahramani}{Chu and
  Ghahramani}{2009}]%
        {chu2009probabilistic}
\bibfield{author}{\bibinfo{person}{Wei Chu} {and} \bibinfo{person}{Zoubin
  Ghahramani}.} \bibinfo{year}{2009}\natexlab{}.
\newblock \showarticletitle{Probabilistic Models for Incomplete
  Multi-dimensional Arrays.}
\newblock \bibinfo{journal}{{\em AISTATS\/}} (\bibinfo{year}{2009}).
\newblock


\bibitem[\protect\citeauthoryear{Cichocki, Zdunek, Phan, and Amari}{Cichocki
  et~al\mbox{.}}{2009}]%
        {cichocki2009nonnegative}
\bibfield{author}{\bibinfo{person}{Andrzej Cichocki}, \bibinfo{person}{Rafal
  Zdunek}, \bibinfo{person}{Anh~Huy Phan}, {and} \bibinfo{person}{Shun-ichi
  Amari}.} \bibinfo{year}{2009}\natexlab{}.
\newblock \bibinfo{booktitle}{{\em Nonnegative matrix and tensor
  factorizations: applications to exploratory multi-way data analysis and blind
  source separation}}.
\newblock \bibinfo{publisher}{John Wiley \&amp; Sons}.
\newblock


\bibitem[\protect\citeauthoryear{Comon}{Comon}{2004}]%
        {comon2004canonical}
\bibfield{author}{\bibinfo{person}{Pierre Comon}.}
  \bibinfo{year}{2004}\natexlab{}.
\newblock \showarticletitle{Canonical tensor decompositions}.
\newblock \bibinfo{journal}{{\em Workshop on Tensor Decompositions, Palo Alto,
  CA\/}} (\bibinfo{year}{2004}).
\newblock


\bibitem[\protect\citeauthoryear{Da~Silva and Herrmann}{Da~Silva and
  Herrmann}{2013}]%
        {da2013hierarchical}
\bibfield{author}{\bibinfo{person}{Curt Da~Silva} {and}
  \bibinfo{person}{Felix~J Herrmann}.} \bibinfo{year}{2013}\natexlab{}.
\newblock \showarticletitle{Hierarchical Tucker tensor
  optimization-Applications to tensor completion}.
\newblock \bibinfo{journal}{{\em International Conference on Sampling Theory
  and Applications\/}} (\bibinfo{year}{2013}).
\newblock


\bibitem[\protect\citeauthoryear{Da~Silva and Herrmann}{Da~Silva and
  Herrmann}{2015}]%
        {da2015optimization}
\bibfield{author}{\bibinfo{person}{Curt Da~Silva} {and}
  \bibinfo{person}{Felix~J Herrmann}.} \bibinfo{year}{2015}\natexlab{}.
\newblock \showarticletitle{Optimization on the Hierarchical Tucker
  manifold--Applications to tensor completion}.
\newblock \bibinfo{journal}{{\it Linear Algebra Appl.}} (\bibinfo{year}{2015}).
\newblock


\bibitem[\protect\citeauthoryear{Dauwels, Garg, Earnest, and Pang}{Dauwels
  et~al\mbox{.}}{2011}]%
        {dauwels2011handling}
\bibfield{author}{\bibinfo{person}{Justin Dauwels}, \bibinfo{person}{Lalit
  Garg}, \bibinfo{person}{Arul Earnest}, {and} \bibinfo{person}{Leong~Khai
  Pang}.} \bibinfo{year}{2011}\natexlab{}.
\newblock \showarticletitle{Handling missing data in medical questionnaires
  using tensor decompositions}.
\newblock \bibinfo{journal}{{\em ICICS\/}} (\bibinfo{year}{2011}).
\newblock


\bibitem[\protect\citeauthoryear{De~Lathauwer, De~Moor, and
  Vandewalle}{De~Lathauwer et~al\mbox{.}}{2000a}]%
        {de2000multilinear}
\bibfield{author}{\bibinfo{person}{Lieven De~Lathauwer}, \bibinfo{person}{Bart
  De~Moor}, {and} \bibinfo{person}{Joos Vandewalle}.}
  \bibinfo{year}{2000}\natexlab{a}.
\newblock \showarticletitle{A multilinear singular value decomposition}.
\newblock \bibinfo{journal}{{\em SIAM journal on Matrix Analysis and
  Applications\/}} (\bibinfo{year}{2000}).
\newblock


\bibitem[\protect\citeauthoryear{De~Lathauwer, De~Moor, and
  Vandewalle}{De~Lathauwer et~al\mbox{.}}{2000b}]%
        {de2000best}
\bibfield{author}{\bibinfo{person}{Lieven De~Lathauwer}, \bibinfo{person}{Bart
  De~Moor}, {and} \bibinfo{person}{Joos Vandewalle}.}
  \bibinfo{year}{2000}\natexlab{b}.
\newblock \showarticletitle{On the best rank-1 and rank-(r 1, r 2,..., rn)
  approximation of higher-order tensors}.
\newblock \bibinfo{journal}{{\it SIAM J. Matrix Anal. Appl.}}
  (\bibinfo{year}{2000}).
\newblock


\bibitem[\protect\citeauthoryear{Dhillon and Sra}{Dhillon and Sra}{2005}]%
        {dhillon2005generalized}
\bibfield{author}{\bibinfo{person}{Inderjit~S Dhillon} {and}
  \bibinfo{person}{Suvrit Sra}.} \bibinfo{year}{2005}\natexlab{}.
\newblock \showarticletitle{Generalized nonnegative matrix approximations with
  Bregman divergences}.
\newblock \bibinfo{journal}{{\em NIPS\/}} (\bibinfo{year}{2005}).
\newblock


\bibitem[\protect\citeauthoryear{Donoho}{Donoho}{2006}]%
        {donoho2006compressed}
\bibfield{author}{\bibinfo{person}{David~L Donoho}.}
  \bibinfo{year}{2006}\natexlab{}.
\newblock \showarticletitle{Compressed sensing}.
\newblock \bibinfo{journal}{{\em IEEE Transactions on information theory\/}}
  (\bibinfo{year}{2006}).
\newblock


\bibitem[\protect\citeauthoryear{Duarte and Baraniuk}{Duarte and
  Baraniuk}{2012}]%
        {duarte2012kronecker}
\bibfield{author}{\bibinfo{person}{Marco~F Duarte} {and}
  \bibinfo{person}{Richard~G Baraniuk}.} \bibinfo{year}{2012}\natexlab{}.
\newblock \showarticletitle{Kronecker compressive sensing}.
\newblock \bibinfo{journal}{{\em IEEE Transactions on Image Processing\/}}
  (\bibinfo{year}{2012}).
\newblock


\bibitem[\protect\citeauthoryear{Dunlavy, Kolda, and Acar}{Dunlavy
  et~al\mbox{.}}{2011}]%
        {dunlavy2011temporal}
\bibfield{author}{\bibinfo{person}{Daniel~M Dunlavy}, \bibinfo{person}{Tamara~G
  Kolda}, {and} \bibinfo{person}{Evrim Acar}.} \bibinfo{year}{2011}\natexlab{}.
\newblock \showarticletitle{Temporal link prediction using matrix and tensor
  factorizations}.
\newblock \bibinfo{journal}{{\em TKDD\/}} (\bibinfo{year}{2011}).
\newblock


\bibitem[\protect\citeauthoryear{Eckstein and Bertsekas}{Eckstein and
  Bertsekas}{1992}]%
        {eckstein1992douglas}
\bibfield{author}{\bibinfo{person}{Jonathan Eckstein} {and}
  \bibinfo{person}{Dimitri~P Bertsekas}.} \bibinfo{year}{1992}\natexlab{}.
\newblock \showarticletitle{On the Douglas---Rachford splitting method and the
  proximal point algorithm for maximal monotone operators}.
\newblock \bibinfo{journal}{{\em Mathematical Programming\/}}
  (\bibinfo{year}{1992}).
\newblock


\bibitem[\protect\citeauthoryear{Eldar and Kutyniok}{Eldar and
  Kutyniok}{2012}]%
        {eldar2012compressed}
\bibfield{author}{\bibinfo{person}{Yonina~C Eldar} {and} \bibinfo{person}{Gitta
  Kutyniok}.} \bibinfo{year}{2012}\natexlab{}.
\newblock \bibinfo{booktitle}{{\em Compressed sensing: theory and
  applications}}.
\newblock \bibinfo{publisher}{Cambridge University Press}.
\newblock


\bibitem[\protect\citeauthoryear{Eldar, Needell, and Plan}{Eldar
  et~al\mbox{.}}{2012}]%
        {eldar2012uniqueness}
\bibfield{author}{\bibinfo{person}{Yonina~C Eldar}, \bibinfo{person}{Deanna
  Needell}, {and} \bibinfo{person}{Yaniv Plan}.}
  \bibinfo{year}{2012}\natexlab{}.
\newblock \showarticletitle{Uniqueness conditions for low-rank matrix
  recovery}.
\newblock \bibinfo{journal}{{\em Applied and Computational Harmonic
  Analysis\/}} (\bibinfo{year}{2012}).
\newblock


\bibitem[\protect\citeauthoryear{Ermi{\c{s}}, Acar, and Cemgil}{Ermi{\c{s}}
  et~al\mbox{.}}{2012}]%
        {ermics2012link}
\bibfield{author}{\bibinfo{person}{Beyza Ermi{\c{s}}}, \bibinfo{person}{Evrim
  Acar}, {and} \bibinfo{person}{A~Taylan Cemgil}.}
  \bibinfo{year}{2012}\natexlab{}.
\newblock \showarticletitle{Link prediction via generalized coupled tensor
  factorisation}.
\newblock \bibinfo{journal}{{\em arXiv preprint arXiv:1208.6231\/}}
  (\bibinfo{year}{2012}).
\newblock


\bibitem[\protect\citeauthoryear{Ermi{\c{s}}, Acar, and Cemgil}{Ermi{\c{s}}
  et~al\mbox{.}}{2015}]%
        {ermics2015link}
\bibfield{author}{\bibinfo{person}{Beyza Ermi{\c{s}}}, \bibinfo{person}{Evrim
  Acar}, {and} \bibinfo{person}{A~Taylan Cemgil}.}
  \bibinfo{year}{2015}\natexlab{}.
\newblock \showarticletitle{Link prediction in heterogeneous data via
  generalized coupled tensor factorization}.
\newblock \bibinfo{journal}{{\em Data Mining and Knowledge Discovery\/}}
  (\bibinfo{year}{2015}).
\newblock


\bibitem[\protect\citeauthoryear{Faber, Bro, and Hopke}{Faber
  et~al\mbox{.}}{2003}]%
        {faber2003recent}
\bibfield{author}{\bibinfo{person}{Nicolaas Klaas~M Faber},
  \bibinfo{person}{Rasmus Bro}, {and} \bibinfo{person}{Philip~K Hopke}.}
  \bibinfo{year}{2003}\natexlab{}.
\newblock \showarticletitle{Recent developments in CANDECOMP/PARAFAC
  algorithms: a critical review}.
\newblock \bibinfo{journal}{{\em Chemometrics and Intelligent Laboratory
  Systems\/}} (\bibinfo{year}{2003}).
\newblock


\bibitem[\protect\citeauthoryear{Fanaee-T and Gama}{Fanaee-T and Gama}{2015}]%
        {fanaee2015multi}
\bibfield{author}{\bibinfo{person}{Hadi Fanaee-T} {and}
  \bibinfo{person}{Jo{\~a}o Gama}.} \bibinfo{year}{2015}\natexlab{}.
\newblock \showarticletitle{Multi-aspect-streaming tensor analysis}.
\newblock \bibinfo{journal}{{\em Knowledge-Based Systems\/}}
  (\bibinfo{year}{2015}).
\newblock


\bibitem[\protect\citeauthoryear{Feng, Xu, and Yan}{Feng et~al\mbox{.}}{2013}]%
        {feng2013online}
\bibfield{author}{\bibinfo{person}{Jiashi Feng}, \bibinfo{person}{Huan Xu},
  {and} \bibinfo{person}{Shuicheng Yan}.} \bibinfo{year}{2013}\natexlab{}.
\newblock \showarticletitle{Online robust pca via stochastic optimization}. In
  \bibinfo{booktitle}{{\em NIPS}}.
\newblock


\bibitem[\protect\citeauthoryear{Filipovi{\'c} and Juki{\'c}}{Filipovi{\'c} and
  Juki{\'c}}{2015}]%
        {filipovic2015tucker}
\bibfield{author}{\bibinfo{person}{Marko Filipovi{\'c}} {and}
  \bibinfo{person}{Ante Juki{\'c}}.} \bibinfo{year}{2015}\natexlab{}.
\newblock \showarticletitle{Tucker factorization with missing data with
  application to low-n-rank tensor completion}.
\newblock \bibinfo{journal}{{\em Multidimensional systems and signal
  processing\/}} (\bibinfo{year}{2015}).
\newblock


\bibitem[\protect\citeauthoryear{Foster, Nascimento, and Amano}{Foster
  et~al\mbox{.}}{2004}]%
        {foster2004information}
\bibfield{author}{\bibinfo{person}{David~H Foster},
  \bibinfo{person}{S{\'e}rgio~MC Nascimento}, {and} \bibinfo{person}{Kinjiro
  Amano}.} \bibinfo{year}{2004}\natexlab{}.
\newblock \showarticletitle{Information limits on neural identification of
  colored surfaces in natural scenes}.
\newblock \bibinfo{journal}{{\em Visual neuroscience\/}}
  (\bibinfo{year}{2004}).
\newblock


\bibitem[\protect\citeauthoryear{Friedland, Li, and Schonfeld}{Friedland
  et~al\mbox{.}}{2014}]%
        {friedland2014compressive}
\bibfield{author}{\bibinfo{person}{Shmuel Friedland}, \bibinfo{person}{Qun Li},
  {and} \bibinfo{person}{Dan Schonfeld}.} \bibinfo{year}{2014}\natexlab{}.
\newblock \showarticletitle{Compressive sensing of sparse tensors}.
\newblock \bibinfo{journal}{{\em IEEE Transactions on Image Processing\/}}
  (\bibinfo{year}{2014}).
\newblock


\bibitem[\protect\citeauthoryear{Gandy, Recht, and Yamada}{Gandy
  et~al\mbox{.}}{2011}]%
        {gandy2011tensor}
\bibfield{author}{\bibinfo{person}{Silvia Gandy}, \bibinfo{person}{Benjamin
  Recht}, {and} \bibinfo{person}{Isao Yamada}.}
  \bibinfo{year}{2011}\natexlab{}.
\newblock \showarticletitle{Tensor completion and low-n-rank tensor recovery
  via convex optimization}.
\newblock \bibinfo{journal}{{\em Inverse Problems\/}} (\bibinfo{year}{2011}).
\newblock


\bibitem[\protect\citeauthoryear{Ge, Caverlee, and Lu}{Ge
  et~al\mbox{.}}{2016a}]%
        {ge2016taper}
\bibfield{author}{\bibinfo{person}{Hancheng Ge}, \bibinfo{person}{James
  Caverlee}, {and} \bibinfo{person}{Haokai Lu}.}
  \bibinfo{year}{2016}\natexlab{a}.
\newblock \showarticletitle{Taper: a contextual tensor-based approach for
  personalized expert recommendation}.
\newblock \bibinfo{journal}{{\em Proc. of RecSys\/}} (\bibinfo{year}{2016}).
\newblock


\bibitem[\protect\citeauthoryear{Ge, Caverlee, Zhang, and Squicciarini}{Ge
  et~al\mbox{.}}{2016b}]%
        {Ge:Tensor}
\bibfield{author}{\bibinfo{person}{Hancheng Ge}, \bibinfo{person}{James
  Caverlee}, \bibinfo{person}{Nan Zhang}, {and} \bibinfo{person}{Anna
  Squicciarini}.} \bibinfo{year}{2016}\natexlab{b}.
\newblock \showarticletitle{Uncovering the spatio-temporal dynamics of memes in
  the presence of incomplete information}.
\newblock \bibinfo{journal}{{\em CIKM\/}} (\bibinfo{year}{2016}).
\newblock


\bibitem[\protect\citeauthoryear{Ge, Zhang, Alfifi, Hu, and Caverlee}{Ge
  et~al\mbox{.}}{2018}]%
        {ge2018distenc}
\bibfield{author}{\bibinfo{person}{Hancheng Ge}, \bibinfo{person}{Kai Zhang},
  \bibinfo{person}{Majid Alfifi}, \bibinfo{person}{Xia Hu}, {and}
  \bibinfo{person}{James Caverlee}.} \bibinfo{year}{2018}\natexlab{}.
\newblock \showarticletitle{DisTenC: A Distributed Algorithm for Scalable
  Tensor Completion on Spark}. In \bibinfo{booktitle}{{\em ICDE}}.
\newblock


\bibitem[\protect\citeauthoryear{Gemulla, Nijkamp, Haas, and Sismanis}{Gemulla
  et~al\mbox{.}}{2011}]%
        {gemulla2011large}
\bibfield{author}{\bibinfo{person}{Rainer Gemulla}, \bibinfo{person}{Erik
  Nijkamp}, \bibinfo{person}{Peter~J Haas}, {and} \bibinfo{person}{Yannis
  Sismanis}.} \bibinfo{year}{2011}\natexlab{}.
\newblock \showarticletitle{Large-scale matrix factorization with distributed
  stochastic gradient descent}. In \bibinfo{booktitle}{{\em Proceedings of the
  17th ACM SIGKDD international conference on Knowledge discovery and data
  mining}}. ACM, \bibinfo{pages}{69--77}.
\newblock


\bibitem[\protect\citeauthoryear{Geng and Smith-Miles}{Geng and
  Smith-Miles}{2009}]%
        {geng2009facial}
\bibfield{author}{\bibinfo{person}{Xin Geng} {and} \bibinfo{person}{Kate
  Smith-Miles}.} \bibinfo{year}{2009}\natexlab{}.
\newblock \showarticletitle{Facial age estimation by multilinear subspace
  analysis}.
\newblock \bibinfo{journal}{{\em ICASSP\/}} (\bibinfo{year}{2009}).
\newblock


\bibitem[\protect\citeauthoryear{Geng, Smith-Miles, Zhou, and Wang}{Geng
  et~al\mbox{.}}{2011}]%
        {geng2011face}
\bibfield{author}{\bibinfo{person}{Xin Geng}, \bibinfo{person}{Kate
  Smith-Miles}, \bibinfo{person}{Zhi-Hua Zhou}, {and} \bibinfo{person}{Liang
  Wang}.} \bibinfo{year}{2011}\natexlab{}.
\newblock \showarticletitle{Face image modeling by multilinear subspace
  analysis with missing values}.
\newblock \bibinfo{journal}{{\em IEEE Transactions on Systems, Man, and
  Cybernetics, Part B (Cybernetics)\/}} (\bibinfo{year}{2011}).
\newblock


\bibitem[\protect\citeauthoryear{Ghadermarzy, Plan, and Y{\i}lmaz}{Ghadermarzy
  et~al\mbox{.}}{2017}]%
        {ghadermarzy2017near}
\bibfield{author}{\bibinfo{person}{Navid Ghadermarzy}, \bibinfo{person}{Yaniv
  Plan}, {and} \bibinfo{person}{{\"O}zg{\"u}r Y{\i}lmaz}.}
  \bibinfo{year}{2017}\natexlab{}.
\newblock \showarticletitle{Near-optimal sample complexity for convex tensor
  completion}.
\newblock \bibinfo{journal}{{\em arXiv preprint arXiv:1711.04965\/}}
  (\bibinfo{year}{2017}).
\newblock


\bibitem[\protect\citeauthoryear{Goldfarb and Qin}{Goldfarb and Qin}{2014}]%
        {goldfarb2014robust}
\bibfield{author}{\bibinfo{person}{Donald Goldfarb} {and}
  \bibinfo{person}{Zhiwei Qin}.} \bibinfo{year}{2014}\natexlab{}.
\newblock \showarticletitle{Robust low-rank tensor recovery: Models and
  algorithms}.
\newblock \bibinfo{journal}{{\it SIAM J. Matrix Anal. Appl.}}
  (\bibinfo{year}{2014}).
\newblock


\bibitem[\protect\citeauthoryear{Grasedyck, Kluge, and Kr{\"a}mer}{Grasedyck
  et~al\mbox{.}}{2015}]%
        {grasedyck2015alternating}
\bibfield{author}{\bibinfo{person}{Lars Grasedyck}, \bibinfo{person}{Melanie
  Kluge}, {and} \bibinfo{person}{Sebastian Kr{\"a}mer}.}
  \bibinfo{year}{2015}\natexlab{}.
\newblock \showarticletitle{Alternating least squares tensor completion in the
  TT-format}.
\newblock \bibinfo{journal}{{\em arXiv preprint arXiv:1509.00311\/}}
  (\bibinfo{year}{2015}).
\newblock


\bibitem[\protect\citeauthoryear{Grasedyck, Kressner, and Tobler}{Grasedyck
  et~al\mbox{.}}{2013}]%
        {grasedyck2013literature}
\bibfield{author}{\bibinfo{person}{Lars Grasedyck}, \bibinfo{person}{Daniel
  Kressner}, {and} \bibinfo{person}{Christine Tobler}.}
  \bibinfo{year}{2013}\natexlab{}.
\newblock \showarticletitle{A literature survey of low-rank tensor
  approximation techniques}.
\newblock \bibinfo{journal}{{\em GAMM-Mitteilungen\/}} (\bibinfo{year}{2013}).
\newblock


\bibitem[\protect\citeauthoryear{Gross}{Gross}{2011}]%
        {gross2011recovering}
\bibfield{author}{\bibinfo{person}{David Gross}.}
  \bibinfo{year}{2011}\natexlab{}.
\newblock \showarticletitle{Recovering low-rank matrices from few coefficients
  in any basis}.
\newblock \bibinfo{journal}{{\em IEEE Transactions on Information Theory\/}}
  (\bibinfo{year}{2011}).
\newblock


\bibitem[\protect\citeauthoryear{Guo, Yao, and Kwok}{Guo et~al\mbox{.}}{2017}]%
        {guo2017efficient}
\bibfield{author}{\bibinfo{person}{Xiawei Guo}, \bibinfo{person}{Quanming Yao},
  {and} \bibinfo{person}{James Tin-Yau Kwok}.} \bibinfo{year}{2017}\natexlab{}.
\newblock \showarticletitle{Efficient Sparse Low-Rank Tensor Completion Using
  the Frank-Wolfe Algorithm.}. In \bibinfo{booktitle}{{\em AAAI}}.
\newblock


\bibitem[\protect\citeauthoryear{Hackbusch}{Hackbusch}{2012}]%
        {hackbusch2012tensor}
\bibfield{author}{\bibinfo{person}{Wolfgang Hackbusch}.}
  \bibinfo{year}{2012}\natexlab{}.
\newblock \bibinfo{booktitle}{{\em Tensor spaces and numerical tensor
  calculus}}.
\newblock \bibinfo{publisher}{Springer Science \& Business Media}.
\newblock


\bibitem[\protect\citeauthoryear{Harchaoui, Juditsky, and Nemirovski}{Harchaoui
  et~al\mbox{.}}{2015}]%
        {harchaoui2015conditional}
\bibfield{author}{\bibinfo{person}{Zaid Harchaoui}, \bibinfo{person}{Anatoli
  Juditsky}, {and} \bibinfo{person}{Arkadi Nemirovski}.}
  \bibinfo{year}{2015}\natexlab{}.
\newblock \showarticletitle{Conditional gradient algorithms for
  norm-regularized smooth convex optimization}.
\newblock \bibinfo{journal}{{\em Mathematical Programming\/}}
  (\bibinfo{year}{2015}).
\newblock


\bibitem[\protect\citeauthoryear{Harshman}{Harshman}{1970}]%
        {harshman1970foundations}
\bibfield{author}{\bibinfo{person}{Richard~A Harshman}.}
  \bibinfo{year}{1970}\natexlab{}.
\newblock \showarticletitle{Foundations of the PARAFAC procedure: Models and
  conditions for an" explanatory" multi-modal factor analysis}.
\newblock \bibinfo{journal}{{\em University of California at Los Angeles Los
  Angeles, CA\/}} (\bibinfo{year}{1970}).
\newblock


\bibitem[\protect\citeauthoryear{H{\aa}stad}{H{\aa}stad}{1990}]%
        {haastad1990tensor}
\bibfield{author}{\bibinfo{person}{Johan H{\aa}stad}.}
  \bibinfo{year}{1990}\natexlab{}.
\newblock \showarticletitle{Tensor rank is NP-complete}.
\newblock \bibinfo{journal}{{\em Journal of Algorithms\/}}
  (\bibinfo{year}{1990}).
\newblock


\bibitem[\protect\citeauthoryear{Henrion}{Henrion}{1994}]%
        {henrion1994n}
\bibfield{author}{\bibinfo{person}{Ren{\'e} Henrion}.}
  \bibinfo{year}{1994}\natexlab{}.
\newblock \showarticletitle{N-way principal component analysis theory,
  algorithms and applications}.
\newblock \bibinfo{journal}{{\em Chemometrics and intelligent laboratory
  systems\/}} (\bibinfo{year}{1994}).
\newblock


\bibitem[\protect\citeauthoryear{Hernandez}{Hernandez}{2010}]%
        {hernandez2010simple}
\bibfield{author}{\bibinfo{person}{David Hernandez}.}
  \bibinfo{year}{2010}\natexlab{}.
\newblock \showarticletitle{Simple tensor products}.
\newblock \bibinfo{journal}{{\em Inventiones mathematicae\/}}
  (\bibinfo{year}{2010}).
\newblock


\bibitem[\protect\citeauthoryear{Hitchcock}{Hitchcock}{[n. d.]}]%
        {Hitchcock:2}
\bibfield{author}{\bibinfo{person}{Frank~L Hitchcock}.} \bibinfo{year}{[n.
  d.]}\natexlab{}.
\newblock \showarticletitle{Multiple invariants and generalized rank of a p-way
  matrix or tensor}.
\newblock \bibinfo{journal}{{\em Journal of Mathematics and Physics\/}}
  (\bibinfo{year}{[n. d.]}).
\newblock


\bibitem[\protect\citeauthoryear{Hitchcock}{Hitchcock}{1927}]%
        {Hitchcock:1}
\bibfield{author}{\bibinfo{person}{Frank~L Hitchcock}.}
  \bibinfo{year}{1927}\natexlab{}.
\newblock \showarticletitle{The expression of a tensor or a polyadic as a sum
  of products}.
\newblock \bibinfo{journal}{{\em Journal of Mathematics and Physics\/}}
  (\bibinfo{year}{1927}).
\newblock


\bibitem[\protect\citeauthoryear{Ho, Ghosh, Steinhubl, Stewart, Denny, Malin,
  and Sun}{Ho et~al\mbox{.}}{2014}]%
        {ho2014limestone}
\bibfield{author}{\bibinfo{person}{Joyce~C Ho}, \bibinfo{person}{Joydeep
  Ghosh}, \bibinfo{person}{Steve~R Steinhubl}, \bibinfo{person}{Walter~F
  Stewart}, \bibinfo{person}{Joshua~C Denny}, \bibinfo{person}{Bradley~A
  Malin}, {and} \bibinfo{person}{Jimeng Sun}.} \bibinfo{year}{2014}\natexlab{}.
\newblock \showarticletitle{Limestone: High-throughput candidate phenotype
  generation via tensor factorization}.
\newblock \bibinfo{journal}{{\em Journal of biomedical informatics\/}}
  (\bibinfo{year}{2014}).
\newblock


\bibitem[\protect\citeauthoryear{Hu, Li, Zhang, Shi, Maybank, and Zhang}{Hu
  et~al\mbox{.}}{2011}]%
        {hu2011incremental}
\bibfield{author}{\bibinfo{person}{Weiming Hu}, \bibinfo{person}{Xi Li},
  \bibinfo{person}{Xiaoqin Zhang}, \bibinfo{person}{Xinchu Shi},
  \bibinfo{person}{Stephen Maybank}, {and} \bibinfo{person}{Zhongfei Zhang}.}
  \bibinfo{year}{2011}\natexlab{}.
\newblock \showarticletitle{Incremental tensor subspace learning and its
  applications to foreground segmentation and tracking}.
\newblock \bibinfo{journal}{{\em IJCV\/}} (\bibinfo{year}{2011}).
\newblock


\bibitem[\protect\citeauthoryear{Huang, Mu, Goldfarb, and Wright}{Huang
  et~al\mbox{.}}{2014}]%
        {huang2014provable}
\bibfield{author}{\bibinfo{person}{Bo Huang}, \bibinfo{person}{Cun Mu},
  \bibinfo{person}{Donald Goldfarb}, {and} \bibinfo{person}{John Wright}.}
  \bibinfo{year}{2014}\natexlab{}.
\newblock \showarticletitle{Provable low-rank tensor recovery}.
\newblock \bibinfo{journal}{{\em Optimization-Online\/}}
  (\bibinfo{year}{2014}).
\newblock


\bibitem[\protect\citeauthoryear{Huang, Niranjan, Hakeem, and Anandkumar}{Huang
  et~al\mbox{.}}{2013}]%
        {huang2013fast}
\bibfield{author}{\bibinfo{person}{Furong Huang}, \bibinfo{person}{UN
  Niranjan}, \bibinfo{person}{M Hakeem}, {and} \bibinfo{person}{Animashree
  Anandkumar}.} \bibinfo{year}{2013}\natexlab{}.
\newblock \showarticletitle{Fast detection of overlapping communities via
  online tensor methods}.
\newblock \bibinfo{journal}{{\em arXiv preprint arXiv:1309.0787\/}}
  (\bibinfo{year}{2013}).
\newblock


\bibitem[\protect\citeauthoryear{Imaizumi, Maehara, and Hayashi}{Imaizumi
  et~al\mbox{.}}{2017}]%
        {imaizumi2017tensor}
\bibfield{author}{\bibinfo{person}{Masaaki Imaizumi}, \bibinfo{person}{Takanori
  Maehara}, {and} \bibinfo{person}{Kohei Hayashi}.}
  \bibinfo{year}{2017}\natexlab{}.
\newblock \showarticletitle{On Tensor Train Rank Minimization: Statistical
  Efficiency and Scalable Algorithm}. In \bibinfo{booktitle}{{\em NIPS}}.
\newblock


\bibitem[\protect\citeauthoryear{Jain and Oh}{Jain and Oh}{2014}]%
        {jain2014provable}
\bibfield{author}{\bibinfo{person}{Prateek Jain} {and} \bibinfo{person}{Sewoong
  Oh}.} \bibinfo{year}{2014}\natexlab{}.
\newblock \showarticletitle{Provable tensor factorization with missing data}.
\newblock \bibinfo{journal}{{\em NIPS\/}} (\bibinfo{year}{2014}).
\newblock


\bibitem[\protect\citeauthoryear{Jain, Gutierrez, and Haupt}{Jain
  et~al\mbox{.}}{2017}]%
        {jain2017noisy}
\bibfield{author}{\bibinfo{person}{Swayambhoo Jain}, \bibinfo{person}{Alexander
  Gutierrez}, {and} \bibinfo{person}{Jarvis Haupt}.}
  \bibinfo{year}{2017}\natexlab{}.
\newblock \showarticletitle{Noisy Tensor Completion for Tensors with a Sparse
  Canonical Polyadic Factor}.
\newblock \bibinfo{journal}{{\em arXiv preprint arXiv:1704.02534\/}}
  (\bibinfo{year}{2017}).
\newblock


\bibitem[\protect\citeauthoryear{Javed, Bouwmans, and Jung}{Javed
  et~al\mbox{.}}{2015}]%
        {javed2015stochastic}
\bibfield{author}{\bibinfo{person}{Sajid Javed}, \bibinfo{person}{Thierry
  Bouwmans}, {and} \bibinfo{person}{Soon~Ki Jung}.}
  \bibinfo{year}{2015}\natexlab{}.
\newblock \showarticletitle{Stochastic decomposition into low rank and sparse
  tensor for robust background subtraction}.
\newblock \bibinfo{journal}{{\em IET\/}} (\bibinfo{year}{2015}).
\newblock


\bibitem[\protect\citeauthoryear{Jeon, Papalexakis, Kang, and Faloutsos}{Jeon
  et~al\mbox{.}}{2015}]%
        {jeon2015haten2}
\bibfield{author}{\bibinfo{person}{Inah Jeon}, \bibinfo{person}{Evangelos~E
  Papalexakis}, \bibinfo{person}{U Kang}, {and} \bibinfo{person}{Christos
  Faloutsos}.} \bibinfo{year}{2015}\natexlab{}.
\newblock \showarticletitle{Haten2: Billion-scale tensor decompositions}.
\newblock \bibinfo{journal}{{\em IEEE ICDE\/}} (\bibinfo{year}{2015}).
\newblock


\bibitem[\protect\citeauthoryear{Ji, Huang, Zhao, Ma, and Liu}{Ji
  et~al\mbox{.}}{2016}]%
        {ji2016tensor}
\bibfield{author}{\bibinfo{person}{Teng-Yu Ji}, \bibinfo{person}{Ting-Zhu
  Huang}, \bibinfo{person}{Xi-Le Zhao}, \bibinfo{person}{Tian-Hui Ma}, {and}
  \bibinfo{person}{Gang Liu}.} \bibinfo{year}{2016}\natexlab{}.
\newblock \showarticletitle{Tensor completion using total variation and
  low-rank matrix factorization}.
\newblock \bibinfo{journal}{{\em Information Sciences\/}}
  (\bibinfo{year}{2016}).
\newblock


\bibitem[\protect\citeauthoryear{Johnson}{Johnson}{1990}]%
        {johnson1990matrix}
\bibfield{author}{\bibinfo{person}{Charles~R Johnson}.}
  \bibinfo{year}{1990}\natexlab{}.
\newblock \showarticletitle{Matrix completion problems: a survey}.
\newblock \bibinfo{journal}{{\em Symposia in Applied Mathematics\/}}
  (\bibinfo{year}{1990}).
\newblock


\bibitem[\protect\citeauthoryear{Kang, Papalexakis, Harpale, and
  Faloutsos}{Kang et~al\mbox{.}}{2012}]%
        {kang2012gigatensor}
\bibfield{author}{\bibinfo{person}{U Kang}, \bibinfo{person}{Evangelos
  Papalexakis}, \bibinfo{person}{Abhay Harpale}, {and}
  \bibinfo{person}{Christos Faloutsos}.} \bibinfo{year}{2012}\natexlab{}.
\newblock \showarticletitle{Gigatensor: scaling tensor analysis up by 100
  times-algorithms and discoveries}.
\newblock \bibinfo{journal}{{\em SIGKDD\/}} (\bibinfo{year}{2012}).
\newblock


\bibitem[\protect\citeauthoryear{Karatzoglou, Amatriain, Baltrunas, and
  Oliver}{Karatzoglou et~al\mbox{.}}{2010}]%
        {karatzoglou2010multiverse}
\bibfield{author}{\bibinfo{person}{Alexandros Karatzoglou},
  \bibinfo{person}{Xavier Amatriain}, \bibinfo{person}{Linas Baltrunas}, {and}
  \bibinfo{person}{Nuria Oliver}.} \bibinfo{year}{2010}\natexlab{}.
\newblock \showarticletitle{Multiverse recommendation: n-dimensional tensor
  factorization for context-aware collaborative filtering}.
\newblock \bibinfo{journal}{{\em fourth ACM conference on Recommender
  systems\/}} (\bibinfo{year}{2010}).
\newblock


\bibitem[\protect\citeauthoryear{Karlsson, Kressner, and Uschmajew}{Karlsson
  et~al\mbox{.}}{2016}]%
        {karlsson2016parallel}
\bibfield{author}{\bibinfo{person}{Lars Karlsson}, \bibinfo{person}{Daniel
  Kressner}, {and} \bibinfo{person}{Andr{\'e} Uschmajew}.}
  \bibinfo{year}{2016}\natexlab{}.
\newblock \showarticletitle{Parallel algorithms for tensor completion in the CP
  format}.
\newblock \bibinfo{journal}{{\it Parallel Comput.}} (\bibinfo{year}{2016}).
\newblock


\bibitem[\protect\citeauthoryear{Kasai}{Kasai}{2016}]%
        {kasai2016online}
\bibfield{author}{\bibinfo{person}{Hiroyuki Kasai}.}
  \bibinfo{year}{2016}\natexlab{}.
\newblock \showarticletitle{Online low-rank tensor subspace tracking from
  incomplete data by CP decomposition using recursive least squares}.
\newblock \bibinfo{journal}{{\em ICASSP\/}} (\bibinfo{year}{2016}).
\newblock


\bibitem[\protect\citeauthoryear{Kasai and Mishra}{Kasai and Mishra}{2015}]%
        {kasai2015riemannian}
\bibfield{author}{\bibinfo{person}{Hiroyuki Kasai} {and}
  \bibinfo{person}{Bamdev Mishra}.} \bibinfo{year}{2015}\natexlab{}.
\newblock \showarticletitle{Riemannian preconditioning for tensor completion}.
\newblock \bibinfo{journal}{{\em arXiv preprint arXiv:1506.02159\/}}
  (\bibinfo{year}{2015}).
\newblock


\bibitem[\protect\citeauthoryear{Kasai and Mishra}{Kasai and Mishra}{2016}]%
        {kasai2016low}
\bibfield{author}{\bibinfo{person}{Hiroyuki Kasai} {and}
  \bibinfo{person}{Bamdev Mishra}.} \bibinfo{year}{2016}\natexlab{}.
\newblock \showarticletitle{Low-rank tensor completion: a Riemannian manifold
  preconditioning approach}.
\newblock \bibinfo{journal}{{\em Technical report, arXiv preprint\/}}
  (\bibinfo{year}{2016}).
\newblock


\bibitem[\protect\citeauthoryear{Kashima, Kato, Yamanishi, Sugiyama, and
  Tsuda}{Kashima et~al\mbox{.}}{2009}]%
        {kashima2009link}
\bibfield{author}{\bibinfo{person}{Hisashi Kashima}, \bibinfo{person}{Tsuyoshi
  Kato}, \bibinfo{person}{Yoshihiro Yamanishi}, \bibinfo{person}{Masashi
  Sugiyama}, {and} \bibinfo{person}{Koji Tsuda}.}
  \bibinfo{year}{2009}\natexlab{}.
\newblock \showarticletitle{Link propagation: A fast semi-supervised learning
  algorithm for link prediction}.
\newblock \bibinfo{journal}{{\em ICDM\/}} (\bibinfo{year}{2009}).
\newblock


\bibitem[\protect\citeauthoryear{Kaya and U{\c{c}}ar}{Kaya and
  U{\c{c}}ar}{2016}]%
        {kaya2016parallel}
\bibfield{author}{\bibinfo{person}{Oguz Kaya} {and} \bibinfo{person}{Bora
  U{\c{c}}ar}.} \bibinfo{year}{2016}\natexlab{}.
\newblock {\em \bibinfo{title}{Parallel CP decomposition of sparse tensors
  using dimension trees}}.
\newblock \bibinfo{thesistype}{Ph.D. Dissertation}.
  \bibinfo{school}{Inria-Research Centre Grenoble--Rh{\^o}ne-Alpes}.
\newblock


\bibitem[\protect\citeauthoryear{Kiers}{Kiers}{1997}]%
        {kiers1997weighted}
\bibfield{author}{\bibinfo{person}{Henk~AL Kiers}.}
  \bibinfo{year}{1997}\natexlab{}.
\newblock \showarticletitle{Weighted least squares fitting using ordinary least
  squares algorithms}.
\newblock \bibinfo{journal}{{\em Psychometrika\/}} (\bibinfo{year}{1997}).
\newblock


\bibitem[\protect\citeauthoryear{Kiers}{Kiers}{2000}]%
        {kiers2000towards}
\bibfield{author}{\bibinfo{person}{Henk~AL Kiers}.}
  \bibinfo{year}{2000}\natexlab{}.
\newblock \showarticletitle{Towards a standardized notation and terminology in
  multiway analysis}.
\newblock \bibinfo{journal}{{\em Journal of chemometrics\/}}
  (\bibinfo{year}{2000}).
\newblock


\bibitem[\protect\citeauthoryear{Kiers and Mechelen}{Kiers and
  Mechelen}{2001}]%
        {kiers2001three}
\bibfield{author}{\bibinfo{person}{Henk~AL Kiers} {and}
  \bibinfo{person}{Iven~Van Mechelen}.} \bibinfo{year}{2001}\natexlab{}.
\newblock \showarticletitle{Three-way component analysis: Principles and
  illustrative application.}
\newblock \bibinfo{journal}{{\em Psychological methods\/}}
  (\bibinfo{year}{2001}).
\newblock


\bibitem[\protect\citeauthoryear{Kiers, Ten~Berge, and Bro}{Kiers
  et~al\mbox{.}}{1999}]%
        {kiers1999parafac2}
\bibfield{author}{\bibinfo{person}{Henk~AL Kiers}, \bibinfo{person}{Jos~MF
  Ten~Berge}, {and} \bibinfo{person}{Rasmus Bro}.}
  \bibinfo{year}{1999}\natexlab{}.
\newblock \showarticletitle{PARAFAC2-Part I. A direct fitting algorithm for the
  PARAFAC2 model}.
\newblock \bibinfo{journal}{{\em Journal of Chemometrics\/}}
  (\bibinfo{year}{1999}).
\newblock


\bibitem[\protect\citeauthoryear{Kilmer, Braman, Hao, and Hoover}{Kilmer
  et~al\mbox{.}}{2013}]%
        {kilmer2013third}
\bibfield{author}{\bibinfo{person}{Misha~E Kilmer}, \bibinfo{person}{Karen
  Braman}, \bibinfo{person}{Ning Hao}, {and} \bibinfo{person}{Randy~C Hoover}.}
  \bibinfo{year}{2013}\natexlab{}.
\newblock \showarticletitle{Third-order tensors as operators on matrices: A
  theoretical and computational framework with applications in imaging}.
\newblock \bibinfo{journal}{{\it SIAM J. Matrix Anal. Appl.}}
  (\bibinfo{year}{2013}).
\newblock


\bibitem[\protect\citeauthoryear{Kline}{Kline}{1990}]%
        {kline1990mathematical}
\bibfield{author}{\bibinfo{person}{Morris Kline}.}
  \bibinfo{year}{1990}\natexlab{}.
\newblock \bibinfo{booktitle}{{\em Mathematical Thought From Ancient to Modern
  Times: Volume 3}}.
\newblock \bibinfo{publisher}{OUP USA}.
\newblock


\bibitem[\protect\citeauthoryear{Kolda and Bader}{Kolda and Bader}{2009}]%
        {Kolda:Tensor}
\bibfield{author}{\bibinfo{person}{Tamara~G Kolda} {and}
  \bibinfo{person}{Brett~W Bader}.} \bibinfo{year}{2009}\natexlab{}.
\newblock \showarticletitle{Tensor decompositions and applications}.
\newblock \bibinfo{journal}{{\em SIAM review\/}} (\bibinfo{year}{2009}).
\newblock


\bibitem[\protect\citeauthoryear{Kolda and Sun}{Kolda and Sun}{2008}]%
        {kolda2008scalable}
\bibfield{author}{\bibinfo{person}{Tamara~G Kolda} {and}
  \bibinfo{person}{Jimeng Sun}.} \bibinfo{year}{2008}\natexlab{}.
\newblock \showarticletitle{Scalable tensor decompositions for multi-aspect
  data mining}. In \bibinfo{booktitle}{{\em IEEE ICDM}}.
\newblock


\bibitem[\protect\citeauthoryear{Koren}{Koren}{2008}]%
        {koren2008factorization}
\bibfield{author}{\bibinfo{person}{Yehuda Koren}.}
  \bibinfo{year}{2008}\natexlab{}.
\newblock \showarticletitle{Factorization meets the neighborhood: a
  multifaceted collaborative filtering model}.
\newblock \bibinfo{journal}{{\em ACM SIGKDD\/}} (\bibinfo{year}{2008}).
\newblock


\bibitem[\protect\citeauthoryear{Koren, Bell, and Volinsky}{Koren
  et~al\mbox{.}}{2009}]%
        {koren2009matrix}
\bibfield{author}{\bibinfo{person}{Yehuda Koren}, \bibinfo{person}{Robert
  Bell}, {and} \bibinfo{person}{Chris Volinsky}.}
  \bibinfo{year}{2009}\natexlab{}.
\newblock \showarticletitle{Matrix factorization techniques for recommender
  systems}.
\newblock \bibinfo{journal}{{\em Computer\/}} (\bibinfo{year}{2009}).
\newblock


\bibitem[\protect\citeauthoryear{Kressner, Steinlechner, and
  Vandereycken}{Kressner et~al\mbox{.}}{2014}]%
        {kressner2014low}
\bibfield{author}{\bibinfo{person}{Daniel Kressner}, \bibinfo{person}{Michael
  Steinlechner}, {and} \bibinfo{person}{Bart Vandereycken}.}
  \bibinfo{year}{2014}\natexlab{}.
\newblock \showarticletitle{Low-rank tensor completion by Riemannian
  optimization}.
\newblock \bibinfo{journal}{{\em BIT Numerical Mathematics\/}}
  (\bibinfo{year}{2014}).
\newblock


\bibitem[\protect\citeauthoryear{Krishnamurthy and Singh}{Krishnamurthy and
  Singh}{2013}]%
        {krishnamurthy2013low}
\bibfield{author}{\bibinfo{person}{Akshay Krishnamurthy} {and}
  \bibinfo{person}{Aarti Singh}.} \bibinfo{year}{2013}\natexlab{}.
\newblock \showarticletitle{Low-rank matrix and tensor completion via adaptive
  sampling}. In \bibinfo{booktitle}{{\em NIPS}}.
\newblock


\bibitem[\protect\citeauthoryear{Kroonenberg}{Kroonenberg}{1983}]%
        {kroonenberg1983three}
\bibfield{author}{\bibinfo{person}{Pieter~M Kroonenberg}.}
  \bibinfo{year}{1983}\natexlab{}.
\newblock \showarticletitle{Three-mode principal component analysis: Theory and
  applications}.
\newblock \bibinfo{journal}{{\em DSWO press\/}} (\bibinfo{year}{1983}).
\newblock


\bibitem[\protect\citeauthoryear{Kroonenberg}{Kroonenberg}{2008}]%
        {kroonenberg2008applied}
\bibfield{author}{\bibinfo{person}{Pieter~M Kroonenberg}.}
  \bibinfo{year}{2008}\natexlab{}.
\newblock \bibinfo{booktitle}{{\em Applied multiway data analysis}}.
  Vol.~\bibinfo{volume}{702}.
\newblock \bibinfo{publisher}{John Wiley \&amp; Sons}.
\newblock


\bibitem[\protect\citeauthoryear{Kroonenberg and De~Leeuw}{Kroonenberg and
  De~Leeuw}{1980}]%
        {kroonenberg1980principal}
\bibfield{author}{\bibinfo{person}{Pieter~M Kroonenberg} {and}
  \bibinfo{person}{Jan De~Leeuw}.} \bibinfo{year}{1980}\natexlab{}.
\newblock \showarticletitle{Principal component analysis of three-mode data by
  means of alternating least squares algorithms}.
\newblock \bibinfo{journal}{{\em Psychometrika\/}} (\bibinfo{year}{1980}).
\newblock


\bibitem[\protect\citeauthoryear{Kruskal}{Kruskal}{1977}]%
        {kruskal1977three}
\bibfield{author}{\bibinfo{person}{Joseph~B Kruskal}.}
  \bibinfo{year}{1977}\natexlab{}.
\newblock \showarticletitle{Three-way arrays: rank and uniqueness of trilinear
  decompositions, with application to arithmetic complexity and statistics}.
\newblock \bibinfo{journal}{{\em Linear algebra and its applications\/}}
  (\bibinfo{year}{1977}).
\newblock


\bibitem[\protect\citeauthoryear{Kruskal}{Kruskal}{1989}]%
        {Kruskal1989rank}
\bibfield{author}{\bibinfo{person}{Joseph~B. Kruskal}.}
  \bibinfo{year}{1989}\natexlab{}.
\newblock \showarticletitle{RANK, DECOMPOSITION, AND UNIQUENESS FOR 3-WAY AND
  iV-WAY ARRAYS, R. Coppi and S. Bolasco (eds.)}.
\newblock \bibinfo{journal}{{\em Multiway Data Analysis\/}}
  (\bibinfo{year}{1989}).
\newblock


\bibitem[\protect\citeauthoryear{Lamba, Nagarajan, Shin, and
  Shajarisales}{Lamba et~al\mbox{.}}{2016}]%
        {lamba2016incorporating}
\bibfield{author}{\bibinfo{person}{Hemank Lamba}, \bibinfo{person}{Vaishnavh
  Nagarajan}, \bibinfo{person}{Kijung Shin}, {and} \bibinfo{person}{Naji
  Shajarisales}.} \bibinfo{year}{2016}\natexlab{}.
\newblock \showarticletitle{Incorporating side information in tensor
  completion}.
\newblock \bibinfo{journal}{{\em WWW\/}} (\bibinfo{year}{2016}).
\newblock


\bibitem[\protect\citeauthoryear{Laurent}{Laurent}{2009}]%
        {laurent2009matrix}
\bibfield{author}{\bibinfo{person}{Monique Laurent}.}
  \bibinfo{year}{2009}\natexlab{}.
\newblock \showarticletitle{Matrix Completion Problems.}
\newblock \bibinfo{journal}{{\em Encyclopedia of Optimization\/}}
  (\bibinfo{year}{2009}).
\newblock


\bibitem[\protect\citeauthoryear{Lee}{Lee}{2012}]%
        {lee2012introduction}
\bibfield{author}{\bibinfo{person}{John Lee}.} \bibinfo{year}{2012}\natexlab{}.
\newblock \bibinfo{booktitle}{{\em Introduction to smooth manifolds}}.
\newblock \bibinfo{publisher}{Springer Science \&amp; Business Media}.
\newblock


\bibitem[\protect\citeauthoryear{Li, Choi, Perros, Sun, and Vuduc}{Li
  et~al\mbox{.}}{2017}]%
        {li2017model}
\bibfield{author}{\bibinfo{person}{Jiajia Li}, \bibinfo{person}{Jee Choi},
  \bibinfo{person}{Ioakeim Perros}, \bibinfo{person}{Jimeng Sun}, {and}
  \bibinfo{person}{Richard Vuduc}.} \bibinfo{year}{2017}\natexlab{}.
\newblock \showarticletitle{Model-driven sparse CP decomposition for
  higher-order tensors}. In \bibinfo{booktitle}{{\em IPDPS, IEEE
  International}}.
\newblock


\bibitem[\protect\citeauthoryear{Li, Schonfeld, and Friedland}{Li
  et~al\mbox{.}}{2013}]%
        {li2013generalized}
\bibfield{author}{\bibinfo{person}{Qun Li}, \bibinfo{person}{Dan Schonfeld},
  {and} \bibinfo{person}{Shmuel Friedland}.} \bibinfo{year}{2013}\natexlab{}.
\newblock \showarticletitle{Generalized tensor compressive sensing}.
\newblock \bibinfo{journal}{{\em IEEE ICME\/}} (\bibinfo{year}{2013}).
\newblock


\bibitem[\protect\citeauthoryear{Li and Yeung}{Li and Yeung}{2009}]%
        {li2009relation}
\bibfield{author}{\bibinfo{person}{Wu-Jun Li} {and} \bibinfo{person}{Dit~Yan
  Yeung}.} \bibinfo{year}{2009}\natexlab{}.
\newblock \showarticletitle{Relation regularized matrix factorization}.
\newblock \bibinfo{journal}{{\em IJCAI\/}} (\bibinfo{year}{2009}).
\newblock


\bibitem[\protect\citeauthoryear{Li, Dai, and Xiong}{Li et~al\mbox{.}}{2016}]%
        {li2016compressive}
\bibfield{author}{\bibinfo{person}{Yong Li}, \bibinfo{person}{Wenrui Dai},
  {and} \bibinfo{person}{Hongkai Xiong}.} \bibinfo{year}{2016}\natexlab{}.
\newblock \showarticletitle{Compressive Tensor Sampling with Structured
  Sparsity}.
\newblock \bibinfo{journal}{{\em IEEE DCC\/}} (\bibinfo{year}{2016}).
\newblock


\bibitem[\protect\citeauthoryear{Li, Yan, Zhou, and Yang}{Li
  et~al\mbox{.}}{2010}]%
        {li2010optimum}
\bibfield{author}{\bibinfo{person}{Yin Li}, \bibinfo{person}{Junchi Yan},
  \bibinfo{person}{Yue Zhou}, {and} \bibinfo{person}{Jie Yang}.}
  \bibinfo{year}{2010}\natexlab{}.
\newblock \showarticletitle{Optimum subspace learning and error correction for
  tensors}.
\newblock \bibinfo{journal}{{\em Computer Vision--ECCV 2010\/}}
  (\bibinfo{year}{2010}).
\newblock


\bibitem[\protect\citeauthoryear{Lin, Sun, Castro, Konuru, Sundaram, and
  Kelliher}{Lin et~al\mbox{.}}{2009}]%
        {lin2009metafac}
\bibfield{author}{\bibinfo{person}{Yu-Ru Lin}, \bibinfo{person}{Jimeng Sun},
  \bibinfo{person}{Paul Castro}, \bibinfo{person}{Ravi Konuru},
  \bibinfo{person}{Hari Sundaram}, {and} \bibinfo{person}{Aisling Kelliher}.}
  \bibinfo{year}{2009}\natexlab{}.
\newblock \showarticletitle{Metafac: community discovery via relational
  hypergraph factorization}.
\newblock \bibinfo{journal}{{\em KDD\/}} (\bibinfo{year}{2009}).
\newblock


\bibitem[\protect\citeauthoryear{Lin, Chen, and Ma}{Lin et~al\mbox{.}}{2010}]%
        {lin2010augmented}
\bibfield{author}{\bibinfo{person}{Zhouchen Lin}, \bibinfo{person}{Minming
  Chen}, {and} \bibinfo{person}{Yi Ma}.} \bibinfo{year}{2010}\natexlab{}.
\newblock \showarticletitle{The augmented lagrange multiplier method for exact
  recovery of corrupted low-rank matrices}.
\newblock \bibinfo{journal}{{\em arXiv preprint arXiv:1009.5055\/}}
  (\bibinfo{year}{2010}).
\newblock


\bibitem[\protect\citeauthoryear{Liu, Musialski, Wonka, and Ye}{Liu
  et~al\mbox{.}}{2009}]%
        {liu2009tensor}
\bibfield{author}{\bibinfo{person}{Ji Liu}, \bibinfo{person}{Przemyslaw
  Musialski}, \bibinfo{person}{Peter Wonka}, {and} \bibinfo{person}{Jieping
  Ye}.} \bibinfo{year}{2009}\natexlab{}.
\newblock \showarticletitle{Tensor completion for estimating missing values in
  visual data}.
\newblock \bibinfo{journal}{{\em ICCV\/}} (\bibinfo{year}{2009}).
\newblock


\bibitem[\protect\citeauthoryear{Liu, Musialski, Wonka, and Ye}{Liu
  et~al\mbox{.}}{2013}]%
        {liu2013tensor}
\bibfield{author}{\bibinfo{person}{Ji Liu}, \bibinfo{person}{Przemyslaw
  Musialski}, \bibinfo{person}{Peter Wonka}, {and} \bibinfo{person}{Jieping
  Ye}.} \bibinfo{year}{2013}\natexlab{}.
\newblock \showarticletitle{Tensor completion for estimating missing values in
  visual data}.
\newblock \bibinfo{journal}{{\em TPAMI\/}} (\bibinfo{year}{2013}).
\newblock


\bibitem[\protect\citeauthoryear{Liu and Shang}{Liu and Shang}{2013}]%
        {liu2013efficient}
\bibfield{author}{\bibinfo{person}{Yuanyuan Liu} {and} \bibinfo{person}{Fanhua
  Shang}.} \bibinfo{year}{2013}\natexlab{}.
\newblock \showarticletitle{An efficient matrix factorization method for tensor
  completion}.
\newblock \bibinfo{journal}{{\em IEEE Signal Processing Letters\/}}
  (\bibinfo{year}{2013}).
\newblock


\bibitem[\protect\citeauthoryear{Liu, Shang, Cheng, Cheng, and Tong}{Liu
  et~al\mbox{.}}{2014a}]%
        {liu2014factor}
\bibfield{author}{\bibinfo{person}{Yuanyuan Liu}, \bibinfo{person}{Fanhua
  Shang}, \bibinfo{person}{Hong Cheng}, \bibinfo{person}{James Cheng}, {and}
  \bibinfo{person}{Hanghang Tong}.} \bibinfo{year}{2014}\natexlab{a}.
\newblock \showarticletitle{Factor matrix trace norm minimization for low-rank
  tensor completion}.
\newblock \bibinfo{journal}{{\em ICDM\/}} (\bibinfo{year}{2014}).
\newblock


\bibitem[\protect\citeauthoryear{Liu, Shang, Fan, Cheng, and Cheng}{Liu
  et~al\mbox{.}}{2014b}]%
        {liu2014generalized}
\bibfield{author}{\bibinfo{person}{Yuanyuan Liu}, \bibinfo{person}{Fanhua
  Shang}, \bibinfo{person}{Wei Fan}, \bibinfo{person}{James Cheng}, {and}
  \bibinfo{person}{Hong Cheng}.} \bibinfo{year}{2014}\natexlab{b}.
\newblock \showarticletitle{Generalized higher-order orthogonal iteration for
  tensor decomposition and completion}. In \bibinfo{booktitle}{{\em Advances in
  Neural Information Processing Systems}}.
\newblock


\bibitem[\protect\citeauthoryear{Liu, Shang, Jiao, Cheng, and Cheng}{Liu
  et~al\mbox{.}}{2015}]%
        {liu2015trace}
\bibfield{author}{\bibinfo{person}{Yuanyuan Liu}, \bibinfo{person}{Fanhua
  Shang}, \bibinfo{person}{Licheng Jiao}, \bibinfo{person}{James Cheng}, {and}
  \bibinfo{person}{Hong Cheng}.} \bibinfo{year}{2015}\natexlab{}.
\newblock \showarticletitle{Trace norm regularized CANDECOMP/PARAFAC
  decomposition with missing data}.
\newblock \bibinfo{journal}{{\em IEEE transactions on cybernetics\/}}
  (\bibinfo{year}{2015}).
\newblock


\bibitem[\protect\citeauthoryear{Lu, Plataniotis, and Venetsanopoulos}{Lu
  et~al\mbox{.}}{2011}]%
        {lu2011survey}
\bibfield{author}{\bibinfo{person}{Haiping Lu}, \bibinfo{person}{Konstantinos~N
  Plataniotis}, {and} \bibinfo{person}{Anastasios~N Venetsanopoulos}.}
  \bibinfo{year}{2011}\natexlab{}.
\newblock \showarticletitle{A survey of multilinear subspace learning for
  tensor data}.
\newblock \bibinfo{journal}{{\em Pattern Recognition\/}}
  (\bibinfo{year}{2011}).
\newblock


\bibitem[\protect\citeauthoryear{Ma, Schonfeld, and Khokhar}{Ma
  et~al\mbox{.}}{2009}]%
        {ma2009dynamic}
\bibfield{author}{\bibinfo{person}{Xiang Ma}, \bibinfo{person}{Dan Schonfeld},
  {and} \bibinfo{person}{Ashfaq Khokhar}.} \bibinfo{year}{2009}\natexlab{}.
\newblock \showarticletitle{Dynamic updating and downdating matrix SVD and
  tensor HOSVD for adaptive indexing and retrieval of motion trajectories}.
\newblock \bibinfo{journal}{{\em ICASSP\/}} (\bibinfo{year}{2009}).
\newblock


\bibitem[\protect\citeauthoryear{Mardani, Mateos, and Giannakis}{Mardani
  et~al\mbox{.}}{2014}]%
        {mardani2014imputation}
\bibfield{author}{\bibinfo{person}{Morteza Mardani}, \bibinfo{person}{Gonzalo
  Mateos}, {and} \bibinfo{person}{Georgios~B Giannakis}.}
  \bibinfo{year}{2014}\natexlab{}.
\newblock \showarticletitle{Imputation of streaming low-rank tensor data}. In
  \bibinfo{booktitle}{{\em IEEE SAM}}.
\newblock


\bibitem[\protect\citeauthoryear{Mardani, Mateos, and Giannakis}{Mardani
  et~al\mbox{.}}{2015}]%
        {Mardani:Tensor}
\bibfield{author}{\bibinfo{person}{Morteza Mardani}, \bibinfo{person}{Gonzalo
  Mateos}, {and} \bibinfo{person}{Georgios~B Giannakis}.}
  \bibinfo{year}{2015}\natexlab{}.
\newblock \showarticletitle{Subspace learning and imputation for streaming big
  data matrices and tensors}.
\newblock \bibinfo{journal}{{\em IEEE TSP\/}} (\bibinfo{year}{2015}).
\newblock


\bibitem[\protect\citeauthoryear{Matsubara, Sakurai, Faloutsos, Iwata, and
  Yoshikawa}{Matsubara et~al\mbox{.}}{2012}]%
        {matsubara2012fast}
\bibfield{author}{\bibinfo{person}{Yasuko Matsubara}, \bibinfo{person}{Yasushi
  Sakurai}, \bibinfo{person}{Christos Faloutsos}, \bibinfo{person}{Tomoharu
  Iwata}, {and} \bibinfo{person}{Masatoshi Yoshikawa}.}
  \bibinfo{year}{2012}\natexlab{}.
\newblock \showarticletitle{Fast mining and forecasting of complex time-stamped
  events}.
\newblock \bibinfo{journal}{{\em KDD\/}} (\bibinfo{year}{2012}).
\newblock


\bibitem[\protect\citeauthoryear{Meng, Morris, and Martin}{Meng
  et~al\mbox{.}}{2003}]%
        {meng2003line}
\bibfield{author}{\bibinfo{person}{X Meng}, \bibinfo{person}{AJ Morris}, {and}
  \bibinfo{person}{EB Martin}.} \bibinfo{year}{2003}\natexlab{}.
\newblock \showarticletitle{On-line monitoring of batch processes using a
  PARAFAC representation}.
\newblock \bibinfo{journal}{{\em Journal of Chemometrics\/}}
  (\bibinfo{year}{2003}).
\newblock


\bibitem[\protect\citeauthoryear{Mishra and Sepulchre}{Mishra and
  Sepulchre}{2014}]%
        {mishra2014r3mc}
\bibfield{author}{\bibinfo{person}{Bamdev Mishra} {and}
  \bibinfo{person}{Rodolphe Sepulchre}.} \bibinfo{year}{2014}\natexlab{}.
\newblock \showarticletitle{R3MC: A Riemannian three-factor algorithm for
  low-rank matrix completion}.
\newblock \bibinfo{journal}{{\em CDC\/}} (\bibinfo{year}{2014}).
\newblock


\bibitem[\protect\citeauthoryear{Montanari and Sun}{Montanari and Sun}{2016}]%
        {montanari2016spectral}
\bibfield{author}{\bibinfo{person}{Andrea Montanari} {and}
  \bibinfo{person}{Nike Sun}.} \bibinfo{year}{2016}\natexlab{}.
\newblock \showarticletitle{Spectral algorithms for tensor completion}.
\newblock \bibinfo{journal}{{\em arXiv preprint arXiv:1612.07866\/}}
  (\bibinfo{year}{2016}).
\newblock


\bibitem[\protect\citeauthoryear{Mu, Huang, Wright, and Goldfarb}{Mu
  et~al\mbox{.}}{2014}]%
        {mu2014square}
\bibfield{author}{\bibinfo{person}{Cun Mu}, \bibinfo{person}{Bo Huang},
  \bibinfo{person}{John Wright}, {and} \bibinfo{person}{Donald Goldfarb}.}
  \bibinfo{year}{2014}\natexlab{}.
\newblock \showarticletitle{Square Deal: Lower Bounds and Improved Relaxations
  for Tensor Recovery.}
\newblock \bibinfo{journal}{{\em ICML\/}} (\bibinfo{year}{2014}).
\newblock


\bibitem[\protect\citeauthoryear{Narita, Hayashi, Tomioka, and Kashima}{Narita
  et~al\mbox{.}}{2011}]%
        {Narita:Tensor}
\bibfield{author}{\bibinfo{person}{Atsuhiro Narita}, \bibinfo{person}{Kohei
  Hayashi}, \bibinfo{person}{Ryota Tomioka}, {and} \bibinfo{person}{Hisashi
  Kashima}.} \bibinfo{year}{2011}\natexlab{}.
\newblock \showarticletitle{Tensor factorization using auxiliary information}.
\newblock \bibinfo{journal}{{\em ECML PKDD\/}} (\bibinfo{year}{2011}).
\newblock


\bibitem[\protect\citeauthoryear{Natarajan and Dhillon}{Natarajan and
  Dhillon}{2014}]%
        {natarajan2014inductive}
\bibfield{author}{\bibinfo{person}{Nagarajan Natarajan} {and}
  \bibinfo{person}{Inderjit~S Dhillon}.} \bibinfo{year}{2014}\natexlab{}.
\newblock \showarticletitle{Inductive matrix completion for predicting
  gene--disease associations}.
\newblock \bibinfo{journal}{{\em Bioinformatics\/}} (\bibinfo{year}{2014}).
\newblock


\bibitem[\protect\citeauthoryear{Nimishakavi, Mishra, Gupta, and
  Talukdar}{Nimishakavi et~al\mbox{.}}{2018}]%
        {nimishakavi2018inductive}
\bibfield{author}{\bibinfo{person}{Madhav Nimishakavi}, \bibinfo{person}{Bamdev
  Mishra}, \bibinfo{person}{Manish Gupta}, {and} \bibinfo{person}{Partha
  Talukdar}.} \bibinfo{year}{2018}\natexlab{}.
\newblock \showarticletitle{Inductive Framework for Multi-Aspect Streaming
  Tensor Completion with Side Information}.
\newblock \bibinfo{journal}{{\em arXiv preprint arXiv:1802.06371\/}}
  (\bibinfo{year}{2018}).
\newblock


\bibitem[\protect\citeauthoryear{Nion and Sidiropoulos}{Nion and
  Sidiropoulos}{2009}]%
        {Nion:Tensor}
\bibfield{author}{\bibinfo{person}{Dimitri Nion} {and}
  \bibinfo{person}{Nicholas~D Sidiropoulos}.} \bibinfo{year}{2009}\natexlab{}.
\newblock \showarticletitle{Adaptive algorithms to track the PARAFAC
  decomposition of a third-order tensor}.
\newblock \bibinfo{journal}{{\em IEEE TSP\/}} (\bibinfo{year}{2009}).
\newblock


\bibitem[\protect\citeauthoryear{Nishimori and Akaho}{Nishimori and
  Akaho}{2005}]%
        {nishimori2005learning}
\bibfield{author}{\bibinfo{person}{Yasunori Nishimori} {and}
  \bibinfo{person}{Shotaro Akaho}.} \bibinfo{year}{2005}\natexlab{}.
\newblock \showarticletitle{Learning algorithms utilizing quasi-geodesic flows
  on the Stiefel manifold}.
\newblock \bibinfo{journal}{{\em Neurocomputing\/}} (\bibinfo{year}{2005}).
\newblock


\bibitem[\protect\citeauthoryear{Oh, Park, Sael, and Kang}{Oh
  et~al\mbox{.}}{2017}]%
        {oh2017scalable}
\bibfield{author}{\bibinfo{person}{Sejoon Oh}, \bibinfo{person}{Namyong Park},
  \bibinfo{person}{Lee Sael}, {and} \bibinfo{person}{U Kang}.}
  \bibinfo{year}{2017}\natexlab{}.
\newblock \showarticletitle{Scalable Tucker Factorization for Sparse
  Tensors-Algorithms and Discoveries}.
\newblock \bibinfo{journal}{{\em arXiv preprint arXiv:1710.02261\/}}
  (\bibinfo{year}{2017}).
\newblock


\bibitem[\protect\citeauthoryear{Oseledets}{Oseledets}{2011}]%
        {oseledets2011tensor}
\bibfield{author}{\bibinfo{person}{Ivan~V Oseledets}.}
  \bibinfo{year}{2011}\natexlab{}.
\newblock \showarticletitle{Tensor-train decomposition}.
\newblock \bibinfo{journal}{{\em SIAM Journal on Scientific Computing\/}}
  (\bibinfo{year}{2011}).
\newblock


\bibitem[\protect\citeauthoryear{Paatero}{Paatero}{1997}]%
        {paatero1997weighted}
\bibfield{author}{\bibinfo{person}{Pentti Paatero}.}
  \bibinfo{year}{1997}\natexlab{}.
\newblock \showarticletitle{A weighted non-negative least squares algorithm for
  three-way `PARAFAC'factor analysis}.
\newblock \bibinfo{journal}{{\em Chemometrics and Intelligent Laboratory
  Systems\/}} (\bibinfo{year}{1997}).
\newblock


\bibitem[\protect\citeauthoryear{Pantraki and Kotropoulos}{Pantraki and
  Kotropoulos}{2015}]%
        {pantraki2015automatic}
\bibfield{author}{\bibinfo{person}{Evangelia Pantraki} {and}
  \bibinfo{person}{Constantine Kotropoulos}.} \bibinfo{year}{2015}\natexlab{}.
\newblock \showarticletitle{Automatic image tagging and recommendation via
  PARAFAC2}.
\newblock \bibinfo{journal}{{\em Machine Learning for Signal Processing (MLSP),
  2015 IEEE 25th International Workshop on\/}} (\bibinfo{year}{2015}).
\newblock


\bibitem[\protect\citeauthoryear{Papalexakis, Faloutsos, and
  Sidiropoulos}{Papalexakis et~al\mbox{.}}{2016}]%
        {Papalexakis:Tensor}
\bibfield{author}{\bibinfo{person}{Evangelos~E. Papalexakis},
  \bibinfo{person}{Christos Faloutsos}, {and} \bibinfo{person}{Nicholas~D.
  Sidiropoulos}.} \bibinfo{year}{2016}\natexlab{}.
\newblock \showarticletitle{Tensors for data mining and data fusion: models,
  applications, and scalable Algorithms}.
\newblock \bibinfo{journal}{{\em TIST\/}} (\bibinfo{year}{2016}).
\newblock


\bibitem[\protect\citeauthoryear{Papastergiou and Megalooikonomou}{Papastergiou
  and Megalooikonomou}{2017}]%
        {papastergiou2017distributed}
\bibfield{author}{\bibinfo{person}{T Papastergiou} {and} \bibinfo{person}{V
  Megalooikonomou}.} \bibinfo{year}{2017}\natexlab{}.
\newblock \showarticletitle{A distributed proximal gradient descent method for
  tensor completion}. In \bibinfo{booktitle}{{\em Big Data (Big Data), 2017
  IEEE International Conference on}}.
\newblock


\bibitem[\protect\citeauthoryear{Peng, Zeng, Zhao, and Wang}{Peng
  et~al\mbox{.}}{2010}]%
        {peng2010collaborative}
\bibfield{author}{\bibinfo{person}{Jing Peng}, \bibinfo{person}{Daniel~Dajun
  Zeng}, \bibinfo{person}{Huimin Zhao}, {and} \bibinfo{person}{Fei-yue Wang}.}
  \bibinfo{year}{2010}\natexlab{}.
\newblock \showarticletitle{Collaborative filtering in social tagging systems
  based on joint item-tag recommendations}.
\newblock \bibinfo{journal}{{\em ACM CIKM\/}} (\bibinfo{year}{2010}).
\newblock


\bibitem[\protect\citeauthoryear{Phan and Cichocki}{Phan and Cichocki}{2011}]%
        {phan2011parafac}
\bibfield{author}{\bibinfo{person}{Anh~Huy Phan} {and} \bibinfo{person}{Andrzej
  Cichocki}.} \bibinfo{year}{2011}\natexlab{}.
\newblock \showarticletitle{PARAFAC algorithms for large-scale problems}.
\newblock \bibinfo{journal}{{\em Neurocomputing\/}} (\bibinfo{year}{2011}).
\newblock


\bibitem[\protect\citeauthoryear{Phien, Tuan, Bengua, and Do}{Phien
  et~al\mbox{.}}{2016}]%
        {phien2016efficient}
\bibfield{author}{\bibinfo{person}{Ho~N Phien}, \bibinfo{person}{Hoang~D Tuan},
  \bibinfo{person}{Johann~A Bengua}, {and} \bibinfo{person}{Minh~N Do}.}
  \bibinfo{year}{2016}\natexlab{}.
\newblock \showarticletitle{Efficient tensor completion: Low-rank tensor
  train}.
\newblock \bibinfo{journal}{{\em arXiv preprint arXiv:1601.01083\/}}
  (\bibinfo{year}{2016}).
\newblock


\bibitem[\protect\citeauthoryear{Rai, Wang, Guo, Chen, Dunson, and Carin}{Rai
  et~al\mbox{.}}{2014}]%
        {rai2014scalable}
\bibfield{author}{\bibinfo{person}{Piyush Rai}, \bibinfo{person}{Yingjian
  Wang}, \bibinfo{person}{Shengbo Guo}, \bibinfo{person}{Gary Chen},
  \bibinfo{person}{David Dunson}, {and} \bibinfo{person}{Lawrence Carin}.}
  \bibinfo{year}{2014}\natexlab{}.
\newblock \showarticletitle{Scalable Bayesian low-rank decomposition of
  incomplete multiway tensors}. In \bibinfo{booktitle}{{\em International
  Conference on Machine Learning}}.
\newblock


\bibitem[\protect\citeauthoryear{Rauhut, Schneider, and Stojanac}{Rauhut
  et~al\mbox{.}}{2015}]%
        {rauhut2015tensor}
\bibfield{author}{\bibinfo{person}{Holger Rauhut}, \bibinfo{person}{Reinhold
  Schneider}, {and} \bibinfo{person}{{\v{Z}}eljka Stojanac}.}
  \bibinfo{year}{2015}\natexlab{}.
\newblock \showarticletitle{Tensor completion in hierarchical tensor
  representations}.
\newblock \bibinfo{journal}{{\em Compressed Sensing and its Applications\/}}
  (\bibinfo{year}{2015}).
\newblock


\bibitem[\protect\citeauthoryear{Rauhut, Schneider, and Stojanac}{Rauhut
  et~al\mbox{.}}{2017}]%
        {rauhut2017low}
\bibfield{author}{\bibinfo{person}{Holger Rauhut}, \bibinfo{person}{Reinhold
  Schneider}, {and} \bibinfo{person}{{\v{Z}}eljka Stojanac}.}
  \bibinfo{year}{2017}\natexlab{}.
\newblock \showarticletitle{Low rank tensor recovery via iterative hard
  thresholding}.
\newblock \bibinfo{journal}{{\it Linear Algebra Appl.}} (\bibinfo{year}{2017}).
\newblock


\bibitem[\protect\citeauthoryear{Rauhut and Stojanac}{Rauhut and
  Stojanac}{2015}]%
        {rauhut2015theta}
\bibfield{author}{\bibinfo{person}{Holger Rauhut} {and}
  \bibinfo{person}{{\v{Z}}eljka Stojanac}.} \bibinfo{year}{2015}\natexlab{}.
\newblock \showarticletitle{Tensor theta norms and low rank recovery}.
\newblock \bibinfo{journal}{{\em arXiv preprint arXiv:1505.05175\/}}
  (\bibinfo{year}{2015}).
\newblock


\bibitem[\protect\citeauthoryear{Recht}{Recht}{2011}]%
        {recht2011simpler}
\bibfield{author}{\bibinfo{person}{Benjamin Recht}.}
  \bibinfo{year}{2011}\natexlab{}.
\newblock \showarticletitle{A simpler approach to matrix completion}.
\newblock \bibinfo{journal}{{\em Journal of Machine Learning Research\/}}
  (\bibinfo{year}{2011}).
\newblock


\bibitem[\protect\citeauthoryear{Recht, Fazel, and Parrilo}{Recht
  et~al\mbox{.}}{2010}]%
        {recht2010guaranteed}
\bibfield{author}{\bibinfo{person}{Benjamin Recht}, \bibinfo{person}{Maryam
  Fazel}, {and} \bibinfo{person}{Pablo~A Parrilo}.}
  \bibinfo{year}{2010}\natexlab{}.
\newblock \showarticletitle{Guaranteed minimum-rank solutions of linear matrix
  equations via nuclear norm minimization}.
\newblock \bibinfo{journal}{{\em SIAM review\/}} (\bibinfo{year}{2010}).
\newblock


\bibitem[\protect\citeauthoryear{Rendle}{Rendle}{2010}]%
        {rendle2010factorization}
\bibfield{author}{\bibinfo{person}{Steffen Rendle}.}
  \bibinfo{year}{2010}\natexlab{}.
\newblock \showarticletitle{Factorization machines}.
\newblock \bibinfo{journal}{{\em ICDM\/}} (\bibinfo{year}{2010}).
\newblock


\bibitem[\protect\citeauthoryear{Rendle, Balby~Marinho, Nanopoulos, and
  Schmidt-Thieme}{Rendle et~al\mbox{.}}{2009}]%
        {rendle2009learning}
\bibfield{author}{\bibinfo{person}{Steffen Rendle}, \bibinfo{person}{Leandro
  Balby~Marinho}, \bibinfo{person}{Alexandros Nanopoulos}, {and}
  \bibinfo{person}{Lars Schmidt-Thieme}.} \bibinfo{year}{2009}\natexlab{}.
\newblock \showarticletitle{Learning optimal ranking with tensor factorization
  for tag recommendation}.
\newblock \bibinfo{journal}{{\em KDD\/}} (\bibinfo{year}{2009}).
\newblock


\bibitem[\protect\citeauthoryear{Rendle and Schmidt-Thieme}{Rendle and
  Schmidt-Thieme}{2010}]%
        {rendle2010pairwise}
\bibfield{author}{\bibinfo{person}{Steffen Rendle} {and} \bibinfo{person}{Lars
  Schmidt-Thieme}.} \bibinfo{year}{2010}\natexlab{}.
\newblock \showarticletitle{Pairwise interaction tensor factorization for
  personalized tag recommendation}.
\newblock \bibinfo{journal}{{\em WSDM\/}} (\bibinfo{year}{2010}).
\newblock


\bibitem[\protect\citeauthoryear{Romera-Paredes and Pontil}{Romera-Paredes and
  Pontil}{2013}]%
        {romera2013new}
\bibfield{author}{\bibinfo{person}{Bernardino Romera-Paredes} {and}
  \bibinfo{person}{Massimiliano Pontil}.} \bibinfo{year}{2013}\natexlab{}.
\newblock \showarticletitle{A new convex relaxation for tensor completion}. In
  \bibinfo{booktitle}{{\em NIPS}}.
\newblock


\bibitem[\protect\citeauthoryear{Sael, Jeon, and Kang}{Sael
  et~al\mbox{.}}{2015}]%
        {Sael:Tensor}
\bibfield{author}{\bibinfo{person}{Lee Sael}, \bibinfo{person}{Inah Jeon},
  {and} \bibinfo{person}{U Kang}.} \bibinfo{year}{2015}\natexlab{}.
\newblock \showarticletitle{Scalable tensor mining}.
\newblock \bibinfo{journal}{{\em Big Data Research\/}} (\bibinfo{year}{2015}).
\newblock


\bibitem[\protect\citeauthoryear{Salakhutdinov and Mnih}{Salakhutdinov and
  Mnih}{2008}]%
        {salakhutdinov2008bayesian}
\bibfield{author}{\bibinfo{person}{Ruslan Salakhutdinov} {and}
  \bibinfo{person}{Andriy Mnih}.} \bibinfo{year}{2008}\natexlab{}.
\newblock \showarticletitle{Bayesian probabilistic matrix factorization using
  Markov chain Monte Carlo}.
\newblock \bibinfo{journal}{{\em ICML\/}} (\bibinfo{year}{2008}).
\newblock


\bibitem[\protect\citeauthoryear{Shin, Sael, and Kang}{Shin
  et~al\mbox{.}}{2017}]%
        {shin2017fully}
\bibfield{author}{\bibinfo{person}{Kijung Shin}, \bibinfo{person}{Lee Sael},
  {and} \bibinfo{person}{U Kang}.} \bibinfo{year}{2017}\natexlab{}.
\newblock \showarticletitle{Fully Scalable Methods for Distributed Tensor
  Factorization}.
\newblock \bibinfo{journal}{{\em IEEE Transactions on Knowledge and Data
  Engineering\/}} (\bibinfo{year}{2017}).
\newblock


\bibitem[\protect\citeauthoryear{Si, Chiang, Hsieh, Rao, and Dhillon}{Si
  et~al\mbox{.}}{2016}]%
        {si2016goal}
\bibfield{author}{\bibinfo{person}{Si Si}, \bibinfo{person}{Kai-Yang Chiang},
  \bibinfo{person}{Cho-Jui Hsieh}, \bibinfo{person}{Nikhil Rao}, {and}
  \bibinfo{person}{Inderjit~S Dhillon}.} \bibinfo{year}{2016}\natexlab{}.
\newblock \showarticletitle{Goal-directed inductive matrix completion}. In
  \bibinfo{booktitle}{{\em SIGKDD}}.
\newblock


\bibitem[\protect\citeauthoryear{Sidiropoulos, De~Lathauwer, Fu, Huang,
  Papalexakis, and Faloutsos}{Sidiropoulos et~al\mbox{.}}{2016}]%
        {sidiropoulos2016tensor}
\bibfield{author}{\bibinfo{person}{Nicholas~D Sidiropoulos},
  \bibinfo{person}{Lieven De~Lathauwer}, \bibinfo{person}{Xiao Fu},
  \bibinfo{person}{Kejun Huang}, \bibinfo{person}{Evangelos~E Papalexakis},
  {and} \bibinfo{person}{Christos Faloutsos}.} \bibinfo{year}{2016}\natexlab{}.
\newblock \showarticletitle{Tensor decomposition for signal processing and
  machine learning}.
\newblock \bibinfo{journal}{{\em arXiv preprint arXiv:1607.01668\/}}
  (\bibinfo{year}{2016}).
\newblock


\bibitem[\protect\citeauthoryear{Sidiropoulos and Kyrillidis}{Sidiropoulos and
  Kyrillidis}{2012}]%
        {sidiropoulos2012multi}
\bibfield{author}{\bibinfo{person}{Nicholas~D Sidiropoulos} {and}
  \bibinfo{person}{Anastasios Kyrillidis}.} \bibinfo{year}{2012}\natexlab{}.
\newblock \showarticletitle{Multi-way compressed sensing for sparse low-rank
  tensors}.
\newblock \bibinfo{journal}{{\em IEEE Signal Processing Letters\/}}
  (\bibinfo{year}{2012}).
\newblock


\bibitem[\protect\citeauthoryear{Signoretto, De~Lathauwer, and
  Suykens}{Signoretto et~al\mbox{.}}{2010}]%
        {signoretto2010nuclear}
\bibfield{author}{\bibinfo{person}{Marco Signoretto}, \bibinfo{person}{Lieven
  De~Lathauwer}, {and} \bibinfo{person}{Johan~AK Suykens}.}
  \bibinfo{year}{2010}\natexlab{}.
\newblock \showarticletitle{Nuclear norms for tensors and their use for convex
  multilinear estimation}.
\newblock \bibinfo{journal}{{\em Submitted to Linear Algebra and Its
  Applications\/}} (\bibinfo{year}{2010}).
\newblock


\bibitem[\protect\citeauthoryear{Signoretto, Dinh, De~Lathauwer, and
  Suykens}{Signoretto et~al\mbox{.}}{2014}]%
        {signoretto2014learning}
\bibfield{author}{\bibinfo{person}{Marco Signoretto},
  \bibinfo{person}{Quoc~Tran Dinh}, \bibinfo{person}{Lieven De~Lathauwer},
  {and} \bibinfo{person}{Johan~AK Suykens}.} \bibinfo{year}{2014}\natexlab{}.
\newblock \showarticletitle{Learning with tensors: a framework based on convex
  optimization and spectral regularization}.
\newblock \bibinfo{journal}{{\em Machine Learning\/}} (\bibinfo{year}{2014}).
\newblock


\bibitem[\protect\citeauthoryear{Signoretto, Van~de Plas, De~Moor, and
  Suykens}{Signoretto et~al\mbox{.}}{2011}]%
        {signoretto2011tensor}
\bibfield{author}{\bibinfo{person}{Marco Signoretto}, \bibinfo{person}{Raf
  Van~de Plas}, \bibinfo{person}{Bart De~Moor}, {and} \bibinfo{person}{Johan~AK
  Suykens}.} \bibinfo{year}{2011}\natexlab{}.
\newblock \showarticletitle{Tensor versus matrix completion: a comparison with
  application to spectral data}.
\newblock \bibinfo{journal}{{\em IEEE Signal Processing Letters\/}}
  (\bibinfo{year}{2011}).
\newblock


\bibitem[\protect\citeauthoryear{Singh and Gordon}{Singh and Gordon}{2008}]%
        {singh2008relational}
\bibfield{author}{\bibinfo{person}{Ajit~P Singh} {and}
  \bibinfo{person}{Geoffrey~J Gordon}.} \bibinfo{year}{2008}\natexlab{}.
\newblock \showarticletitle{Relational learning via collective matrix
  factorization}.
\newblock \bibinfo{journal}{{\em KDD\/}} (\bibinfo{year}{2008}).
\newblock


\bibitem[\protect\citeauthoryear{Smilde, Bro, and Geladi}{Smilde
  et~al\mbox{.}}{2005}]%
        {smilde2005multi}
\bibfield{author}{\bibinfo{person}{Age Smilde}, \bibinfo{person}{Rasmus Bro},
  {and} \bibinfo{person}{Paul Geladi}.} \bibinfo{year}{2005}\natexlab{}.
\newblock \bibinfo{booktitle}{{\em Multi-way analysis: applications in the
  chemical sciences}}.
\newblock \bibinfo{publisher}{John Wiley \& Sons}.
\newblock


\bibitem[\protect\citeauthoryear{Smilde, Westerhuis, and Boque}{Smilde
  et~al\mbox{.}}{2000}]%
        {smilde2000multiway}
\bibfield{author}{\bibinfo{person}{Age~K Smilde}, \bibinfo{person}{Johan~A
  Westerhuis}, {and} \bibinfo{person}{Ricard Boque}.}
  \bibinfo{year}{2000}\natexlab{}.
\newblock \showarticletitle{Multiway multiblock component and covariates
  regression models}.
\newblock \bibinfo{journal}{{\em Journal of Chemometrics\/}}
  (\bibinfo{year}{2000}).
\newblock


\bibitem[\protect\citeauthoryear{Smith, Beri, and Karypis}{Smith
  et~al\mbox{.}}{2017}]%
        {smith2017constrained}
\bibfield{author}{\bibinfo{person}{Shaden Smith}, \bibinfo{person}{Alec Beri},
  {and} \bibinfo{person}{George Karypis}.} \bibinfo{year}{2017}\natexlab{}.
\newblock \showarticletitle{Constrained tensor factorization with accelerated
  AO-ADMM}. In \bibinfo{booktitle}{{\em Parallel Processing (ICPP), 2017 46th
  International Conference on}}.
\newblock


\bibitem[\protect\citeauthoryear{Smith and Karypis}{Smith and Karypis}{2017}]%
        {smith2017accelerating}
\bibfield{author}{\bibinfo{person}{Shaden Smith} {and} \bibinfo{person}{George
  Karypis}.} \bibinfo{year}{2017}\natexlab{}.
\newblock \showarticletitle{Accelerating the tucker decomposition with
  compressed sparse tensors}. In \bibinfo{booktitle}{{\em European Conference
  on Parallel Processing}}.
\newblock


\bibitem[\protect\citeauthoryear{Sobral, Baker, Bouwmans, and Zahzah}{Sobral
  et~al\mbox{.}}{2014}]%
        {sobral2014incremental}
\bibfield{author}{\bibinfo{person}{Andrews Sobral},
  \bibinfo{person}{Christopher~G Baker}, \bibinfo{person}{Thierry Bouwmans},
  {and} \bibinfo{person}{El-hadi Zahzah}.} \bibinfo{year}{2014}\natexlab{}.
\newblock \showarticletitle{Incremental and multi-feature tensor subspace
  learning applied for background modeling and subtraction}.
\newblock \bibinfo{journal}{{\em ICIAR\/}} (\bibinfo{year}{2014}).
\newblock


\bibitem[\protect\citeauthoryear{Solo and Kong}{Solo and Kong}{1994}]%
        {solo1994adaptive}
\bibfield{author}{\bibinfo{person}{Victor Solo} {and} \bibinfo{person}{Xuan
  Kong}.} \bibinfo{year}{1994}\natexlab{}.
\newblock \bibinfo{booktitle}{{\em Adaptive signal processing algorithms:
  stability and performance}}.
\newblock \bibinfo{publisher}{Prentice-Hall, Inc.}
\newblock


\bibitem[\protect\citeauthoryear{Song, Huang, Ge, Caverlee, and Hu}{Song
  et~al\mbox{.}}{2017}]%
        {song2017multi}
\bibfield{author}{\bibinfo{person}{Qingquan Song}, \bibinfo{person}{Xiao
  Huang}, \bibinfo{person}{Hancheng Ge}, \bibinfo{person}{James Caverlee},
  {and} \bibinfo{person}{Xia Hu}.} \bibinfo{year}{2017}\natexlab{}.
\newblock \showarticletitle{Multi-aspect streaming tensor completion}. In
  \bibinfo{booktitle}{{\em KDD}}.
\newblock


\bibitem[\protect\citeauthoryear{Srebro and Shraibman}{Srebro and
  Shraibman}{2005}]%
        {srebro2005rank}
\bibfield{author}{\bibinfo{person}{Nathan Srebro} {and} \bibinfo{person}{Adi
  Shraibman}.} \bibinfo{year}{2005}\natexlab{}.
\newblock \showarticletitle{Rank, trace-norm and max-norm}.
\newblock \bibinfo{journal}{{\em International Conference on Computational
  Learning Theory\/}} (\bibinfo{year}{2005}).
\newblock


\bibitem[\protect\citeauthoryear{Su and Khoshgoftaar}{Su and
  Khoshgoftaar}{2009}]%
        {su2009survey}
\bibfield{author}{\bibinfo{person}{Xiaoyuan Su} {and} \bibinfo{person}{Taghi~M
  Khoshgoftaar}.} \bibinfo{year}{2009}\natexlab{}.
\newblock \showarticletitle{A survey of collaborative filtering techniques}.
\newblock \bibinfo{journal}{{\em Advances in artificial intelligence\/}}
  (\bibinfo{year}{2009}).
\newblock


\bibitem[\protect\citeauthoryear{Sun, Tao, and Faloutsos}{Sun
  et~al\mbox{.}}{2006}]%
        {sun2006beyond}
\bibfield{author}{\bibinfo{person}{Jimeng Sun}, \bibinfo{person}{Dacheng Tao},
  {and} \bibinfo{person}{Christos Faloutsos}.} \bibinfo{year}{2006}\natexlab{}.
\newblock \showarticletitle{Beyond streams and graphs: dynamic tensor
  analysis}.
\newblock \bibinfo{journal}{{\em KDD\/}} (\bibinfo{year}{2006}).
\newblock


\bibitem[\protect\citeauthoryear{Sun, Tao, Papadimitriou, Yu, and
  Faloutsos}{Sun et~al\mbox{.}}{2008}]%
        {sun2008incremental}
\bibfield{author}{\bibinfo{person}{Jimeng Sun}, \bibinfo{person}{Dacheng Tao},
  \bibinfo{person}{Spiros Papadimitriou}, \bibinfo{person}{Philip~S Yu}, {and}
  \bibinfo{person}{Christos Faloutsos}.} \bibinfo{year}{2008}\natexlab{}.
\newblock \showarticletitle{Incremental tensor analysis: theory and
  applications}.
\newblock \bibinfo{journal}{{\em TKDD\/}} (\bibinfo{year}{2008}).
\newblock


\bibitem[\protect\citeauthoryear{Symeonidis}{Symeonidis}{2016}]%
        {symeonidis2016matrix}
\bibfield{author}{\bibinfo{person}{Panagiotis Symeonidis}.}
  \bibinfo{year}{2016}\natexlab{}.
\newblock \showarticletitle{Matrix and tensor decomposition in recommender
  systems}. In \bibinfo{booktitle}{{\em Proceedings of the 10th ACM Conference
  on Recommender Systems}}.
\newblock


\bibitem[\protect\citeauthoryear{Symeonidis, Nanopoulos, and
  Manolopoulos}{Symeonidis et~al\mbox{.}}{2008}]%
        {symeonidis2008tag}
\bibfield{author}{\bibinfo{person}{Panagiotis Symeonidis},
  \bibinfo{person}{Alexandros Nanopoulos}, {and} \bibinfo{person}{Yannis
  Manolopoulos}.} \bibinfo{year}{2008}\natexlab{}.
\newblock \showarticletitle{Tag recommendations based on tensor dimensionality
  reduction}.
\newblock \bibinfo{journal}{{\em RecSys\/}} (\bibinfo{year}{2008}).
\newblock


\bibitem[\protect\citeauthoryear{Takeuchi and Ueda}{Takeuchi and Ueda}{2016}]%
        {takeuchi2016graph}
\bibfield{author}{\bibinfo{person}{Koh Takeuchi} {and} \bibinfo{person}{Naonori
  Ueda}.} \bibinfo{year}{2016}\natexlab{}.
\newblock \showarticletitle{Graph regularized Non-negative Tensor Completion
  for spatio-temporal data analysis}.
\newblock \bibinfo{journal}{{\em 2nd International Workshop on Smart\/}}
  (\bibinfo{year}{2016}).
\newblock


\bibitem[\protect\citeauthoryear{Tan, Cheng, Feng, Feng, Wang, and Zhang}{Tan
  et~al\mbox{.}}{2013a}]%
        {tan2013low}
\bibfield{author}{\bibinfo{person}{Huachun Tan}, \bibinfo{person}{Bin Cheng},
  \bibinfo{person}{Jianshuai Feng}, \bibinfo{person}{Guangdong Feng},
  \bibinfo{person}{Wuhong Wang}, {and} \bibinfo{person}{Yu-Jin Zhang}.}
  \bibinfo{year}{2013}\natexlab{a}.
\newblock \showarticletitle{Low-n-rank tensor recovery based on multi-linear
  augmented lagrange multiplier method}.
\newblock \bibinfo{journal}{{\em Neurocomputing\/}} (\bibinfo{year}{2013}).
\newblock


\bibitem[\protect\citeauthoryear{Tan, Cheng, Wang, Zhang, and Ran}{Tan
  et~al\mbox{.}}{2014}]%
        {tan2014tensor}
\bibfield{author}{\bibinfo{person}{Huachun Tan}, \bibinfo{person}{Bin Cheng},
  \bibinfo{person}{Wuhong Wang}, \bibinfo{person}{Yu-Jin Zhang}, {and}
  \bibinfo{person}{Bin Ran}.} \bibinfo{year}{2014}\natexlab{}.
\newblock \showarticletitle{Tensor completion via a multi-linear low-n-rank
  factorization model}.
\newblock \bibinfo{journal}{{\em Neurocomputing\/}} (\bibinfo{year}{2014}).
\newblock


\bibitem[\protect\citeauthoryear{Tan, Feng, Feng, Wang, and Zhang}{Tan
  et~al\mbox{.}}{2013b}]%
        {tan2013traffic}
\bibfield{author}{\bibinfo{person}{Huachun Tan}, \bibinfo{person}{Jianshuai
  Feng}, \bibinfo{person}{Guangdong Feng}, \bibinfo{person}{Wuhong Wang}, {and}
  \bibinfo{person}{Yu-Jin Zhang}.} \bibinfo{year}{2013}\natexlab{b}.
\newblock \showarticletitle{Traffic volume data outlier recovery via tensor
  model}.
\newblock \bibinfo{journal}{{\em Mathematical Problems in Engineering\/}}
  (\bibinfo{year}{2013}).
\newblock


\bibitem[\protect\citeauthoryear{Tan, Wu, Feng, Wang, and Ran}{Tan
  et~al\mbox{.}}{2013c}]%
        {tan2013new}
\bibfield{author}{\bibinfo{person}{Huachun Tan}, \bibinfo{person}{Yuankai Wu},
  \bibinfo{person}{Guangdong Feng}, \bibinfo{person}{Wuhong Wang}, {and}
  \bibinfo{person}{Bin Ran}.} \bibinfo{year}{2013}\natexlab{c}.
\newblock \showarticletitle{A new traffic prediction method based on dynamic
  tensor completion}.
\newblock \bibinfo{journal}{{\em Procedia-Social and Behavioral Sciences\/}}
  (\bibinfo{year}{2013}).
\newblock


\bibitem[\protect\citeauthoryear{Tang, Shu, Qi, Li, Wang, Yan, and Jain}{Tang
  et~al\mbox{.}}{2017}]%
        {tang2017tri}
\bibfield{author}{\bibinfo{person}{Jinhui Tang}, \bibinfo{person}{Xiangbo Shu},
  \bibinfo{person}{Guo-Jun Qi}, \bibinfo{person}{Zechao Li},
  \bibinfo{person}{Meng Wang}, \bibinfo{person}{Shuicheng Yan}, {and}
  \bibinfo{person}{Ramesh Jain}.} \bibinfo{year}{2017}\natexlab{}.
\newblock \showarticletitle{Tri-clustered tensor completion for social-aware
  image tag refinement}.
\newblock \bibinfo{journal}{{\em IEEE transactions on pattern analysis and
  machine intelligence\/}} (\bibinfo{year}{2017}).
\newblock


\bibitem[\protect\citeauthoryear{Tomasi and Bro}{Tomasi and Bro}{2005}]%
        {tomasi2005parafac}
\bibfield{author}{\bibinfo{person}{Giorgio Tomasi} {and}
  \bibinfo{person}{Rasmus Bro}.} \bibinfo{year}{2005}\natexlab{}.
\newblock \showarticletitle{PARAFAC and missing values}.
\newblock \bibinfo{journal}{{\em Chemometrics and Intelligent Laboratory
  Systems\/}} (\bibinfo{year}{2005}).
\newblock


\bibitem[\protect\citeauthoryear{Tomioka, Hayashi, and Kashima}{Tomioka
  et~al\mbox{.}}{2010}]%
        {tomioka2010estimation}
\bibfield{author}{\bibinfo{person}{Ryota Tomioka}, \bibinfo{person}{Kohei
  Hayashi}, {and} \bibinfo{person}{Hisashi Kashima}.}
  \bibinfo{year}{2010}\natexlab{}.
\newblock \showarticletitle{Estimation of low-rank tensors via convex
  optimization}.
\newblock \bibinfo{journal}{{\em arXiv\/}} (\bibinfo{year}{2010}).
\newblock


\bibitem[\protect\citeauthoryear{Tomioka, Suzuki, Hayashi, and Kashima}{Tomioka
  et~al\mbox{.}}{2011}]%
        {tomioka2011statistical}
\bibfield{author}{\bibinfo{person}{Ryota Tomioka}, \bibinfo{person}{Taiji
  Suzuki}, \bibinfo{person}{Kohei Hayashi}, {and} \bibinfo{person}{Hisashi
  Kashima}.} \bibinfo{year}{2011}\natexlab{}.
\newblock \showarticletitle{Statistical performance of convex tensor
  decomposition}.
\newblock \bibinfo{journal}{{\em NIPS\/}} (\bibinfo{year}{2011}).
\newblock


\bibitem[\protect\citeauthoryear{Tropp and Gilbert}{Tropp and Gilbert}{2007}]%
        {tropp2007signal}
\bibfield{author}{\bibinfo{person}{Joel~A Tropp} {and} \bibinfo{person}{Anna~C
  Gilbert}.} \bibinfo{year}{2007}\natexlab{}.
\newblock \showarticletitle{Signal recovery from random measurements via
  orthogonal matching pursuit}.
\newblock \bibinfo{journal}{{\em IEEE Transactions on information theory\/}}
  (\bibinfo{year}{2007}).
\newblock


\bibitem[\protect\citeauthoryear{Tucker}{Tucker}{1966}]%
        {tucker1966some}
\bibfield{author}{\bibinfo{person}{Ledyard~R Tucker}.}
  \bibinfo{year}{1966}\natexlab{}.
\newblock \showarticletitle{Some mathematical notes on three-mode factor
  analysis}.
\newblock \bibinfo{journal}{{\em Psychometrika\/}} (\bibinfo{year}{1966}).
\newblock


\bibitem[\protect\citeauthoryear{Vasilescu and Terzopoulos}{Vasilescu and
  Terzopoulos}{2002}]%
        {vasilescu2002multilinear}
\bibfield{author}{\bibinfo{person}{M Vasilescu} {and} \bibinfo{person}{Demetri
  Terzopoulos}.} \bibinfo{year}{2002}\natexlab{}.
\newblock \showarticletitle{Multilinear analysis of image ensembles:
  Tensorfaces}.
\newblock \bibinfo{journal}{{\em Computer Vision---ECCV 2002\/}}
  (\bibinfo{year}{2002}).
\newblock


\bibitem[\protect\citeauthoryear{Vervliet, Debals, and De~Lathauwer}{Vervliet
  et~al\mbox{.}}{2017}]%
        {vervliet2017canonical}
\bibfield{author}{\bibinfo{person}{NICO Vervliet}, \bibinfo{person}{OTTO
  Debals}, {and} \bibinfo{person}{LIEVEN De~Lathauwer}.}
  \bibinfo{year}{2017}\natexlab{}.
\newblock \bibinfo{booktitle}{{\em Canonical polyadic decomposition of
  incomplete tensors with linearly constrained factors}}.
\newblock \bibinfo{type}{{T}echnical {R}eport}. \bibinfo{institution}{Technical
  Report 16--172, ESAT--STADIUS, KU Leuven, Belgium}.
\newblock


\bibitem[\protect\citeauthoryear{Walczak and Massart}{Walczak and
  Massart}{2001}]%
        {walczak2001dealing}
\bibfield{author}{\bibinfo{person}{B Walczak} {and} \bibinfo{person}{DL
  Massart}.} \bibinfo{year}{2001}\natexlab{}.
\newblock \showarticletitle{Dealing with missing data: Part I}.
\newblock \bibinfo{journal}{{\em Chemometrics and Intelligent Laboratory
  Systems\/}} (\bibinfo{year}{2001}).
\newblock


\bibitem[\protect\citeauthoryear{Wang, Nie, and Huang}{Wang
  et~al\mbox{.}}{2014a}]%
        {wang2014low}
\bibfield{author}{\bibinfo{person}{Hua Wang}, \bibinfo{person}{Feiping Nie},
  {and} \bibinfo{person}{Heng Huang}.} \bibinfo{year}{2014}\natexlab{a}.
\newblock \showarticletitle{Low-Rank Tensor Completion with Spatio-Temporal
  Consistency.}. In \bibinfo{booktitle}{{\em AAAI}}.
\newblock


\bibitem[\protect\citeauthoryear{Wang, Chen, Ghosh, Denny, Kho, Chen, Malin,
  and Sun}{Wang et~al\mbox{.}}{2015}]%
        {wang2015rubik}
\bibfield{author}{\bibinfo{person}{Yichen Wang}, \bibinfo{person}{Robert Chen},
  \bibinfo{person}{Joydeep Ghosh}, \bibinfo{person}{Joshua~C Denny},
  \bibinfo{person}{Abel Kho}, \bibinfo{person}{You Chen},
  \bibinfo{person}{Bradley~A Malin}, {and} \bibinfo{person}{Jimeng Sun}.}
  \bibinfo{year}{2015}\natexlab{}.
\newblock \showarticletitle{Rubik: Knowledge guided tensor factorization and
  completion for health data analytics}.
\newblock \bibinfo{journal}{{\em KDD\/}} (\bibinfo{year}{2015}).
\newblock


\bibitem[\protect\citeauthoryear{Wang, Zheng, and Xue}{Wang
  et~al\mbox{.}}{2014b}]%
        {wang2014travel}
\bibfield{author}{\bibinfo{person}{Yilun Wang}, \bibinfo{person}{Yu Zheng},
  {and} \bibinfo{person}{Yexiang Xue}.} \bibinfo{year}{2014}\natexlab{b}.
\newblock \showarticletitle{Travel time estimation of a path using sparse
  trajectories}.
\newblock \bibinfo{journal}{{\em KDD\/}} (\bibinfo{year}{2014}).
\newblock


\bibitem[\protect\citeauthoryear{Wimalawarne, Yamada, and
  Mamitsuka}{Wimalawarne et~al\mbox{.}}{2017}]%
        {wimalawarne2017convex}
\bibfield{author}{\bibinfo{person}{Kishan Wimalawarne}, \bibinfo{person}{Makoto
  Yamada}, {and} \bibinfo{person}{Hiroshi Mamitsuka}.}
  \bibinfo{year}{2017}\natexlab{}.
\newblock \showarticletitle{Convex Coupled Matrix and Tensor Completion}.
\newblock \bibinfo{journal}{{\em arXiv preprint arXiv:1705.05197\/}}
  (\bibinfo{year}{2017}).
\newblock


\bibitem[\protect\citeauthoryear{Xia and Yuan}{Xia and Yuan}{2017}]%
        {xia2017polynomial}
\bibfield{author}{\bibinfo{person}{Dong Xia} {and} \bibinfo{person}{Ming
  Yuan}.} \bibinfo{year}{2017}\natexlab{}.
\newblock \showarticletitle{On Polynomial Time Methods for Exact Low Rank
  Tensor Completion}.
\newblock \bibinfo{journal}{{\em arXiv preprint arXiv:1702.06980\/}}
  (\bibinfo{year}{2017}).
\newblock


\bibitem[\protect\citeauthoryear{Xiao, Li, Gao, Wang, Ge, Fan, Vu, and
  Turaga}{Xiao et~al\mbox{.}}{2015}]%
        {xiao2015believe}
\bibfield{author}{\bibinfo{person}{Houping Xiao}, \bibinfo{person}{Yaliang Li},
  \bibinfo{person}{Jing Gao}, \bibinfo{person}{Fei Wang},
  \bibinfo{person}{Liang Ge}, \bibinfo{person}{Wei Fan},
  \bibinfo{person}{Long~H Vu}, {and} \bibinfo{person}{Deepak~S Turaga}.}
  \bibinfo{year}{2015}\natexlab{}.
\newblock \showarticletitle{Believe it today or tomorrow? detecting
  untrustworthy information from dynamic multi-source data}. In
  \bibinfo{booktitle}{{\em SDM}}.
\newblock


\bibitem[\protect\citeauthoryear{Xiong, Liu, Xiong, Li, Wang, and Liang}{Xiong
  et~al\mbox{.}}{2018}]%
        {xiong2018field}
\bibfield{author}{\bibinfo{person}{Biao Xiong}, \bibinfo{person}{Qiegen Liu},
  \bibinfo{person}{Jiaojiao Xiong}, \bibinfo{person}{Sanqian Li},
  \bibinfo{person}{Shanshan Wang}, {and} \bibinfo{person}{Dong Liang}.}
  \bibinfo{year}{2018}\natexlab{}.
\newblock \showarticletitle{Field-of-Experts Filters Guided Tensor Completion}.
\newblock \bibinfo{journal}{{\em IEEE Transactions on Multimedia\/}}
  (\bibinfo{year}{2018}).
\newblock


\bibitem[\protect\citeauthoryear{Xiong, Chen, Huang, Schneider, and
  Carbonell}{Xiong et~al\mbox{.}}{2010}]%
        {xiong2010temporal}
\bibfield{author}{\bibinfo{person}{Liang Xiong}, \bibinfo{person}{Xi Chen},
  \bibinfo{person}{Tzu-Kuo Huang}, \bibinfo{person}{Jeff Schneider}, {and}
  \bibinfo{person}{Jaime~G Carbonell}.} \bibinfo{year}{2010}\natexlab{}.
\newblock \showarticletitle{Temporal collaborative filtering with bayesian
  probabilistic tensor factorization}.
\newblock \bibinfo{journal}{{\em ICDM\/}} (\bibinfo{year}{2010}).
\newblock


\bibitem[\protect\citeauthoryear{Xu, Hao, Yin, and Su}{Xu
  et~al\mbox{.}}{2013}]%
        {xu2013parallel}
\bibfield{author}{\bibinfo{person}{Yangyang Xu}, \bibinfo{person}{Ruru Hao},
  \bibinfo{person}{Wotao Yin}, {and} \bibinfo{person}{Zhixun Su}.}
  \bibinfo{year}{2013}\natexlab{}.
\newblock \showarticletitle{Parallel matrix factorization for low-rank tensor
  completion}.
\newblock \bibinfo{journal}{{\em arXiv preprint arXiv:1312.1254\/}}
  (\bibinfo{year}{2013}).
\newblock


\bibitem[\protect\citeauthoryear{Xu and Yin}{Xu and Yin}{2013}]%
        {xu2013block}
\bibfield{author}{\bibinfo{person}{Yangyang Xu} {and} \bibinfo{person}{Wotao
  Yin}.} \bibinfo{year}{2013}\natexlab{}.
\newblock \showarticletitle{A block coordinate descent method for regularized
  multiconvex optimization with applications to nonnegative tensor
  factorization and completion}.
\newblock \bibinfo{journal}{{\em SIAM Journal on imaging sciences\/}}
  (\bibinfo{year}{2013}).
\newblock


\bibitem[\protect\citeauthoryear{Xu, Yin, Wen, and Zhang}{Xu
  et~al\mbox{.}}{2012}]%
        {xu2012alternating}
\bibfield{author}{\bibinfo{person}{Yangyang Xu}, \bibinfo{person}{Wotao Yin},
  \bibinfo{person}{Zaiwen Wen}, {and} \bibinfo{person}{Yin Zhang}.}
  \bibinfo{year}{2012}\natexlab{}.
\newblock \showarticletitle{An alternating direction algorithm for matrix
  completion with nonnegative factors}.
\newblock \bibinfo{journal}{{\em Frontiers of Mathematics in China\/}}
  (\bibinfo{year}{2012}).
\newblock


\bibitem[\protect\citeauthoryear{Y{\i}lmaz, Cemgil, and Simsekli}{Y{\i}lmaz
  et~al\mbox{.}}{2011}]%
        {yilmaz2011generalised}
\bibfield{author}{\bibinfo{person}{Kenan~Y Y{\i}lmaz}, \bibinfo{person}{Ali~T
  Cemgil}, {and} \bibinfo{person}{Umut Simsekli}.}
  \bibinfo{year}{2011}\natexlab{}.
\newblock \showarticletitle{Generalised coupled tensor factorisation}.
\newblock \bibinfo{journal}{{\em NIPS\/}} (\bibinfo{year}{2011}).
\newblock


\bibitem[\protect\citeauthoryear{Yin and Kaynak}{Yin and Kaynak}{2015}]%
        {yin2015big}
\bibfield{author}{\bibinfo{person}{Shen Yin} {and} \bibinfo{person}{Okyay
  Kaynak}.} \bibinfo{year}{2015}\natexlab{}.
\newblock \showarticletitle{Big data for modern industry: challenges and trends
  [point of view]}.
\newblock \bibinfo{journal}{{\it Proc. IEEE}} (\bibinfo{year}{2015}).
\newblock


\bibitem[\protect\citeauthoryear{Ying, Lu, Wei, Cai, Guo, Wu, Chen, and
  Qu}{Ying et~al\mbox{.}}{2017}]%
        {ying2017hankel}
\bibfield{author}{\bibinfo{person}{Jiaxi Ying}, \bibinfo{person}{Hengfa Lu},
  \bibinfo{person}{Qingtao Wei}, \bibinfo{person}{Jian-Feng Cai},
  \bibinfo{person}{Di Guo}, \bibinfo{person}{Jihui Wu}, \bibinfo{person}{Zhong
  Chen}, {and} \bibinfo{person}{Xiaobo Qu}.} \bibinfo{year}{2017}\natexlab{}.
\newblock \showarticletitle{Hankel Matrix Nuclear Norm Regularized Tensor
  Completion for N-dimensional Exponential Signals}.
\newblock \bibinfo{journal}{{\em IEEE Transactions on Signal Processing\/}}
  (\bibinfo{year}{2017}).
\newblock


\bibitem[\protect\citeauthoryear{Yokota, Zhao, and Cichocki}{Yokota
  et~al\mbox{.}}{2016}]%
        {yokota2016smooth}
\bibfield{author}{\bibinfo{person}{Tatsuya Yokota}, \bibinfo{person}{Qibin
  Zhao}, {and} \bibinfo{person}{Andrzej Cichocki}.}
  \bibinfo{year}{2016}\natexlab{}.
\newblock \showarticletitle{Smooth PARAFAC decomposition for tensor
  completion}.
\newblock \bibinfo{journal}{{\em IEEE Transactions on Signal Processing\/}}
  (\bibinfo{year}{2016}).
\newblock


\bibitem[\protect\citeauthoryear{Yu, Hsieh, Si, and Dhillon}{Yu
  et~al\mbox{.}}{2012}]%
        {yu2012scalable}
\bibfield{author}{\bibinfo{person}{Hsiang-Fu Yu}, \bibinfo{person}{Cho-Jui
  Hsieh}, \bibinfo{person}{Si Si}, {and} \bibinfo{person}{Inderjit Dhillon}.}
  \bibinfo{year}{2012}\natexlab{}.
\newblock \showarticletitle{Scalable coordinate descent approaches to parallel
  matrix factorization for recommender systems}. In \bibinfo{booktitle}{{\em
  Data Mining (ICDM), 2012 IEEE 12th International Conference on}}. IEEE,
  \bibinfo{pages}{765--774}.
\newblock


\bibitem[\protect\citeauthoryear{Yu, Cheng, and Liu}{Yu et~al\mbox{.}}{2015}]%
        {Rose:Tucker}
\bibfield{author}{\bibinfo{person}{Rose Yu}, \bibinfo{person}{Dehua Cheng},
  {and} \bibinfo{person}{Yan Liu}.} \bibinfo{year}{2015}\natexlab{}.
\newblock \showarticletitle{Accelerated online low-rank tensor learning for
  multivariate spatio-temporal streams}.
\newblock \bibinfo{journal}{{\em ICML\/}} (\bibinfo{year}{2015}).
\newblock


\bibitem[\protect\citeauthoryear{Yuan and Zhang}{Yuan and Zhang}{2016}]%
        {yuan2016incoherent}
\bibfield{author}{\bibinfo{person}{Ming Yuan} {and} \bibinfo{person}{Cun-Hui
  Zhang}.} \bibinfo{year}{2016}\natexlab{}.
\newblock \showarticletitle{Incoherent tensor norms and their applications in
  higher order tensor completion}.
\newblock \bibinfo{journal}{{\em arXiv preprint arXiv:1606.03504\/}}
  (\bibinfo{year}{2016}).
\newblock


\bibitem[\protect\citeauthoryear{Zafarani, Abbasi, and Liu}{Zafarani
  et~al\mbox{.}}{2014}]%
        {zafarani2014social}
\bibfield{author}{\bibinfo{person}{Reza Zafarani},
  \bibinfo{person}{Mohammad~Ali Abbasi}, {and} \bibinfo{person}{Huan Liu}.}
  \bibinfo{year}{2014}\natexlab{}.
\newblock \bibinfo{booktitle}{{\em Social media mining: an introduction}}.
\newblock \bibinfo{publisher}{Cambridge University Press}.
\newblock


\bibitem[\protect\citeauthoryear{Zhang and Aeron}{Zhang and Aeron}{2017}]%
        {zhang2017exact}
\bibfield{author}{\bibinfo{person}{Zemin Zhang} {and} \bibinfo{person}{Shuchin
  Aeron}.} \bibinfo{year}{2017}\natexlab{}.
\newblock \showarticletitle{Exact tensor completion using t-svd}.
\newblock \bibinfo{journal}{{\em IEEE Transactions on Signal Processing\/}}
  (\bibinfo{year}{2017}).
\newblock


\bibitem[\protect\citeauthoryear{Zhang, Ely, Aeron, Hao, and Kilmer}{Zhang
  et~al\mbox{.}}{2014}]%
        {zhang2014novel}
\bibfield{author}{\bibinfo{person}{Zemin Zhang}, \bibinfo{person}{Gregory Ely},
  \bibinfo{person}{Shuchin Aeron}, \bibinfo{person}{Ning Hao}, {and}
  \bibinfo{person}{Misha Kilmer}.} \bibinfo{year}{2014}\natexlab{}.
\newblock \showarticletitle{Novel methods for multilinear data completion and
  de-noising based on tensor-SVD}.
\newblock \bibinfo{journal}{{\em IEEE Conference on Computer Vision and Pattern
  Recognition\/}} (\bibinfo{year}{2014}).
\newblock


\bibitem[\protect\citeauthoryear{Zhao, Zhang, and Cichocki}{Zhao
  et~al\mbox{.}}{2015a}]%
        {zhao2015bayesian}
\bibfield{author}{\bibinfo{person}{Qibin Zhao}, \bibinfo{person}{Liqing Zhang},
  {and} \bibinfo{person}{Andrzej Cichocki}.} \bibinfo{year}{2015}\natexlab{a}.
\newblock \showarticletitle{Bayesian CP factorization of incomplete tensors
  with automatic rank determination}.
\newblock \bibinfo{journal}{{\em IEEE transactions on pattern analysis and
  machine intelligence\/}} (\bibinfo{year}{2015}).
\newblock


\bibitem[\protect\citeauthoryear{Zhao, Zhang, and Cichocki}{Zhao
  et~al\mbox{.}}{2015b}]%
        {zhao2015bayesianTucker}
\bibfield{author}{\bibinfo{person}{Qibin Zhao}, \bibinfo{person}{Liqing Zhang},
  {and} \bibinfo{person}{Andrzej Cichocki}.} \bibinfo{year}{2015}\natexlab{b}.
\newblock \showarticletitle{Bayesian sparse Tucker models for dimension
  reduction and tensor completion}.
\newblock \bibinfo{journal}{{\em arXiv preprint arXiv:1505.02343\/}}
  (\bibinfo{year}{2015}).
\newblock


\bibitem[\protect\citeauthoryear{Zhao, Zhou, Zhang, Cichocki, and Amari}{Zhao
  et~al\mbox{.}}{2014}]%
        {zhao2014robust}
\bibfield{author}{\bibinfo{person}{Qibin Zhao}, \bibinfo{person}{Guoxu Zhou},
  \bibinfo{person}{Liqing Zhang}, \bibinfo{person}{Andrzej Cichocki}, {and}
  \bibinfo{person}{Shun-ichi Amari}.} \bibinfo{year}{2014}\natexlab{}.
\newblock \showarticletitle{Robust bayesian tensor factorization for incomplete
  multiway data}.
\newblock \bibinfo{journal}{{\em CoRR abs/1410.2386\/}} (\bibinfo{year}{2014}).
\newblock


\bibitem[\protect\citeauthoryear{Zheng, Cao, Zheng, Xie, and Yang}{Zheng
  et~al\mbox{.}}{2010}]%
        {zheng2010collaborative}
\bibfield{author}{\bibinfo{person}{Vincent~Wenchen Zheng}, \bibinfo{person}{Bin
  Cao}, \bibinfo{person}{Yu Zheng}, \bibinfo{person}{Xing Xie}, {and}
  \bibinfo{person}{Qiang Yang}.} \bibinfo{year}{2010}\natexlab{}.
\newblock \showarticletitle{Collaborative Filtering Meets Mobile
  Recommendation: A User-Centered Approach.}
\newblock \bibinfo{journal}{{\em AAAI\/}} (\bibinfo{year}{2010}).
\newblock


\bibitem[\protect\citeauthoryear{Zhou, Vinh, Bailey, Jia, and Davidson}{Zhou
  et~al\mbox{.}}{2016}]%
        {zhou2016accelerating}
\bibfield{author}{\bibinfo{person}{Shuo Zhou}, \bibinfo{person}{Nguyen~Xuan
  Vinh}, \bibinfo{person}{James Bailey}, \bibinfo{person}{Yunzhe Jia}, {and}
  \bibinfo{person}{Ian Davidson}.} \bibinfo{year}{2016}\natexlab{}.
\newblock \showarticletitle{Accelerating online CP decompositions for higher
  order tensors}.
\newblock \bibinfo{journal}{{\em KDD\/}} (\bibinfo{year}{2016}).
\newblock


\bibitem[\protect\citeauthoryear{Zhou, Qian, Shen, Zhang, and Xu}{Zhou
  et~al\mbox{.}}{2017}]%
        {zhou2017tensor}
\bibfield{author}{\bibinfo{person}{Tengfei Zhou}, \bibinfo{person}{Hui Qian},
  \bibinfo{person}{Zebang Shen}, \bibinfo{person}{Chao Zhang}, {and}
  \bibinfo{person}{Congfu Xu}.} \bibinfo{year}{2017}\natexlab{}.
\newblock \showarticletitle{Tensor completion with side information: A
  riemannian manifold approach}. In \bibinfo{booktitle}{{\em IJCAI}}.
\newblock


\bibitem[\protect\citeauthoryear{Zhou, Wilkinson, Schreiber, and Pan}{Zhou
  et~al\mbox{.}}{2008}]%
        {zhou2008large}
\bibfield{author}{\bibinfo{person}{Yunhong Zhou}, \bibinfo{person}{Dennis
  Wilkinson}, \bibinfo{person}{Robert Schreiber}, {and} \bibinfo{person}{Rong
  Pan}.} \bibinfo{year}{2008}\natexlab{}.
\newblock \showarticletitle{Large-scale parallel collaborative filtering for
  the netflix prize}. In \bibinfo{booktitle}{{\em International Conference on
  Algorithmic Applications in Management}}. Springer,
  \bibinfo{pages}{337--348}.
\newblock


\bibitem[\protect\citeauthoryear{Zinkevich, Weimer, Li, and Smola}{Zinkevich
  et~al\mbox{.}}{2010}]%
        {zinkevich2010parallelized}
\bibfield{author}{\bibinfo{person}{Martin Zinkevich}, \bibinfo{person}{Markus
  Weimer}, \bibinfo{person}{Lihong Li}, {and} \bibinfo{person}{Alex~J Smola}.}
  \bibinfo{year}{2010}\natexlab{}.
\newblock \showarticletitle{Parallelized stochastic gradient descent}. In
  \bibinfo{booktitle}{{\em Advances in neural information processing systems}}.
  \bibinfo{pages}{2595--2603}.
\newblock


\end{thebibliography}

\end{document}